\documentclass[runningheads]{llncs}

\usepackage{eccv}

\usepackage{eccvabbrv}
\usepackage{amsmath}
\usepackage{graphicx}
\usepackage{booktabs}
\usepackage{multirow}
\usepackage{stmaryrd}
\usepackage{pgfplots}
\usepackage{xfakebold}

\usepackage[accsupp]{axessibility}  % Improves PDF readability for those with disabilities.
\usepackage{comment}
\usepackage{fontawesome5}
\usepackage{hyperref}

\usepackage{orcidlink}

\newcommand{\mstd}[2]{$\text{#1}_{\pm#2}$}
\newcommand*{\yes}{\checkmark}
\newcommand*{\no}{\textcolor{gray}{--}}
\newcommand*{\blarrow}{\rotatebox[origin=c]{270}{$\Rsh$}}
\newcommand{\pqs}{$\text{PQ}_{16}$}

\begin{document}

%\title{ML-PanDA: Mask-level Confidence\\ for Panoptic Domain Adaptation}
%\title{MLC-PanDA: Mask-level Confidence\\ for Panoptic Domain Adaptation}
%\title{$\mu$-PanDA: Mask-level Confidence\\ for Panoptic Domain Adaptation}
\title{MC-PanDA: Mask Confidence
  \texorpdfstring{\\}{}
  for Panoptic Domain Adaptation}

\titlerunning{MC-PanDA: Mask Confidence for Panoptic Domain Adaptation}

\author{Ivan Martinović\inst{1}\orcidlink{0009-0001-6497-8417} \and
Josip Šarić\inst{1}\orcidlink{0000-0001-7262-550X} \and
Siniša Šegvić\inst{1}\orcidlink{0000-0001-7378-0536}}

\authorrunning{I.~Martinović et al.}

\institute{University of Zagreb, Faculty of Electrical Engineering and Computing, Croatia
\email{\{ivan.martinovic,josip.saric,sinisa.segvic\}@fer.hr}}

\maketitle

\begin{abstract}
Domain adaptive panoptic segmentation
promises to resolve
the long tail of corner cases
in natural scene understanding.
Previous state of the art addresses this problem
with cross-task consistency, 
careful system-level optimization
and heuristic improvement
of teacher predictions.
In contrast, we propose to build upon
remarkable capability of mask transformers
to estimate their own prediction uncertainty.
Our method 
avoids noise amplification
by leveraging fine-grained confidence
of panoptic teacher predictions.
In particular, we modulate the loss 
with mask-wide confidence and discourage
back-propagation in pixels
with uncertain teacher or confident student.
Experimental evaluation on standard benchmarks
reveals a substantial contribution
of the proposed selection techniques. 
We report 47.4 PQ
on Synthia$\rightarrow$Cityscapes,
which corresponds to an improvement 
of 6.2 
percentage points 
over the state of the art.
The source code
is available at
\texttt{\href{https://github.com/helen1c/MC-PanDA}{github.com/helen1c/MC-PanDA}}.

\end{abstract}

\section{Introduction}
\label{sec:intro}
Panoptic segmentation is
a recently developed technique
that aims to unify
all computer vision tasks
related to scene understanding~\cite{kirillov2019panoptic,carion2020end,cheng2021per}.
Consistent performance improvement
has brought exciting applications
within reach of practitioners~\cite{cheng2022masked,yu2022k,li2023mask}.
However, many important domains
require application-specific datasets
that are very expensive to annotate~\cite{zlateski18cvpr,zendel19cvprw}.
Furthermore, even fairly standard domains
suffer from the long tail of corner cases
that are not easily collected~\cite{zendel18eccv,sakaridis21iccv,uijlings22eccv,zendel22cvpr}.
\begin{figure}[b!]
  \centering
  \includegraphics[width=\textwidth]
    {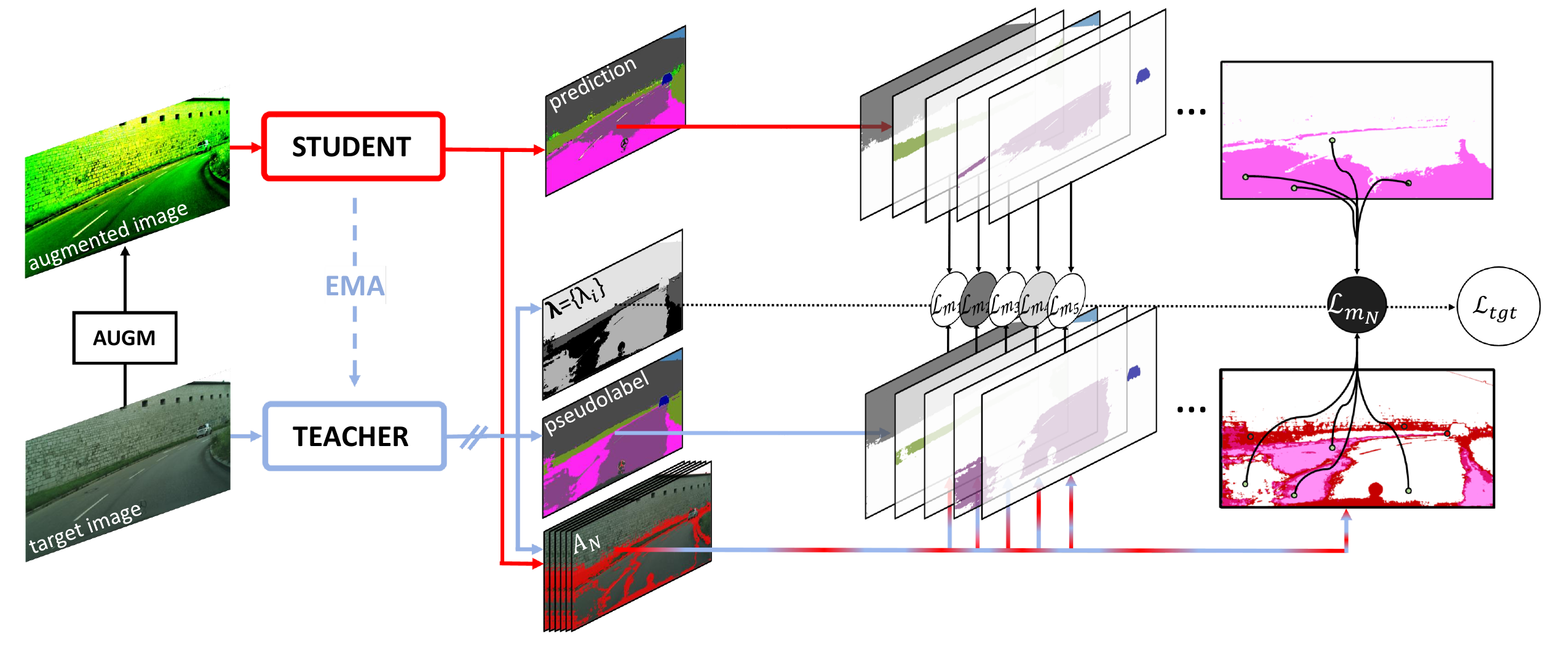}
  \caption{
MC-PanDA complements 
the Mean Teacher 
for domain-adaptive panoptics 
with fine-grained uncertainty quantification. 
We modulate the dense localization loss 
of the $i$-th mask $\mathcal{L}_{m_{i}}$ with the mask-wide confidence $\lambda_i$, 
and sample it with respect to pixel-level
affinity $A_i$ that blends 
teacher confidence $\Phi$
and student uncertainty (\cf~Eq.~\ref{eq:unc}).
Our contributions encourage cautious 
self-learning from uncertain pseudo-labels.
Note that we show detailed 
loss sampling and weighting
only for the mask $N$. 
This procedure is carried out 
for each student mask that maps
to a non-empty teacher mask. 
For simplicity, the figure omits source domain branch
and SegMix augmentation.}
  \label{fig:overview}
\end{figure}
Synthetic datasets offer a promising avenue
to address the shortage of training labels, 
provided that domain shift 
is somehow accounted for~\cite{Ros_2016_CVPR,richter16eccv,kim20cvpr}.
Reducing the sensitivity to domain shift 
appears as a reasonable stance,
since there will always be
some divergence between
the train and the test data~\cite{blum21ijcv,lambert23pami,zhou23pami,bevandic24ijcv}.
This state of affairs makes
unsupervised domain adaptation
an extremely attractive goal~\cite{bendavid10ml,ganin15icml,sun16eccvw}.

Most of the previous work
in dense domain adaptation 
addresses semantic segmentation~\cite{zheng2021rectifying,zhou2022uncertainty ,chen23acmmm,hoyer24pami, hoyer23cvpr}.
However, semantic segmentation methods 
are not easily 
upgraded to panoptic segmentation
since they can not distinguish instances 
of the same class.
This incompatibilty contributes 
to the relative scarceness 
of domain adaptive panoptic approaches 
in the literature~\cite{huang2021cross,zhang23cvpr,Saha_2023_ICCV}.

Many approaches
to discriminative learning on unlabeled images 
encourage smooth decision boundaries by consistency training.
In practice, this boils down to learning 
invariance over a family of perturbations~\cite{lenc19ijcv}
by requiring similar predictions 
in related training samples~\cite{chapelle06book}.
The most effective methods involve strong perturbations~\cite{french20bmvc}
and disallow loss propagation
from the perturbed samples 
towards the clean ones~\cite{miyato19pami,grubisic21mva}.
This leads to self-learning setups with two branches,
each of which holds an instance
of the desired production model.
The student branch receives the perturbed image
and propagates the gradients 
to model parameters.
The teacher branch receives the corresponding clean image,
and stops the gradients.
Best results are often obtained
in the Mean Teacher setup 
where the teacher corresponds
to the exponentially averaged student~\cite{tarvainen17neurips}.

Consistency learning powers 
the most successful domain adaptive approaches
for semantic segmentation~\cite{chen23acmmm,hoyer24pami, hoyer23cvpr}.
Previous work in domain adaptive panoptics
also start from this setup.
CVRN~\cite{huang2021cross} relies on cross-task and cross-style terms.
UniDAformer~\cite{zhang23cvpr} improves the teacher predictions
through hierarchical mask calibration.
EDAPS~\cite{Saha_2023_ICCV} outperforms all previous approaches 
by a wide margin through powerful shared encoder, separate decoders for things and stuff, as well as
modulating the self-supervised loss
with image-wide confidence.

This paper empowers domain adaptive panoptics 
with fine-grained uncertainty quantification.
We denote our method MC-PanDA
since we implement it atop 
a mask transformer~\cite{cheng2022masked}
that localizes instances and stuff segments
with dense activation maps called masks.
Figure~\ref{fig:overview} shows that the teacher recovers 
mask-wide confidences $\lambda_i$ and 
panoptic predictions.
Moreover, we recover 
the dense sampling affinity $A_i$ for each mask $i$
by blending fine-grained teacher  
confidence $\Phi$ 
with student uncertainty. 
Mask-wide confidences $\lambda_i$ and sampling affinities $A_i$ allow us
to discourage self-learning in uncertain masks
and at locations with inappropriate pixel-level uncertainty.
In summary, we contribute
two novel techniques
for domain adaptive panoptics:
i) mask-wide loss scaling (MLS) 
according to aggregated region-wide confidence $\lambda_i$,
and ii) confidence-based point filtering (CBPF) 
that favours learning
in points with
confident teacher 
and uncertain 
student as indicated by $A_i$.
Experiments on standard benchmarks
for domain adaptive panoptics 
reveal substantial improvements 
of the generalization performance.
Our method outperforms 
the state of the art
by a large margin and 
brings us a step closer 
to real-world applications of synthetic training data.

\section{Related Work}\label{sec:related_work}
\textbf{Panoptic Segmentation}.
Although many 
real-world applications
require both
semantic and 
instance segmentation, 
most of the early research
considered these tasks
in isolation.
Recent
definition of 
panoptic segmentation 
as the joint task
brought much attention
to this problem~\cite{kirillov2019panoptic}.
Early attempts 
extend Mask R-CNN~\cite{he2017mask}
with a semantic segmentation branch
and consider different
fusion strategies
for things and stuff predictions~\cite{kirillov2019panoptic,kirillov2019pfpn,li2018tascnet,xiong2019upsnet}.
Panoptic Deeplab extends 
the semantic segmentation pipeline
with class-agnostic instance predictions
based on center and offset regression~\cite{cheng2020panoptic,saric23rs}.

\noindent
\textbf{Unified Panoptics with Mask Transformers.}
Recent work proposes a unified
framework that
localizes instances and stuff segments
with dense sigmoidal maps called \emph{masks}~\cite{yu2022k,cheng2022masked,li2023mask}.
Masks are recovered by scoring dense features 
with the associated embeddings~\cite{cheng2021per}.
Mask embeddings are recovered 
through direct set prediction
with an appropriate transformer module~\cite{carion2020end}.
Besides the simplified inference,
these models 
currently achieve the
state-of-the-art on
panoptic segmentation benchmarks~\cite{zhou2017scene,lin2014microsoft}.
Moreover, several recent works 
point out remarkable capability 
of mask transformers 
to estimate their own prediction uncertainty~\cite{grcic23cvprw}.
Methods based on mask transformers
represent the current state of the art
in dense anomaly detection~\cite{rai23iccv,nayal23iccv,ackermann23bmvc,delic24arxiv}.

\noindent\textbf{Unsupervised Domain Adaptation}.
This field involves 
a source domain
and the target domain.
The source domain contains labeled data
that involves
a domain shift
with respect to the target domain.
Thus, standard training on the 
source domain
fails to bring 
satisfactory generalization
on target 
data~\cite{sakaridis21iccv,zendel18eccv}.
Domain adaptation aims 
to improve the performance
by joint training with
unlabeled target data,
which results 
in the following compound loss~\cite{bendavid10ml,sun16eccvw}:
\begin{align}
        \label{eq:uda}
	\mathcal{L}_{uda} = \mathcal{L}_{src} + \mathcal{L}_{tgt}
\end{align}
Domain adaptation focuses on
the second loss term
to find a way 
to exploit unlabeled target data~\cite{zou2018unsupervised,vu2019advent,zhou2022context,olsson2021classmix}.
The most performant 
approaches build upon 
consistency learning
with Mean Teacher
as described 
in the introduction~\cite{hoyer23cvpr,Saha_2023_ICCV}.
Several approaches~\cite{zheng2021rectifying, zhou2022uncertainty}
for domain-adaptive
semantic segmentation
leverage pixel-level
uncertainty. 
Different from them,
our method aggregates
region-wide uncertainties,
exploits both
the student uncertainty and the teacher uncertainty,
and relies on
sparse point sampling~\cite{kirillov2020pointrend}
instead of per-pixel loss scaling~\cite{zheng2021rectifying,zhou2022uncertainty}.

\noindent
\textbf{Domain Adaptive Panoptics.}
The current state of the art
is achieved by EDAPS~\cite{Saha_2023_ICCV},
which extends consistency learning
with image-wide scaling
of the self-supervised loss~\cite{hoyer2022daformer,tranheden2021dacs,hoyer2022hrda}.
This method encourages 
stable learning by reducing 
the gradient magnitude
in images with 
uncertain teacher predictions.
However,
we find this suboptimal
because it equally 
downgrades all image pixels,
even when the prediction uncertainty 
varies across the image.
This will certainly be the case
when some kind of 
mixing data augmentation is used,
where training images 
are crafted by pasting
source scenery 
into target images~\cite{french20bmvc,tranheden2021dacs}.

Our method is most closely related 
to UniDAformer~\cite{zhang2022hierarchical},
the only prior work 
that considers domain adaptation 
of a mask transformer.
However, their work relies on
handcrafted improvement
of teacher predictions, 
while
we present the first
panoptic adaptation approach
that relies 
upon region-wide and fine-grained uncertainty.
Recent strong performance
of mask transformers
in dense outlier detection~\cite{rai23iccv,grcic23cvprw,nayal23iccv}
suggests that our approach 
has a considerable chance of success.
Contrary, 
UniDAformer underperforms
with mask transformers
and reports the best performance with 
a multi-branch architecture (\cf~Table~\ref{tab:synthetic_to_real_mean} and~\cite{zhang2022hierarchical}).

\section{Method}
We first recap
panoptic segmentation with direct set prediction
in subsection~\ref{sec:panseg_mask}.
Then, we describe
a baseline domain adaptation 
with a mask transformer 
in~\ref{sec:uda_basics}.
Finally,
subsections~\ref{sec:mls} and~\ref{sec:cbpf}
present our contributions
based on per-mask loss modulation 
and loss subsampling
according to pixel-level confidence.

\subsection{Panoptic Segmentation with Direct Set Prediction}
\label{sec:panseg_mask}
Recent methods for scene understanding~\cite{cheng2022masked,li2023mask,yu2022k,carion2020end} 
can handle
semantic, instance or panoptic segmentation 
without changing the loss function
or the model architecture.
These models directly detect 
instances and stuff segments,
and describe them with 
distinct segmentation masks.
The dense feature extractor 
produces pixel embeddings 
$E_p \in \mathbb{R}^{H\times W \times d}$, 
where $H$ and $W$ stand for height and width.
The transformer decoder observes the features
and classifies each of the $N$ mask queries across $(C + 1)$ classes. 
The $(C+1)$-th class (no-object) indicates 
that the corresponding mask 
is unused in this particular image. 
The transformer decoder also produces 
mask embeddings $E_m \in \mathbb{R}^{N \times d}$
that identify the corresponding pixel embeddings $E_p$
through dot-product similarity. 
Combining the two embeddings 
through generalized matmul 
and sigmoid activation
produces pixel-to-mask 
assignments $\sigma \in \mathbb{R}^{N \times H \times W}$.
Thus, 
each mask
is defined with 
$\sigma_i \in \mathbb{R}^{H \times W}$
and class distribution $P_i = (p_{i}^{(1)}, p_{i}^{(2)}, ..., p_{i}^{(C+1)})$.

The training process
minimizes the difference
between the 
ground truth masks $\{(\sigma_i^{GT}, y_i^{GT})\}_i^{N^{GT}}$
and the predictions $\{(\sigma_i, P_i)\}_i^N$.
The loss computation requires
bipartite matching  $\mathcal{M}$
which maps predictions onto 
their ground truth counterparts.
Note that the set of ground truth masks
has to be extended with empty masks 
$(\sigma_i^{GT}=\textbf{0}, y_i^{GT}=C + 1)$
in order to match the number of predictions.
Given $\mathcal{M}$,
we can decompose the loss 
into recognition
and localization:
\begin{align}
    \label{eq:m2f_loss}
    \mathcal{L}^{MT} = \sum_i^N 
    \mathcal{L}_{cls}(P_i, y^{GT}_{\mathcal{M}(i)}) + 
    \hspace{-1em}
    \sum_{ y_{\mathcal{M}(i)}^{GT} \neq C + 1}{
    \hspace{-1.5em}
    \mathcal{L}_{mask}(\sigma_i, \sigma^{GT}_{\mathcal{M}(i)})}
\end{align}
The recognition terms $\mathcal{L}_{cls}$
require correct semantic classification,
while the localization terms $\mathcal{L}_{mask}$
optimize per-pixel assignments.
Note that $\mathcal{L}_{mask}$
is computed only for masks
matched with non-empty ground truth, and
that $\mathcal{L}^{MT}$ from~(\ref{eq:m2f_loss}) 
corresponds to $\mathcal{L}_{src}$ from~(\ref{eq:uda}). 
During inference, 
each pixel (\underline{r}ow, \underline{c}olumn) is assigned 
the mask $M$ that maximizes 
the pixel-level confidence $\rho$ 
that is  expressed as 
a product 
of recognition and localization scores:
\begin{align}
        \label{eq:m2f_inference}
	M(r,c) = \text{argmax}_i(\rho_{i, r,c});
	\quad 
	\rho_{i,r,c}=\max_{y \neq C+1} P_i(y) \cdot \sigma_i (r, c)
\end{align}

\subsection{Baseline Consistency Learning with Mean Teacher}
\label{sec:uda_basics}
This section presents our baseline Mean Teacher setup
for panoptic self-training with mask transformers.
We feed the teacher with clean images
and block the gradients 
through the corresponding branch.
We perturb 
the student images 
with color augmentation \cite{chen2020simple},
random application
of Gaussian smoothing,
and SegMix  - 
our panoptic adaptation
of ClassMix~\cite{olsson2021classmix,tranheden2021dacs}.
Instead of classes,
SegMix
samples half of the 
panoptic segments
from a random source image
and pastes them atop
the target image.
We recover the teacher predictions
and convert them 
to hard pseudo-labels
consisting of the semantic class
and dense assignment map.
We obtain the student loss $\mathcal{L}_{tgt}$ 
from the mask transformer loss $\mathcal{L}^{MT}$
by replacing the ground truth
with the pseudo-labels,
\cf~equations~(\ref{eq:uda}) and~(\ref{eq:m2f_loss}).
We initialize the teacher
with pretrained weights
and keep them frozen
in the early stage of 
domain adaptive training \cite{berrada2024guided}.
After that,
we set the teacher to the
temporal exponential moving average 
(EMA) of the student
\cite{tarvainen2017mean}.
Nevertheless, 
our baseline students
still tend to 
deteriorate due 
to noisy pseudo-labels.
Moreover,
we notice that
our baseline teachers 
produce many 
false positive masks.
Training on such pseudo-labels tends to amplify the noise
due to Mean Teacher setup.
The following subsections 
present our solution to this problem.

\subsection{Mask-wide Loss Scaling}
\label{sec:mls}
Our baseline underperforms
due to positive feedback loop 
between the student and the teacher.
We propose to alleviate this effect
by modulating the per-mask 
localization term $\mathcal{L}_{mask}$ 
of the student loss $\mathcal{L}_{tgt}$
with mask-wide teacher confidence $\lambda_i$. 
This allows the student to learn 
from the confident teacher masks
while postponing the learning
from the uncertain ones.
We scale our localization loss as follows:
\begin{align}
\label{eq:lambda}
	\mathcal{L}_\text{loc}^{\text{MC}}=
	\hspace{-1.5em}
	\sum_{ y_{\mathcal{M}(i)}^\text{teach} \neq C + 1}{
	\hspace{-1.5em}
	\lambda_i \cdot \mathcal{L}_{mask}
	(\sigma_i, \sigma^\text{teach}_{\mathcal{M}(i)})};\quad
	\lambda_i = \frac{\sum_{r,c \in M^{F}_i} 
        \llbracket \rho_{i, r,c} > \tau_1 \rrbracket }
        {|M^{F}_i|}
\end{align}
Note that $M^{F}_i$ represents the foreground locations for mask i,
$\llbracket . \rrbracket$ --- Iverson indicator function, 
and $\tau_1$ a threshold. 
The pixel-level confidence $\rho$ is defined in~(\ref{eq:m2f_inference}).
The mask-wide teacher confidence $\lambda_i$ 
corresponds to the ratio 
of the foreground pixels 
where the pixel-level confidence 
is larger than the threshold. 

\subsection{Confidence-based Point Filtering}
\label{sec:cbpf}
Training dense prediction models
on high-resolution images
can be extremely memory intensive.
This problem can be alleviated by training
on a carefully chosen 
sample of $N_p$ dense predictions
instead of on 
the whole prediction tensor~\cite{kirillov2020pointrend}.
The procedure starts by sampling
a random oversized set 
of $3N_p$ floating-point locations.
We subsample the 
initial oversized set
by choosing 
$\beta \cdot N_p$ points  
with the largest uncertainty ($\beta \in [0, 1]$),
and random $(1-\beta) \cdot N_p$ points.
However, the original procedure
is inappropriate for
consistency training.
In fact, the teacher and the student
will often be uncertain
at the same locations
since they 
are presented 
with the same image 
(up to a perturbation).
Thus, blind favoring
of uncertain student points
would increase the chance
of sampling incorrect 
pseudo-labels.
On the other hand,
favouring highly confident
points
would impair the
learning process
by providing
uninformative gradients.
Thus, we propose to 
favour points with 
low student 
and high teacher confidence 
by means of
a dense sampling
affinity $A_i$:
\begin{align}
    \label{eq:unc}
    \text{A}_i\left(r,c\right) = 
    \begin{cases}
               -\infty & : {\Phi}^\text{teach}_{r,c} < \tau_2\\
       -|s_{i,r,c}| & : \text{otherwise}
    \end{cases}.
\end{align}
Note that $\Phi^\text{teach}_{r,c}$ 
denotes the teacher confidence
that will be defined 
before the end of this subsection,
while $s_{i,r,c}$ denotes the pre-activation 
of the student mask-assignment $\sigma$.
Thus, if the teacher confidence 
in some point (r,c)
is lower than the threshold $\tau_2$,
we prevent its sampling by setting 
the sampling affinity to $-\infty$.
If the teacher is confident,
we set the sampling affinity
as the negative absolute value
of the student pixel-to-mask pre-activation.
We estimate the teacher 
confidence as follows:
\begin{align}
    \label{eq:conf_def}
    {\Phi}^\text{teach}_{r,c} = \max_i \rho^\text{teach}_{i,r,c}
\end{align}
This score assigns the 
lowest confidence
to the locations
that are not 
claimed by any of the masks,
and to the locations
claimed by masks
with low classification confidence.
This formulation of point sampling
is conservative as it prevents
any mask from training on 
low confidence teacher predictions.

\begin{figure}[h]
    \centering
    \includegraphics[width=\textwidth]{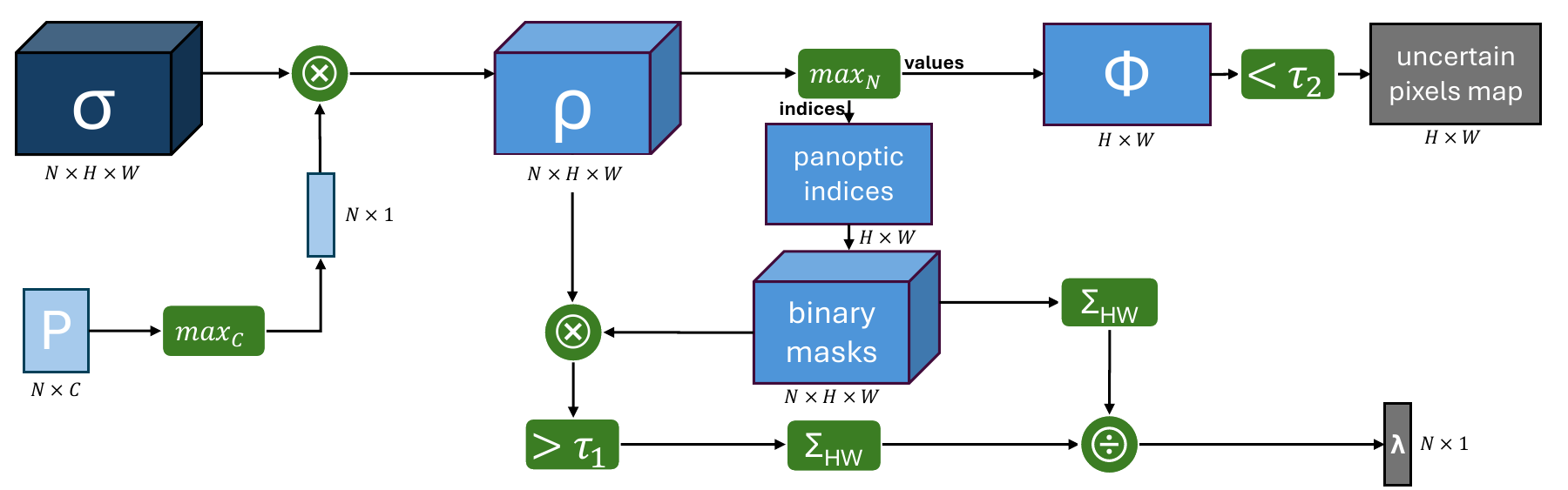}
    \caption{The teacher multiplies
the dense assignment masks $\sigma$
with mask-wide max-softmax
in order to recover
dense per-mask confidence $\rho$.
Taking the max along the mask axis of $\rho$
and thresholding with respect to $\tau_2$
reveals binary map 
of uncertain pixels~(\ref{eq:unc}).
Furthermore, the mask axis arg-max of $\rho$
delivers the map of panoptic indices,
which we further convert
to per-mask binary maps
by converting indices
to their one-hot encodings.
Finally, we determine 
per-mask teacher confidence  
$\boldsymbol{\lambda} = \left\{ \lambda_i \right\}$~(\ref{eq:lambda})
as the relative count
of dense per-mask confidences
that are greater than $\tau_1$.}
    \label{fig:mcpanda_comp_graph}
\end{figure}

Figure~\ref{fig:mcpanda_comp_graph}
illustrates the teacher's
computational graph
given the pixel-to-mask
assignment scores $\sigma$
and classification probabilities $P$.
We observe that 
the proposed additions 
to the baseline consistency
represent small
computational overhead
over the standard 
panoptic inference.

\section{Experiments}
\label{sec:exp}
\subsection{Implementation Details}
\textbf{Datasets:}
    We experiment on two real-world datasets:
    Cityscapes~\cite{Cordts_2016_CVPR} and
    Mapillary Vistas~\cite{Neuhold_2017_ICCV}.
    Cityscapes comprises 2975 training and 500
    validation images of European urban scenes 
    in fair weather and $1024\times2048$ resolution.
    Vistas comprises 18,000 training and 2000
    validation images of worldwide scenes
    in varying weather conditions
    and mean resolution of 8.4MPix. 
    For synthetic data,
    we consider Synthia~\cite{Ros_2016_CVPR}
    with 9400 images at $1280\times760$
    resolution, and Foggy Cityscapes~\cite{sakaridis2018semantic} with 2975 training
    and 500 validation images of $1024\times 2048$ pixels, 
    and attenuation factor $\beta=0.02$.\\
\textbf{Architecture:} 
    Our experiments involve panoptic 
    Mask2Former~\cite{cheng2022masked} 
    with Swin-B~\cite{liu2021swin} 
    backbone, and multi-scale deformable
    attention~\cite{zhu2020deformable}
    in the pixel decoder.
    We use the MiT-B5 backbone~\cite{NEURIPS2021_segformer} in comparisons with EDAPS~\cite{Saha_2023_ICCV}.\\
\textbf{Training:}
    We conduct the training
    in two stages~\cite{berrada2024guided}.
    The first stage initializes the backbone with ImageNet weights
    and pre-trains the teacher
    on 
    the source domain.
    The second stage 
    performs the
    domain adaptation training
    on unlabeled target domain  
    and labeled source domain.
    We pre-train the teacher on
    Synthia/Cityscapes for 20k/40k
    iterations and freeze the teacher
    for initial 15k/20k iterations
    of the second stage~\cite{berrada2024guided}.
    Subsequently,
    we set the teacher weights
    to the exponential moving average~\cite{tarvainen2017mean} 
    of the student
    with the decay factor $\alpha = 0.999$.
    We train
    for 90k iterations
    on batches of 
    two source domain
    and two target domain images.
    We oversample
    the rare classes 
    in the source domain~\cite{wu2019detectron2},
    and apply random scaling,
    horizontal flipping,
    color augmentation, 
    and random
    $512 \times 1024$ 
    cropping in both domains.
    We apply additional strong augmentations
    to the student image
    as described in section~\ref{sec:uda_basics}.
    We recover target 
    domain pseudo-labels
    through default 
    panoptic inference~\cite{cheng2022masked} 
    with the teacher,
    and train the student with 
    consistency and fine-grained confidence.
    We set the initial
    learning rate to 0.0001, weight
    decay to 0.05, and use AdamW~\cite{loshchilov2018decoupled}. 
    We set
    $\tau_1=0.99$ and $\tau_2=0.8$
    except on Synthia$\rightarrow$Vistas 
    where we report experiments
    for three values 
    of ($\tau_1$, $\tau_2$).
    All experiments
    are conducted 
    on a single A100-40GB.
\\
\textbf{Evaluation:}
We evaluate our models according to
panoptic quality (PQ)~\cite{kirillov2019panoptic}
that can be factored into segmentation quality (SQ) 
and recognition quality (RQ).
We report PQ for each category,
as well as the mean PQ.
Synthia~\cite{Ros_2016_CVPR} comprises
16 annotated classes that correspond
to a subset of the standard Cityscapes taxonomy.
Consequently, the experiments on Synthia
report $\text{PQ}_{16}$ as the mean
over the 16 Synthia classes.

\subsection{Comparison with the State-of-the-Art}
Table~\ref{tab:synthetic_to_real_mean} 
compares MC-PanDA
with the domain-adaptive panoptic 
segmentation methods from the literature
in two setups:
Synthia$\rightarrow$Cityscapes (left)
and Synthia$\rightarrow$Vistas (right).
Our method performs well
across all metrics in both setups.
Our Cityscapes performance (47.4 \pqs)
outperforms the previous state of the art
for 6.2 percentage points.
On Synthia$\rightarrow$Vistas,
our model achieves 37.9 \pqs
with the default hyperparameters.
We had noticed a deterioration 
of validation performance 
for a single training run.
Hence, we have performed
another experiment with a stricter 
loss subsampling threshold 
$\tau_2$ = 0.9,
which resulted in 38.7 \pqs.
We have also tried to account
for greater variety of stuff classes
by setting $\tau_1^{\text{stuff}}=0.9$.
This experiment yielded 39.6 \pqs.
In the end, we have included the median 
of the three compound assays in the table.
Note that all reported performances
are averaged over three training runs.
\begin{table}[htb]
\caption{Performance evaluation on Synthia$\rightarrow$Cityscapes and Synthia$\rightarrow$Vistas. Two-branch UniDAFormer~\cite{zhang23cvpr}~(UniDAF-PSN) works better 
than the mask transformer counterpart~(UniDAF-DETR). All our experiments are averaged over 3 random seeds.}
    \label{tab:synthetic_to_real_mean}
    \centering
    \footnotesize
    \begin{tabular}{lccc@{\quad}ccc}
        \toprule 
        & \multicolumn{3}{c}{Synthia$\rightarrow$City} & \multicolumn{3}{c}{Synthia$\rightarrow$Vistas} \\
        Method & $\text{SQ}_{16}$ & $\text{RQ}_{16}$ & $\text{PQ}_{16}$ & $\text{SQ}_{16}$ & $\text{RQ}_{16}$ & $\text{PQ}_{16}$ \\
        \midrule
        CVRN~\cite{huang2021cross}              & 66.6 & 40.9 & 32.1 & 65.3 & 28.1 & 21.3  \\
        UniDAF-DETR~\cite{zhang23cvpr}    & 64.7 & 42.2 & 33.0 & \no & \no & \no \\
        UniDAF-PSN~\cite{zhang23cvpr,kirillov2019panoptic}   & 66.9 & 44.3 & 34.2 & \no & \no & \no \\
        EDAPS~\cite{Saha_2023_ICCV}             & {72.7} & {53.6} & {41.2} & {71.7} & {46.1} & {36.6}\\
        \midrule
        MC-PanDA (ours) & \textbf{76.7} & \textbf{59.3} & \textbf{47.4} & {71.0} & \textbf{49.8} & \textbf{38.7} \\ 
%        MC-PanDA (ours) & \textbf{76.7} & \textbf{59.3} & \textbf{47.4} & \textbf{74.5} & \textbf{50.7} & \textbf{39.6} \\ 
        \bottomrule
    \end{tabular}
\end{table}

Figure~\ref{fig:sota_comparison}
shows qualitative results
on two Cityscapes scenes (rows 1-2)
and a single Vistas scene (row 3).
The columns show
the input image,
ground truth,
predictions of the current SOTA~\cite{Saha_2023_ICCV},
and our predictions.
In the first scene,
both models
fail to recognize terrain 
due to the absence of this class
in the source Synthia dataset.
Besides that,
we observe that our model delivers much better recognition of the road surface.
Moreover, it 
correctly localizes 
and recognizes the bus
while EDAPS fails in both tasks.
In the second scene,
our model prevails on the 
traffic sign and rider.
In the third scene,
our method delivers
significantly better
segmentation of the car
and the traffic light.
Both models fail
on the crosswalk:
EDAPS classifies it as vegetation,  
while our model rejects prediction
in some pixels and predicts 
sidewalk in others.
\newcommand{\myw}{0.245\textwidth}
\begin{figure}[t]
    \centering
    \begin{tabular}{c@{\,}c@{\,}c@{\,}c}
        \footnotesize Image & \footnotesize GT & \footnotesize EDAPS~\cite{Saha_2023_ICCV} & \footnotesize MC-PanDA \\
         \includegraphics[width=\myw]{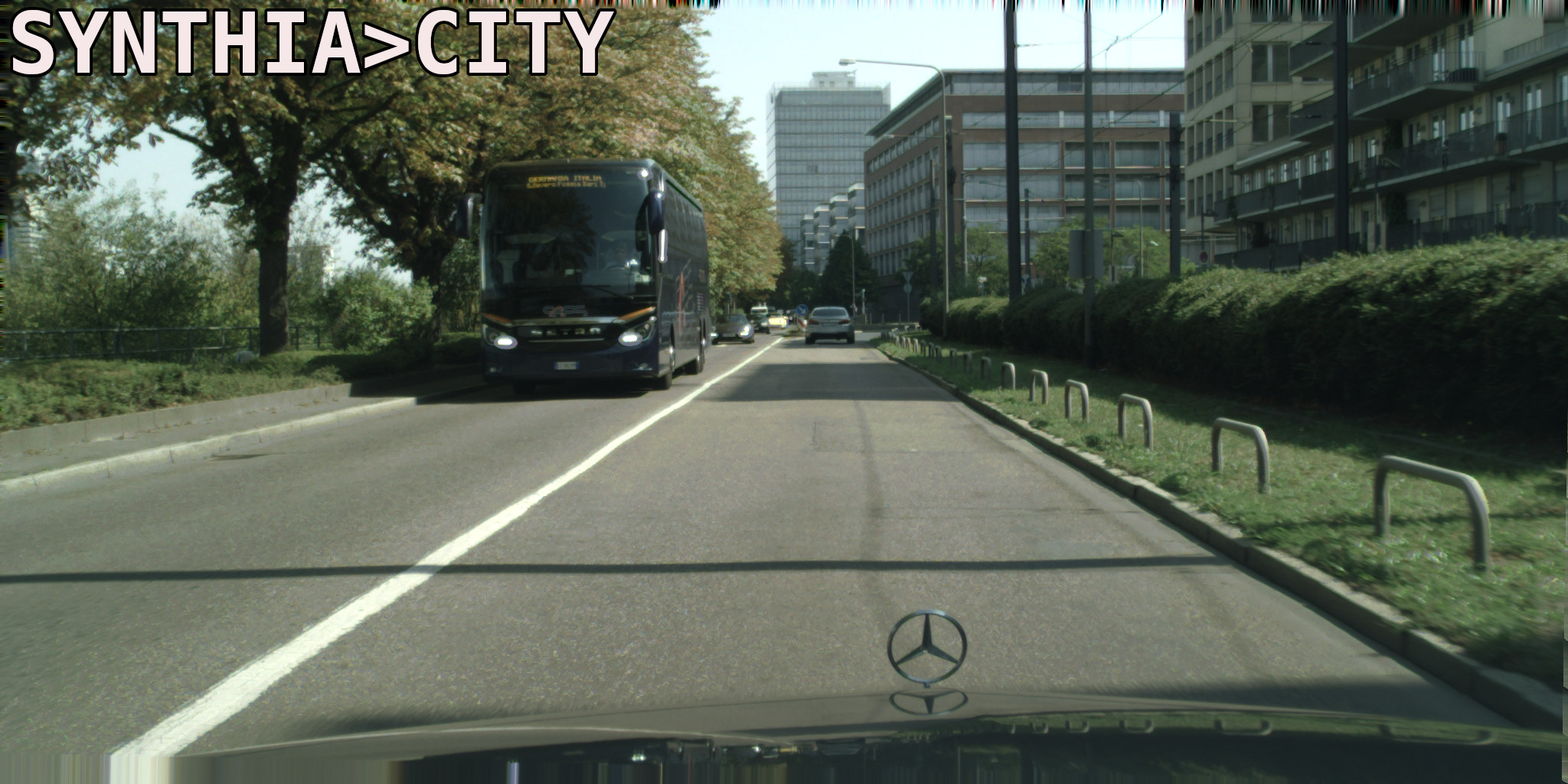} & 
         \includegraphics[width=\myw]{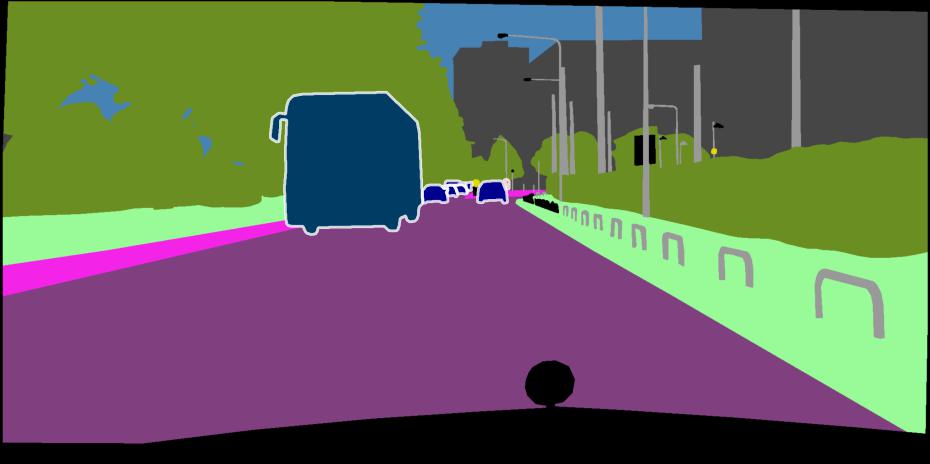} &
         \includegraphics[width=\myw]{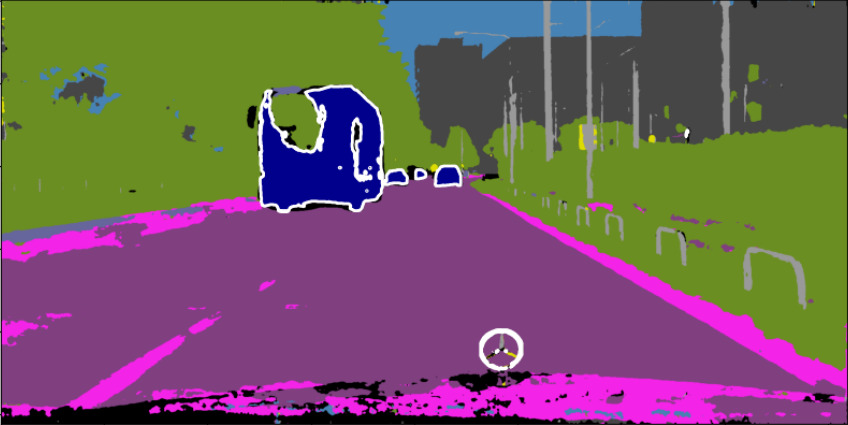} &
         \includegraphics[width=\myw]{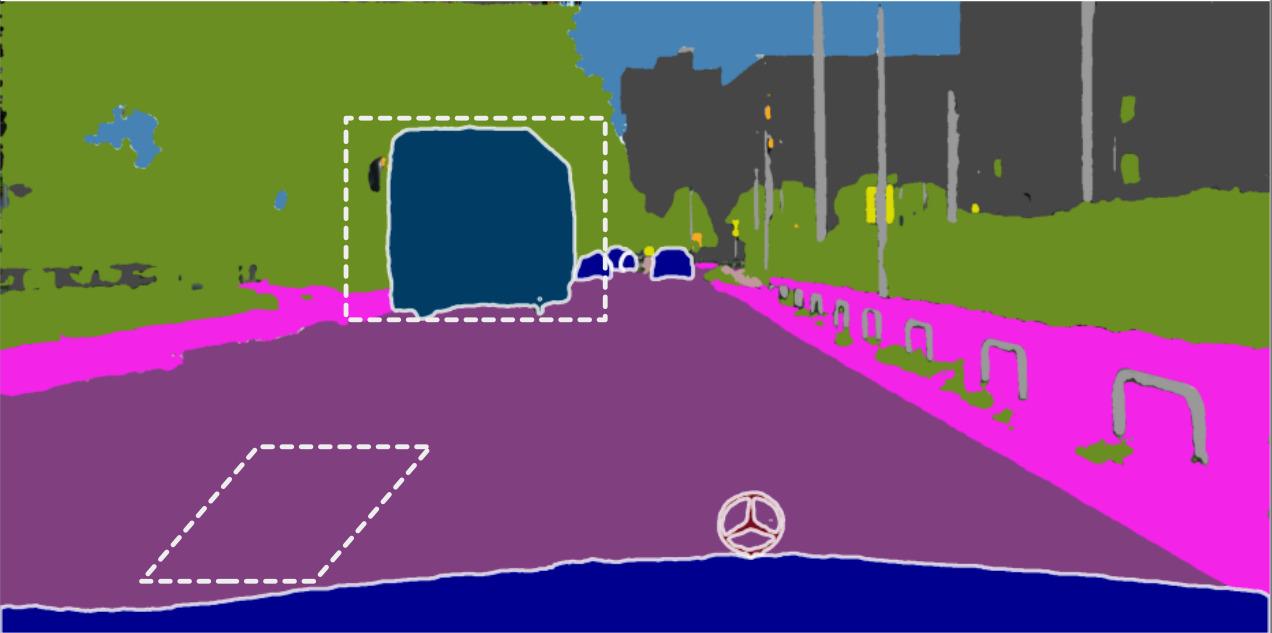} \\[-0.2em]

         \includegraphics[width=\myw]{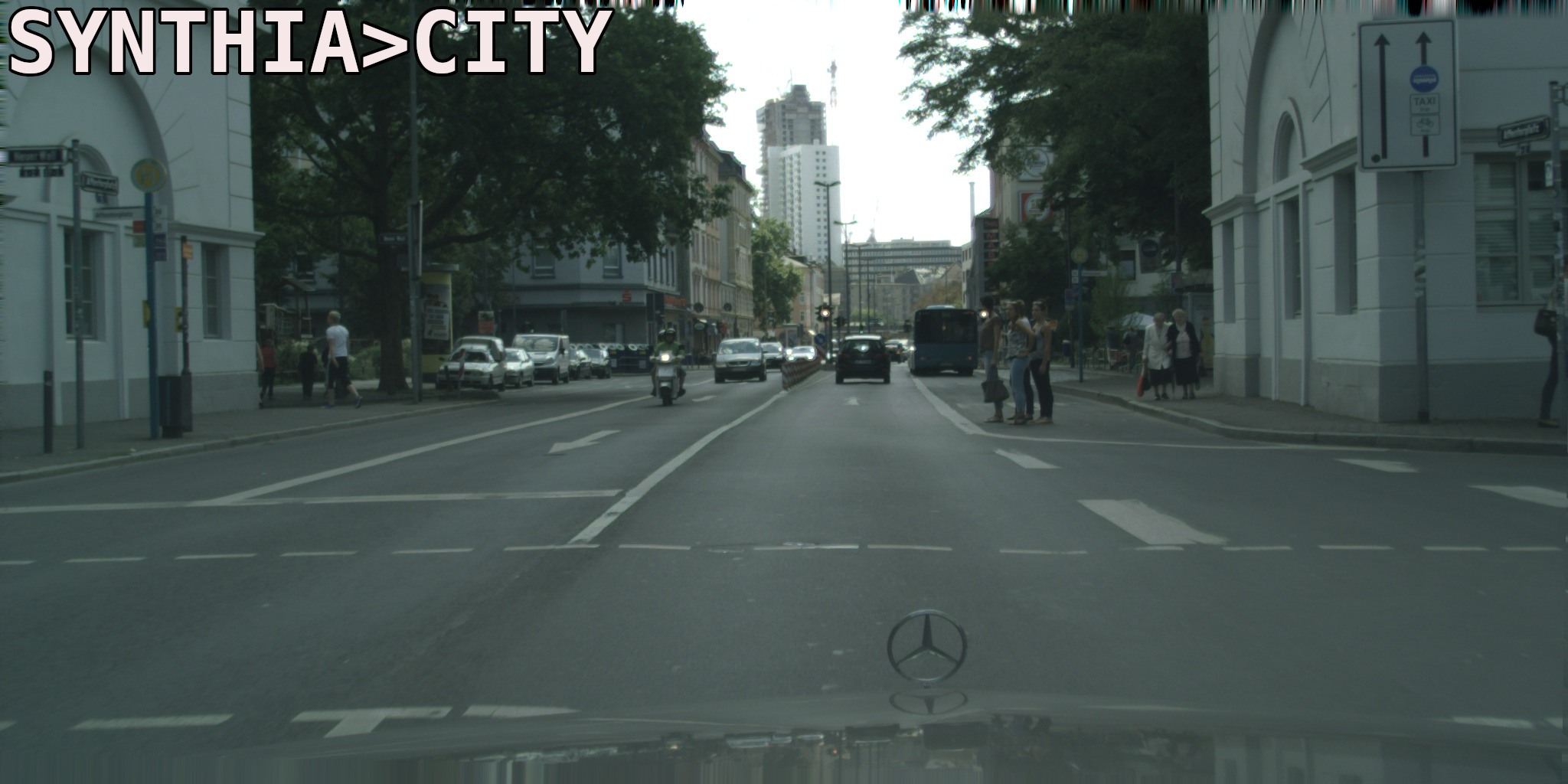} & 
         \includegraphics[width=\myw]{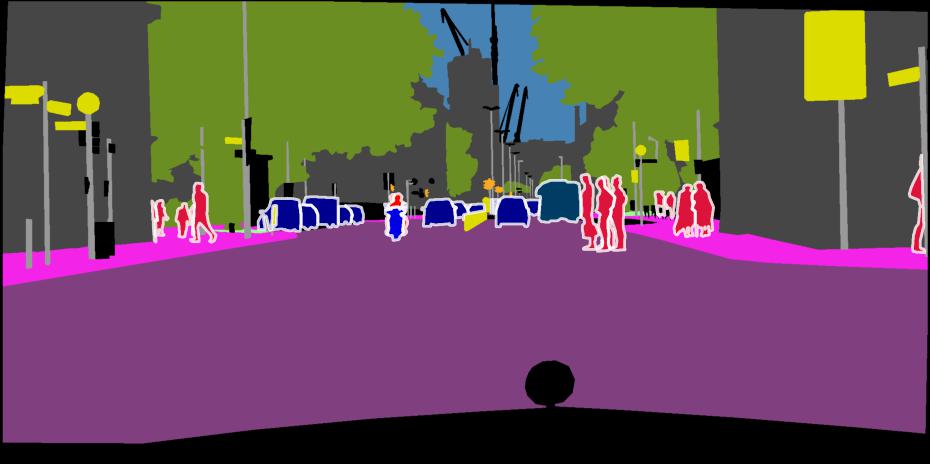} &
         \includegraphics[width=\myw]{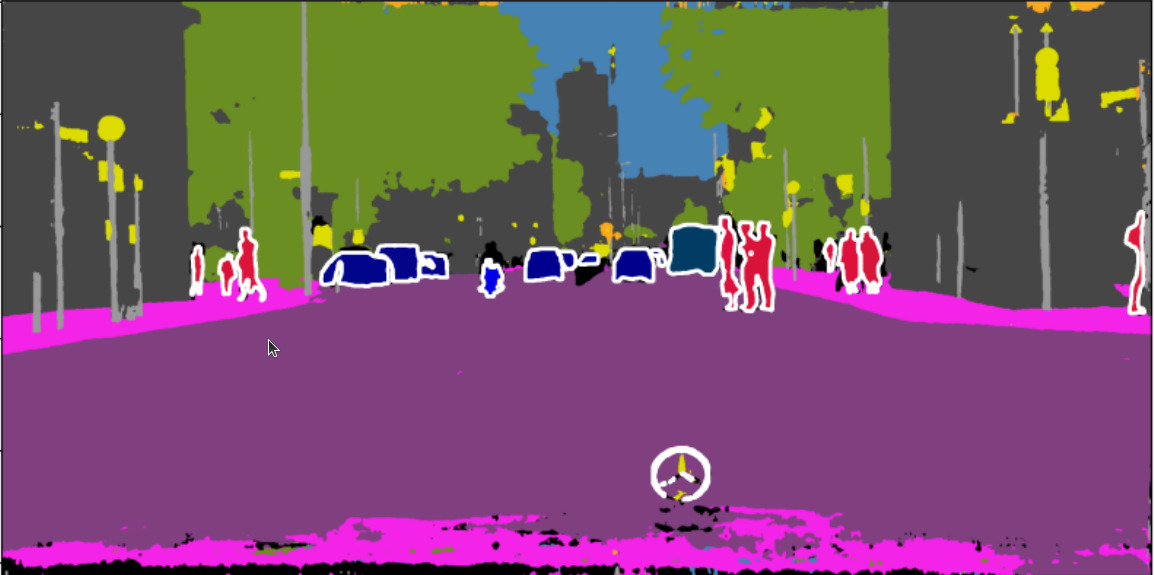} &
         \includegraphics[width=\myw]{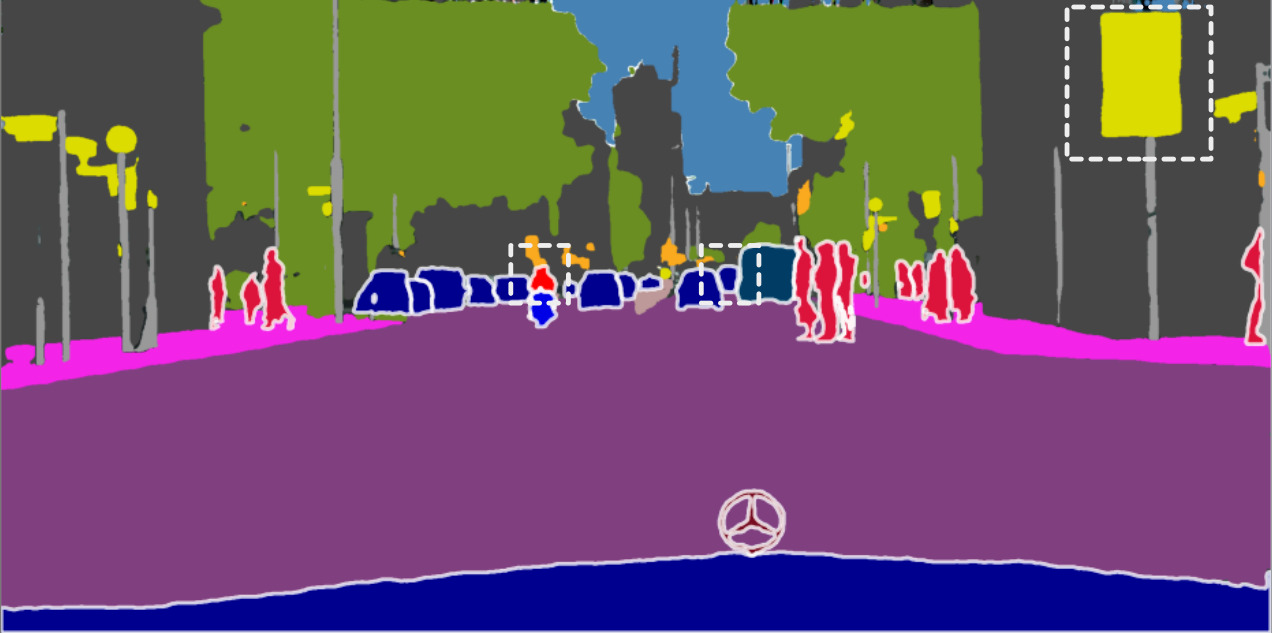} \\[-0.2em]
         \includegraphics[width=\myw]{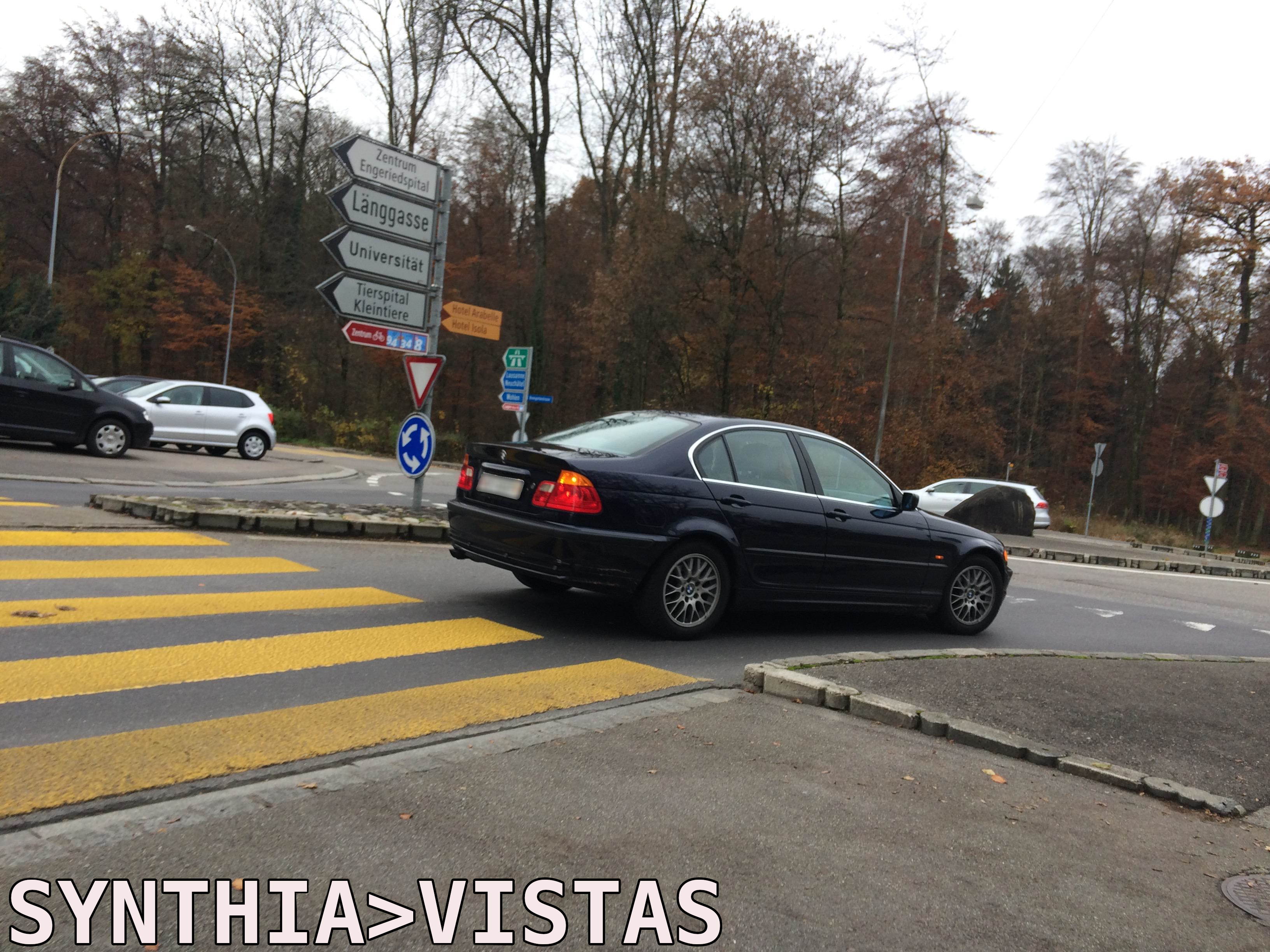} & 
         \includegraphics[width=\myw]{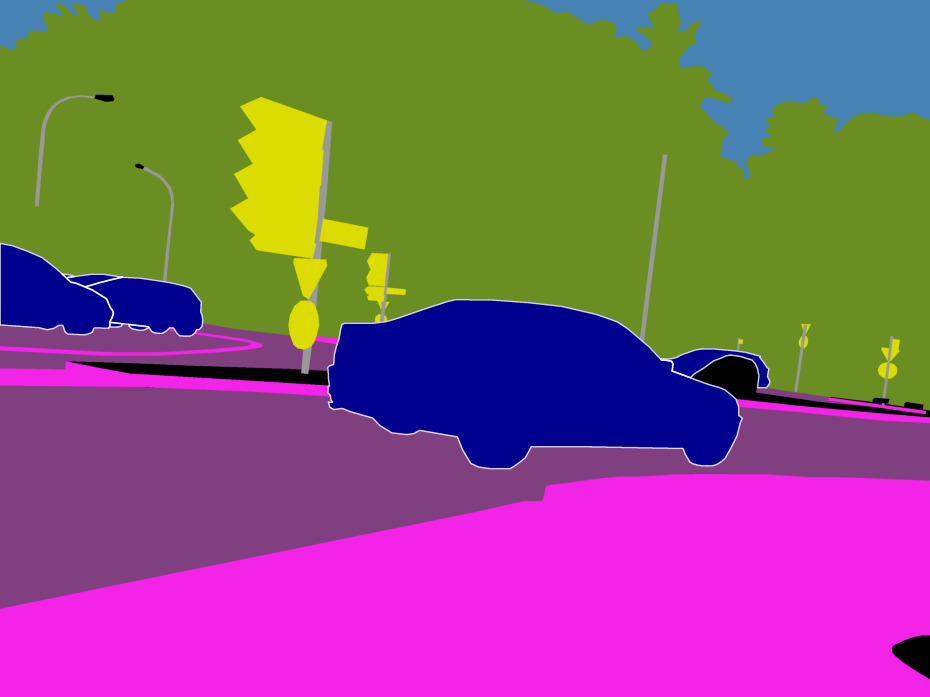} &
         \includegraphics[width=\myw]{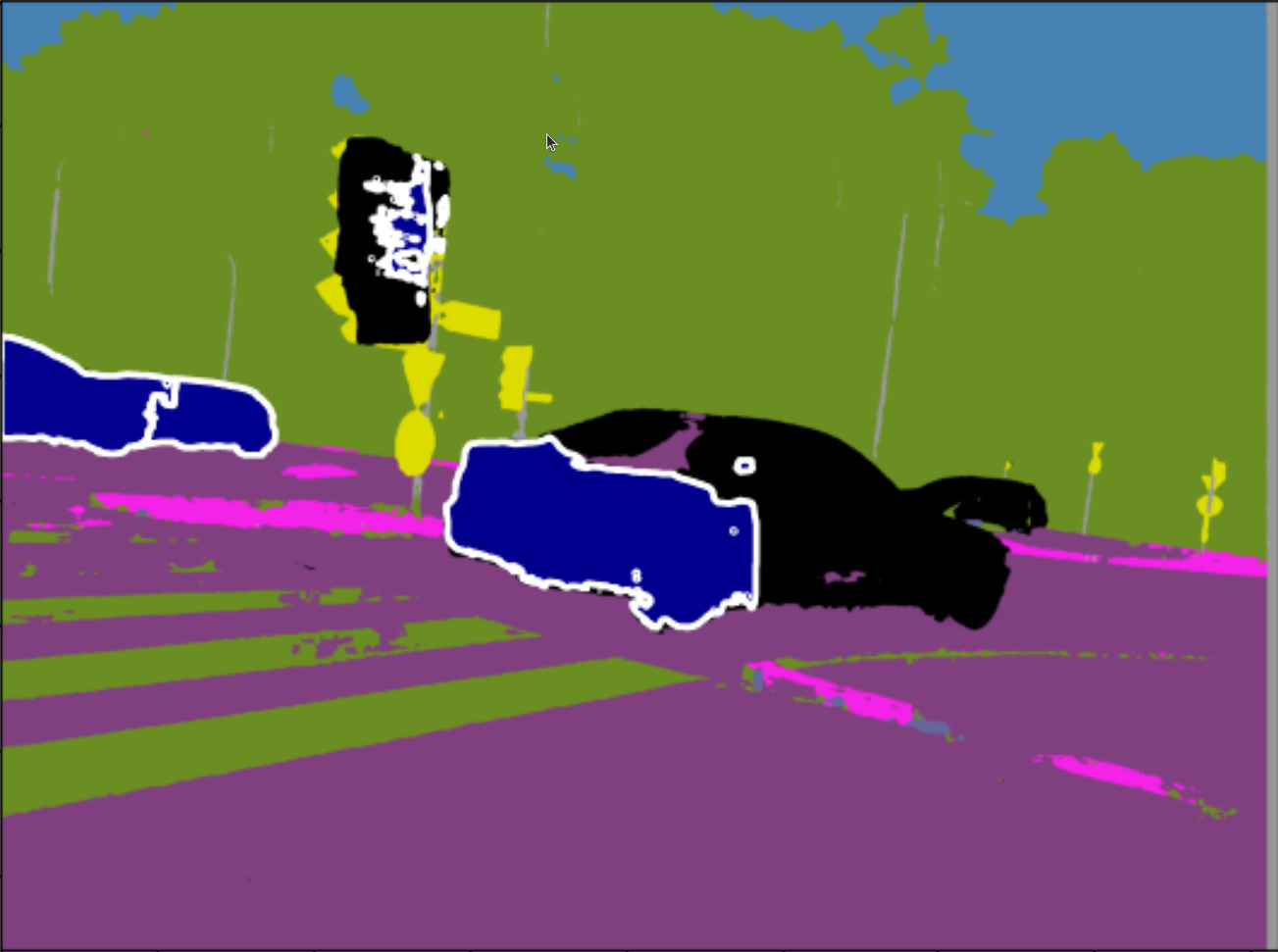} &
         \includegraphics[width=\myw]{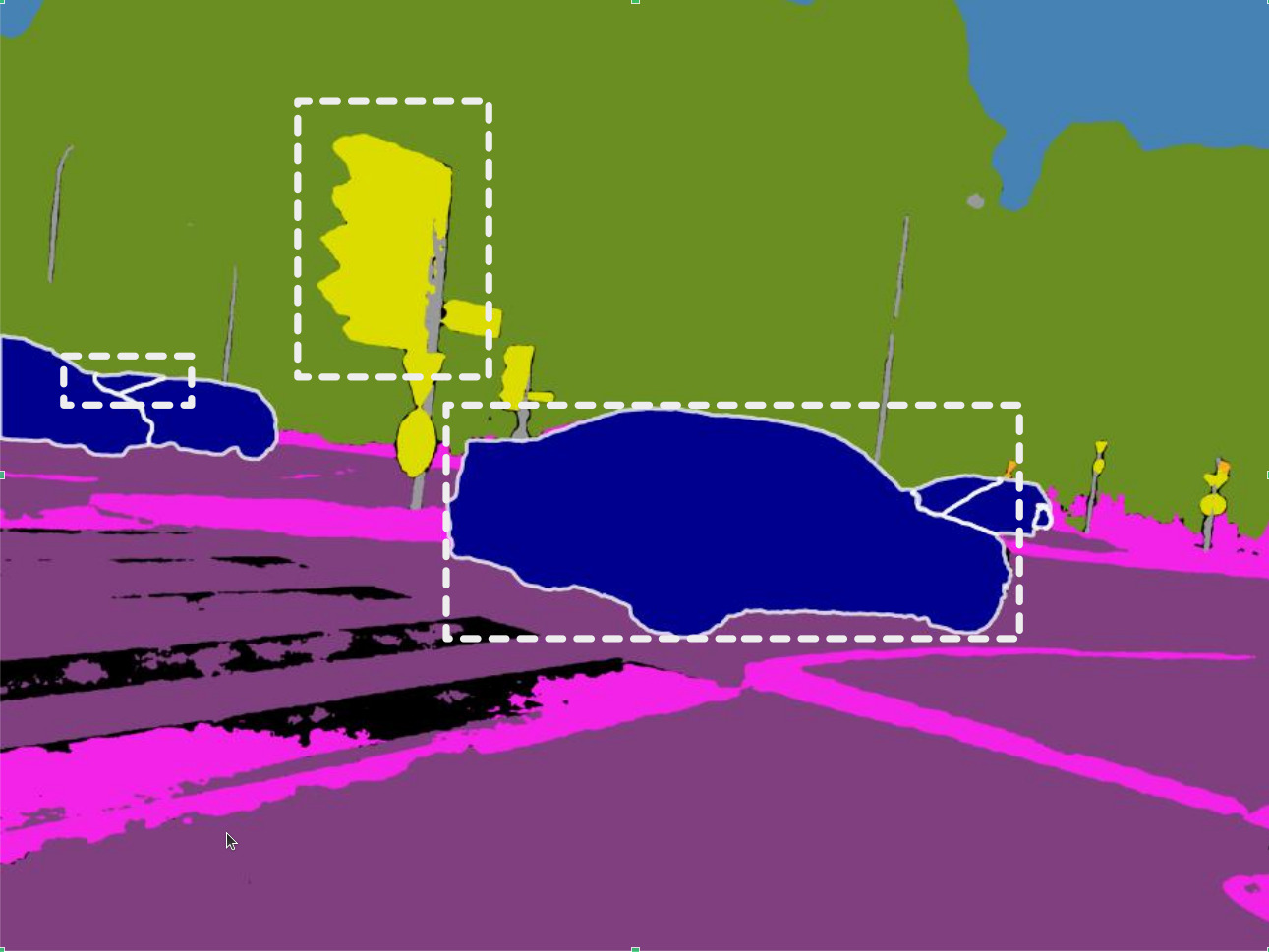}
    \end{tabular}
    \caption{
        Qualitative comparison with the state-of-the-art~\cite{Saha_2023_ICCV} on Synthia$\rightarrow$Cityscapes and Synthia$\rightarrow$Vistas. Dashed polygons indicate regions where our method prevails.}
    \label{fig:sota_comparison}
\end{figure}

\begin{table}[b]
    \caption{Performance evaluation on Cityscapes$\rightarrow$Foggy Cityscapes and Cityscapes $\rightarrow$ Vistas. All our experiments are averaged over three random seeds.}
    \label{tab:city_to_other_mean}
    \centering
    \footnotesize
    \begin{tabular}{lcccc@{\quad}cccc}
        \toprule 
        & \multicolumn{4}{c}{City$\rightarrow$Foggy} & \multicolumn{4}{c}{City$\rightarrow$Vistas} \\
        Method & $\text{SQ}_{16}$ & $\text{RQ}_{16}$ & $\text{PQ}_{16}$ & $\text{PQ}_{19}$ & $\text{SQ}_{16}$ & $\text{RQ}_{16}$ & $\text{PQ}_{16}$& $\text{PQ}_{19}$  \\
        \midrule
        CVRN \cite{huang2021cross} & 72.7 & 46.7 & 35.7 & \no &  73.8 & 42.8 & 33.5 & \no \\
        UniDAF \cite{zhang23cvpr} &  72.9 & 49.5 & 37.6 & \no & \no & \no & \no & \no \\
        EDAPS \cite{Saha_2023_ICCV} & {79.2} & {70.5} & {56.7} & \no & {75.9} & {53.4} & {41.2} & \no \\
        \midrule
        MC-PanDA & \textbf{82.5}  & \textbf{76.3}  & \textbf{63.7}  & 62.0 & \textbf{79.3} & \textbf{66.6} & \textbf{53.8} & 51.7  \\ 
        % prikazuje mPQ_16, mozda ce trebati prikazati mPQ 19 u ovom slucaju?
        \bottomrule
    \end{tabular}
\end{table}
Table~\ref{tab:city_to_other_mean}
compares MC-PanDA
with the methods from the literature
when we use Cityscapes 
as the source dataset,
while Foggy Cityscapes
and Vistas
serve as targets.
Note that
in the latter case,
both source and target domain
are based on real images.
Following the previous work,
we report $\text{PQ}_{16}$ 
averaged over 16 Synthia classes,
as well as $\text{PQ}_{19}$ 
averaged over 19 Cityscapes classes.
Our models
outperform previous
work 
by a large margin
and set a new 
state of the art.
We observe 
a substantial performance improvement
on Vistas in comparison 
to the Synthia experiments.
This reveals 
a significant space for improvement
in bridging the domain shift
between the synthetic
and real domains.

Table~\ref{tab:per_class_results}
complements the previous two tables
with per-class evaluations
of panoptic quality.
In experiments with Synthia
as the source domain,
our method achieves the
best performance
on most classes
and competitive peformance
elsewhere.
We observe that fence 
represents the hardest class
in these setups,
while sky and road
are the easiest.
When we use Cityscapes
as the source dataset,
our method prevails on all classes,
except on rider
when evaluating
on Foggy Cityscapes.
We observe significantly
higher panoptic quality
for class fence
than in domain adaptation from Synthia.
This suggests a large 
domain shift between fences
in Synthia and the two real datasets.
\begin{table}[h!]
    \caption{Per-class performance evaluation 
    on four domain adaptation setups.
    All our experiments are averaged over three random seeds.}
    \label{tab:per_class_results}
    \centering
    %\tiny
    \fontsize{6pt}{7pt}\selectfont
    \setlength{\tabcolsep}{1.2pt}
    \renewcommand{\arraystretch}{1}
    \newcommand{\wall}{\includegraphics[width=1em]{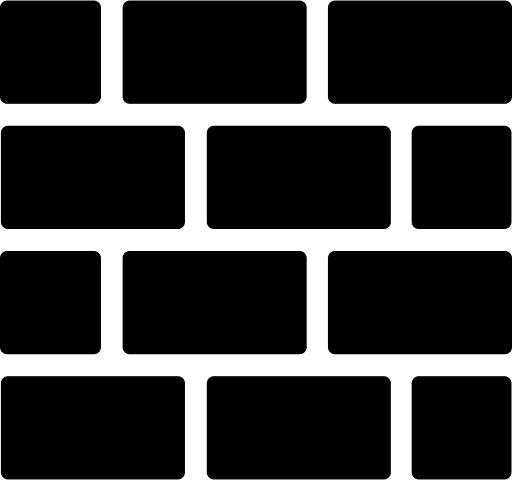}}
    \newcommand{\fence}{\includegraphics[width=1em]{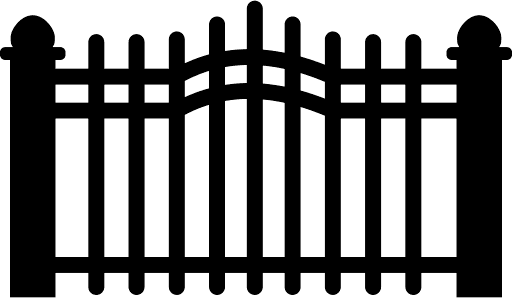}}
    \newcommand{\pole}{\includegraphics[width=0.6em]{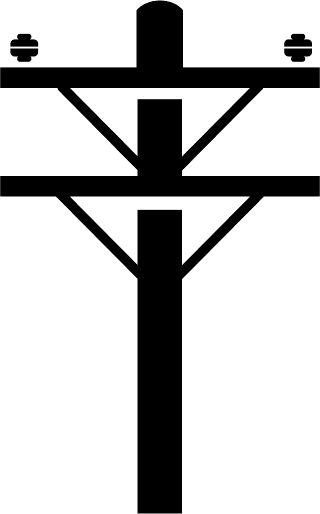}}
    \newcommand{\trsign}{\includegraphics[width=1em]{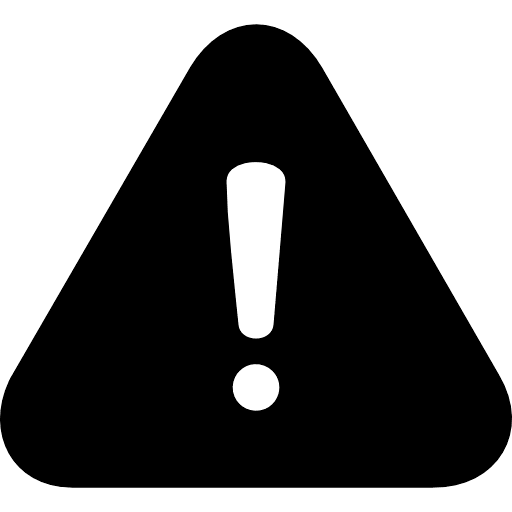}}
    \newcommand{\mybf}[1]{\textbf{#1}}
    {
    \begin{tabular}{lcccccccccccccccc@{\hskip 0.15cm}c}
        \toprule 
         & \faRoad & \faShoePrints & \faBuilding & \wall & \fence & \pole & \faTrafficLight & \trsign  & \faTree & \faCloud & \faMale & \faBiking & \faCar & \faBus&  \faMotorcycle & \faBicycle & {$\text{PQ}_{16}$} \\
        % & \rots{road} & \rots{sidewalk} & \rots{building} & \rots{wall} & \rots{fence} & \rots{pole} & \rots{tr.light} & \rots{tr.sign} & \rots{veg} & \rots{sky} & \rots{person} & \rots{rider} & \rots{car} & \rots{bus} & \rots{m.bike} & \rots{bike} \\
        \midrule
        Method & \multicolumn{17}{c}{Synthia$\rightarrow$Cityscapes}\\
        \midrule
        % FDA \cite{yang2020fda}                  & 79.0      & 22.0      & 61.8      & 1.1  & 0.0 & 5.6  & 5.5   & 9.5  & 51.6 & 70.7 & 23.4 & 16.3 & 34.1 & 31.0 & 5.2  & 8.8  \\
        % CRST \cite{zou2019confidence}           & 75.4      & 19.0      & 70.8      & 1.4  & 0.0 & 7.3  & 0.0   & 5.2  & 74.1 & 69.2 & 23.7 & 19.9 & 33.4 & 26.6 & 2.4  & 4.8  \\
        % AdvEnt \cite{vu2019advent}              & 87.1      & 32.4      & 69.7      & 1.1  & 0.0 & 3.8  & 0.7   & 2.3  & 71.7 & 72.0 & 28.2 & 17.7 & 31.0 & 21.1 & 6.3  & 4.9  \\
        CVRN \cite{huang2021cross}              & 86.6  & 33.8      & 74.6      & 3.4  & 0.0 & 10.0 & 5.7   & 13.5 & 80.3 & 76.3 & 26.0 & 18.0 & 34.1 & 37.4 & 7.3  & 6.2 & {32.1} \\
        UniDAF \cite{zhang23cvpr}    & 73.7      & 26.5      & 71.9      & 1.0 & 0.0 & 7.6 & 9.9 & 12.4 & 81.4 & 77.4 & 27.4 & 23.1 & {47.0} & {40.9} & 12.6 & 15.4 & {33.0}\\
        UniDAF$^\dagger$ \cite{zhang23cvpr} & \mybf{87.7} & 34.0 & 73.2 & 1.3 & 0.0 & 8.1 & 9.9 & 6.7 & 78.2 & 74.0 & 37.6 & 25.3 & 40.7 & 37.4 & 15.0 & {18.8} & {34.2}\\
        EDAPS \cite{Saha_2023_ICCV}             & 77.5      & {36.9}  & {80.1} & \mybf{17.2} & \mybf{1.8} & {29.2} & \mybf{33.5} & {40.9} & {82.6} & {80.4} & {43.5} & {33.8} & 45.6 & 35.6 & {18.0} & 2.8 & {41.2}\\
        [0.2em]
        MC-PanDA & {87.2}     & \mybf{51.8}  & \mybf{82.5} & 16.1 & {1.7} & \mybf{36.3} & {26.1} & \mybf{54.3} & \mybf{86.3} & \mybf{86.4} & \mybf{48.3} & \mybf{37.7} & \mybf{46.9} & \mybf{45.8} & \mybf{27.4} & \mybf{23.9} & \mybf{47.4}\\
        \toprule
         & \multicolumn{17}{c}{Synthia$\rightarrow$Vistas}\\
        \midrule
        
        % FDA \cite{yang2020fda}          & 44.1 & 7.1 & 26.6 & 1.3 & 0.0 & 3.2 & 0.2 & 5.5 & 45.2 & 61.3 & 30.1 & 13.9 & 39.4 & 12.1 & 8.5 & 7.0  \\
        % CRST \cite{zou2019confidence}   & 36.0 & 6.4 & 29.1 & 0.2 & 0.0 & 2.8 & 0.5 & 4.6 & 47.7 & 68.9 & 28.3 & 13.0 & 42.4 & 13.6 & 5.1 & 2.0  \\
        % AdvEnt \cite{vu2019advent}      & 27.7 & 6.1 & 28.1 & 0.3 & 0.0 & 3.4 & 1.6 & 5.2 & 48.1 & 66.5 & 28.4 & 13.4 & 40.5 & 14.6 & 5.2 & 3.3  \\
        CVRN \cite{huang2021cross}      & 33.4 & 7.4 & 32.9 & 1.6 & 0.0 & 4.3 & 0.4 & 6.5 & 50.8 & 76.8 & 30.6 & 15.2 & 44.8 & 18.8 & 7.9 & {9.5} & 21.3 \\
        EDAPS \cite{Saha_2023_ICCV}      & {77.5} & {25.3} & {59.9} & \mybf{14.9} & 0.0 & {27.5} & \text{33.1} & {37.1} & \mybf{72.6} & \mybf{92.2} & {32.9} & \mybf{16.4} & {47.5} & \mybf{31.4} & {13.9} & 3.7 & 36.6 \\[0.2em]
%        MC-PanDA & \mybf{82.4} & \text{22.2} &\mybf{65.3} &\text{13.2} &\mybf{0.1} &\mybf{39.3} &\text{30.6} &\mybf{51.9} &\text{70.7} &\text{89.8} &\mybf{40.9} &\text{10.5} &\mybf{50.6} &\text{23.4} &\mybf{23.2} &\mybf{20.4} \\
        MC-PanDA & \mybf{82.7} & \mybf{26.5} &\mybf{61.0} &\text{5.5} &\text{0.0} &\mybf{39.9} &\mybf{34.4} &\mybf{51.3} &\text{62.1} &\text{85.6} &\mybf{41.9} &\text{11.4} &\mybf{50.6} &\text{25.1} &\mybf{23.0} &\mybf{18.1} & \mybf{38.7} \\
        \toprule
         & \multicolumn{17}{c}{Cityscapes$\rightarrow$Foggy Cityscapes}\\
        \midrule
        
        % DAF & 94.0 & 54.5 & 57.7 & 6.7  & 10.0 & 7.0 & 6.6  & 25.5  & 44.6  & 59.1  & 26.7  & 16.7  & 42.2  & 36.6 & 4.5  & 16.9  \\
        % AdvEnt \cite{vu2019advent}           & 93.8  & 52.7  & 56.3  &5.7 &13.5  & 10.0  &10.9  & 27.7  & 40.7  & 57.9  & 27.8  &29.4  & 44.7 & 28.6 & 11.6 & 20.8   \\
        % CRST \cite{zou2019confidence}           &91.8 & 49.7 & 66.1 & 6.4 &14.5 &5.2 &8.6& 21.5& 56.3 &50.7& 30.5& 30.7& 46.3& 34.2& 11.7 &22.1    \\
        CVRN \cite{huang2021cross}          
        &93.6&52.3&65.3&7.5&15.9&5.2&7.4&22.3&57.8&48.7&32.9&30.9&49.6&38.9&18.0&25.2 & 35.7 \\
        UniDAF \cite{zhang23cvpr} 
        &93.9&53.1&63.9&8.7&14.0&3.8&10.0&26.0&53.5&49.6&38.0&35.4&57.5&44.2&28.9&29.8& 37.6  \\
        EDAPS \cite{Saha_2023_ICCV} 
        & {91.0} & {68.5} & {80.9} & {24.1} & 29.0 & {50.1} & {47.2} & {67.0} & {85.3} & {71.8} & {50.9} & \mybf{51.2} & {64.7} & {47.7} & {36.9} & 41.5 & 56.7 \\[0.2em]
        MC-PanDA & \mybf{98.0} & \mybf{80.6} & \mybf{85.8} & \mybf{45.7} & \mybf{43.4} & \mybf{60.3} & \mybf{49.4} & \mybf{73.8} & \mybf{87.9} & \mybf{81.7} & \mybf{53.5} & 47.8 & \mybf{65.2} & \mybf{61.6} & \mybf{40.2} & \mybf{44.3} & \mybf{63.7}\\ 
        \toprule
         & \multicolumn{17}{c}{Cityscapes$\rightarrow$Vistas}\\
        \midrule
        CVRN \cite{huang2021cross} & 77.3& 21.0& 47.8& 10.5& 13.4& 7.5& 14.1& 25.1& 62.1& 86.4& 37.7& 20.4& 55.0& 21.7& 14.3& 21.4 & 33.5 \\
        EDAPS \cite{Saha_2023_ICCV}      & {58.8} & {43.4} & {57.1} & {25.6} & 29.1 & {34.3} & {35.5} & {41.2} & {77.8} & {59.1} & {35.0} & {23.8} & {56.7} & {36.0} & {24.3} & 25.5 & 41.2  \\[0.2em]
        MC-PanDA & \mybf{88.4} & \mybf{49.1} & \mybf{75.2} & \mybf{35.2} & \mybf{39.7} & \mybf{50.3} & \mybf{45.2} & \mybf{54.0} & \mybf{81.1} & \mybf{96.2} & \mybf{46.1} & \mybf{30.1} & \mybf{57.2} & \mybf{42.0} & \mybf{33.9} & \mybf{37.3} & \mybf{53.8} \\ 
        \bottomrule
    \end{tabular}
    }
\end{table}

\subsection{Ablations and Validations}
This section validates the performance 
impact of our contributions on
Synthia $\rightarrow$ Cityscapes.
For all domain adaptation experiments
we  report mean and standard deviation
over three runs.
Please find 
additional ablations in the supplement,
including a detailed comparison
with the consistency baseline,
the impact of $\tau_1$ and $\tau_2$
on the final performance,
and ablations on Synthia$\rightarrow$Vistas.

Table~\ref{tab:main_ablation_2nd_version}
quantifies the contributions of
mask-wide loss scaling (MLS)
and confidence-based point filtering (CBPF)
to the panoptic performance of adapted models.
We report results 
for MLS with
two different thresholds:
$\tau_1=0.968$~\cite{tranheden2021dacs,Saha_2023_ICCV}
and $\tau_1=0.99$.
The first row shows the 
domain generalization performance
of a supervised model trained only on Synthia images.
We notice that our
consistency baseline from section~\ref{sec:uda_basics}
achieves the improvement
of  8.8 points.
Furthermore,
self-training with MLS
brings additional improvement
of around 5 points,
which is almost 14 PQ points
better than the  
domain generalization baseline.
Confidence-based point filtering (CBPF)
contributes a further improvement of around 3 PQ points.
Finally, our complete method
achieves $47.4$ PQ points
which represents
an improvement of 16.6 PQ points
w.r.t.~the supervised baseline 
and 7.8 PQ points
w.r.t.~baseline consistency training.
\begin{table}[h!]
\centering
\caption{Ablation study on Synthia$\rightarrow$Cityscapes: baseline consistency (BC), mask-level loss scaling (MLS), and confidence-based point filtering (CBPF). Top row represents supervised training on Synthia. We report mean$_{\pm\text{std}}$ over three random seeds.}
\label{tab:main_ablation_2nd_version}
\scriptsize
\setlength{\tabcolsep}{3pt}
\begin{tabular}{cccccccccc}
\toprule
BC & $\text{MLS}_{.968}$ & $\text{MLS}_{.99}$  & $\text{CBPF}_{0.8}$  & \multicolumn{2}{c}{$\text{SQ}_{16}$} & \multicolumn{2}{c}{$\text{RQ}_{16}$}& \multicolumn{2}{c}{$\text{PQ}_{16}$} \\
\midrule
\no & \no & \no & \no & 70.7 & \blarrow & 40.4 & \blarrow & 30.8 & \blarrow\\
\yes & \no & \no & \no &  \mstd{73.4}{0.6} & +2.7 & \mstd{50.0}{1.7} & +9.6 & \mstd{39.6}{1.5} & +8.8\\
\yes & \yes & \no & \no &  \mstd{75.1}{0.5} & +4.4 & \mstd{56.1}{1.8} & +15.7 & \mstd{44.4}{1.5} & +13.6\\
\yes & \no & \yes & \no &  \mstd{75.3}{0.2} & +4.6 & \mstd{56.5}{0.9} & +16.1 & \mstd{44.6}{0.7} & +13.8\\
\yes & \no & \no & \yes &  {\mstd{75.8}{0.4}} & {+5.1} & {\mstd{55.7}{0.2}} & {+15.3} & {\mstd{44.1}{0.2}} & {+13.3}\\
\yes & \yes & \no & \yes &  \mstd{76.5}{0.0} & +5.8 & \mstd{59.2}{1.0} & {+18.8} & \mstd{47.2}{0.5} & +16.4\\
\yes & \no & \yes & \yes & \textbf{\mstd{76.7}{0.4}} & \textbf{+6.0} & \textbf{\mstd{59.3}{1.0}} & \textbf{+18.9} & \textbf{\mstd{47.4}{0.8}} & \textbf{+16.6}\\
\bottomrule
\end{tabular}
\end{table}

\begin{figure}[h!]
    \renewcommand{\myw}{0.325\textwidth}
    \centering
    \begin{tabular}{c@{\,}c@{\,}c}
         \scriptsize 15k & 
         \scriptsize 40k & 
         \scriptsize 90k \\
         \includegraphics[width=\myw]{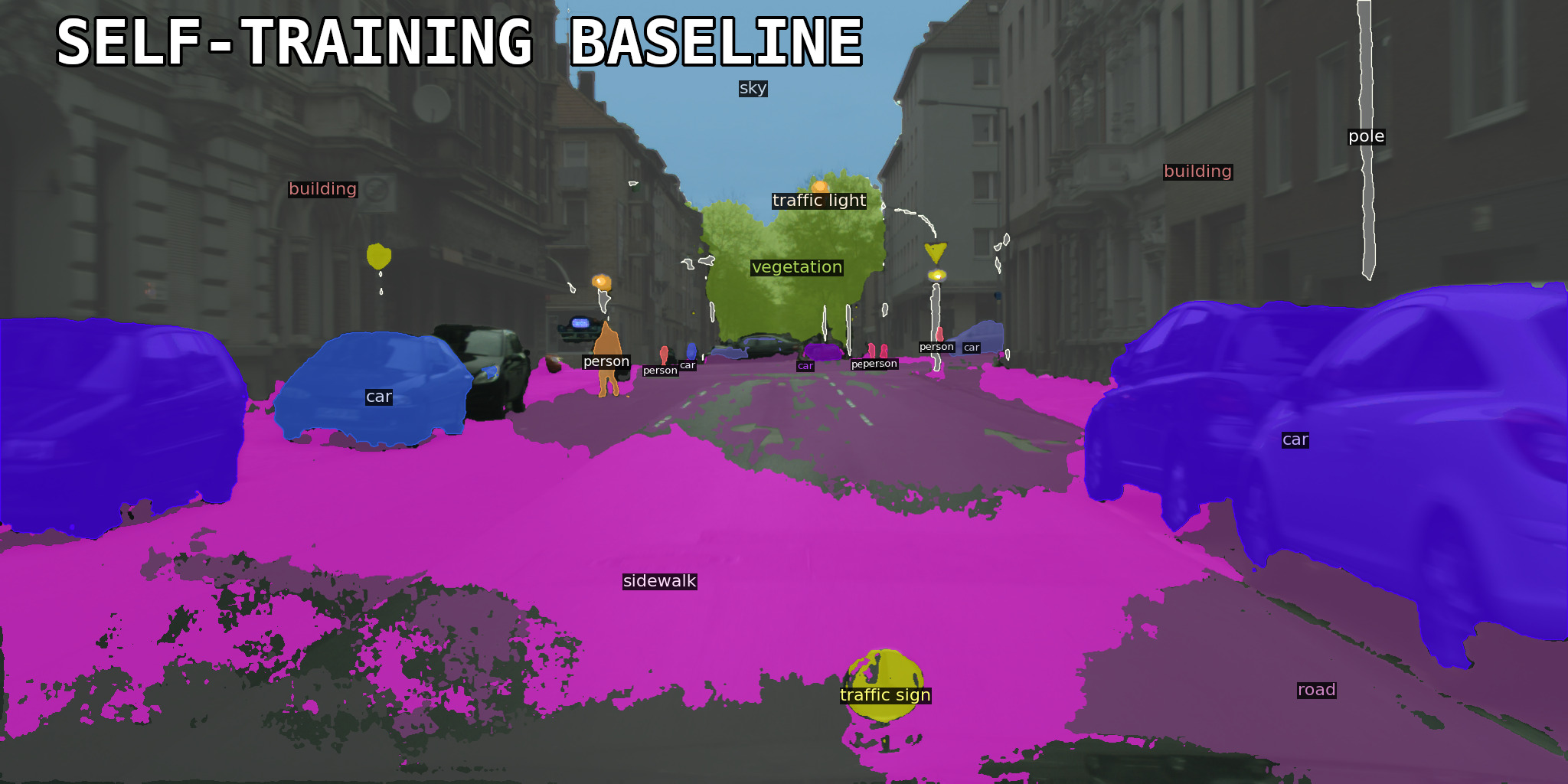} & 
         \includegraphics[width=\myw]{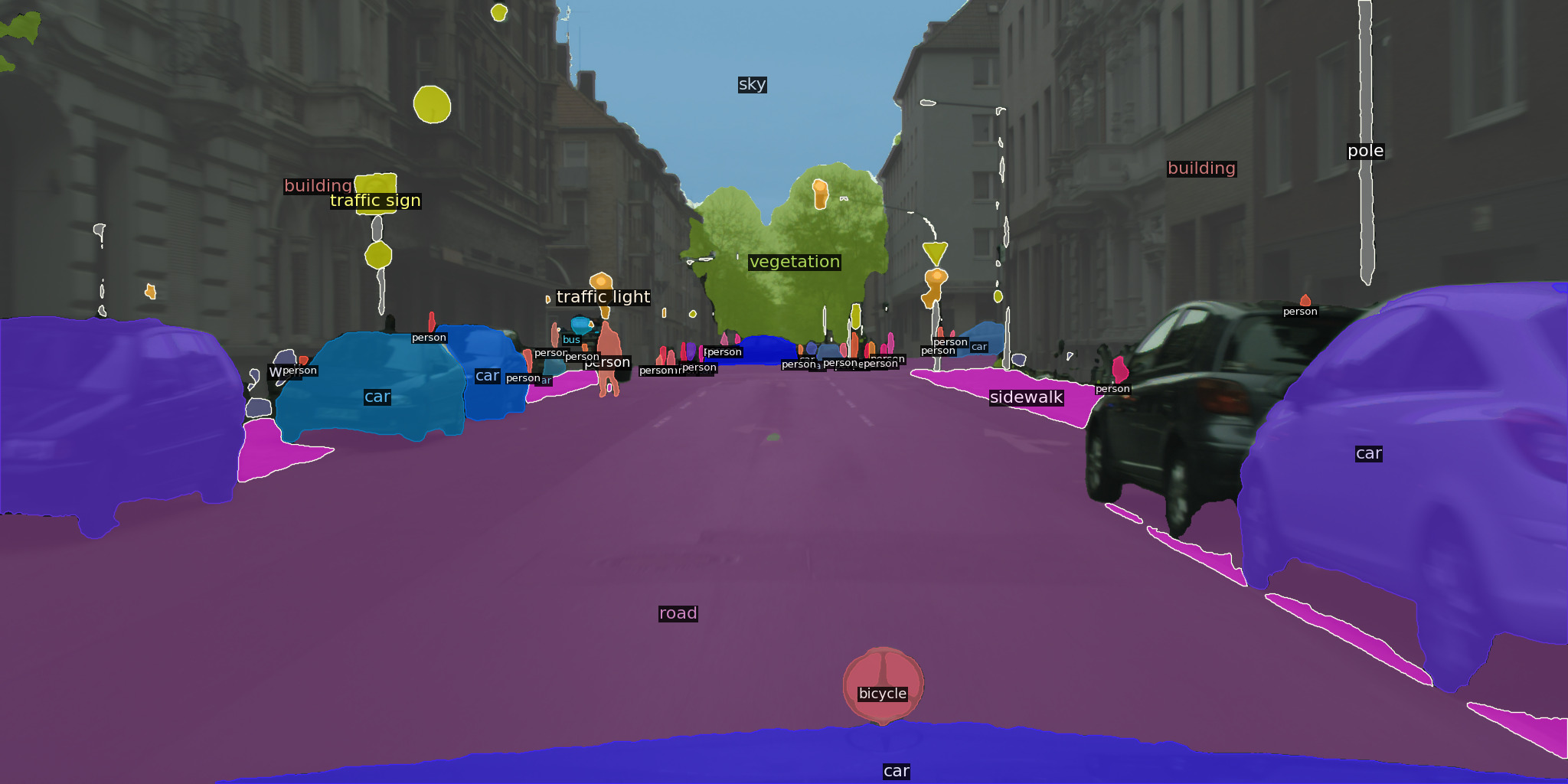} &
         \includegraphics[width=\myw]{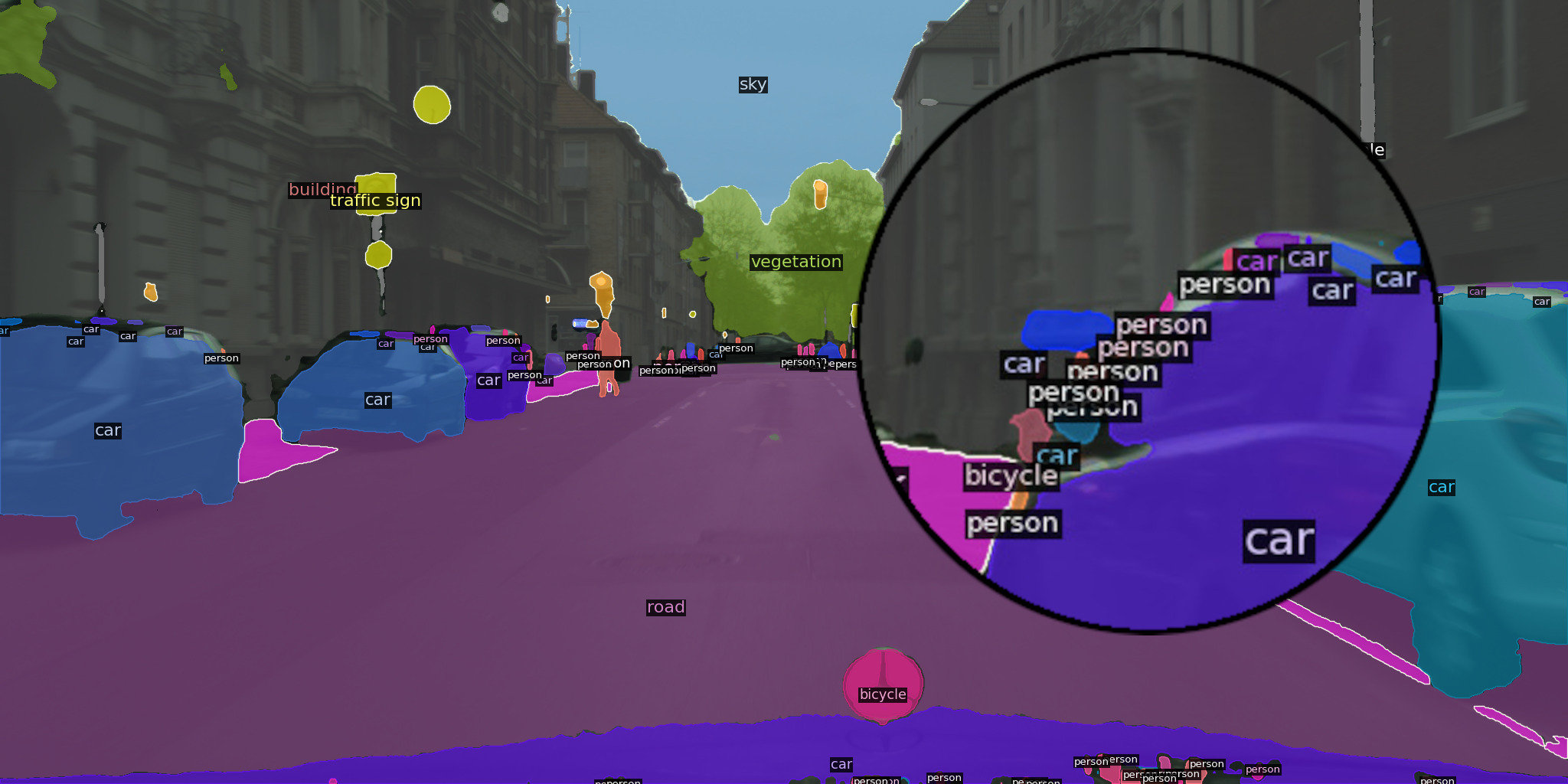} \\ [-0.2em]
         
         \includegraphics[width=\myw]{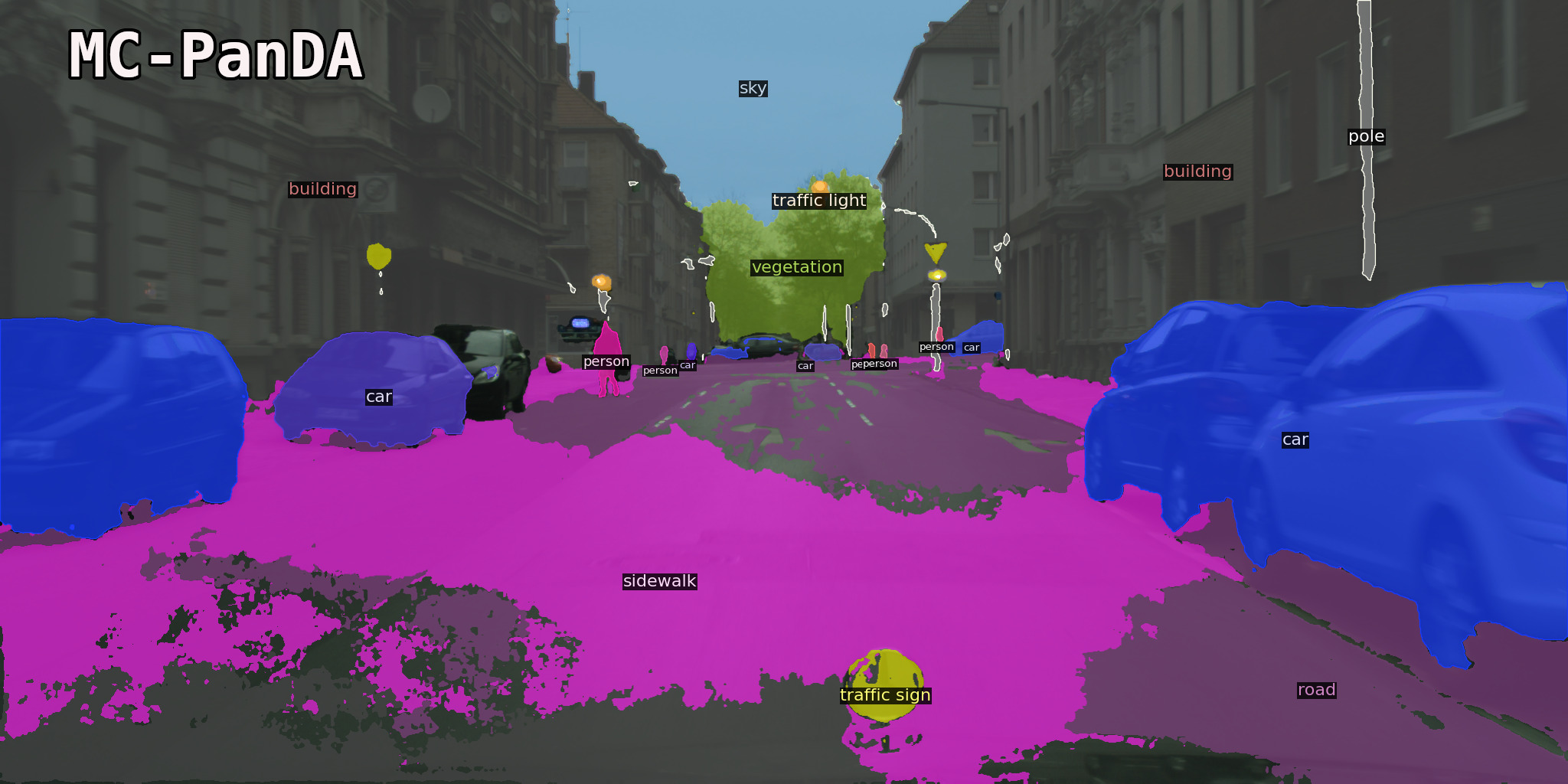} &
         \includegraphics[width=\myw]{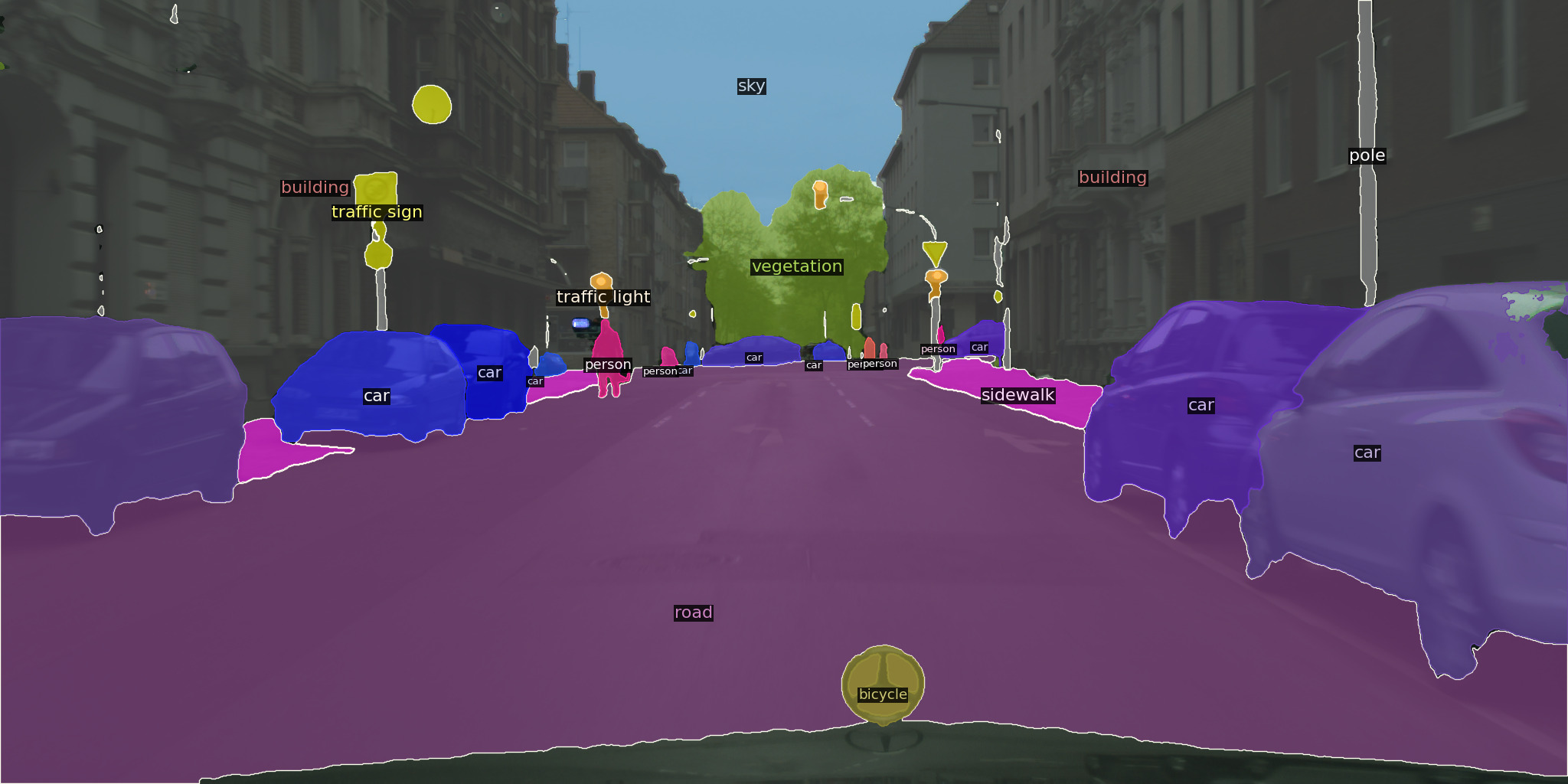} &
         \includegraphics[width=\myw]{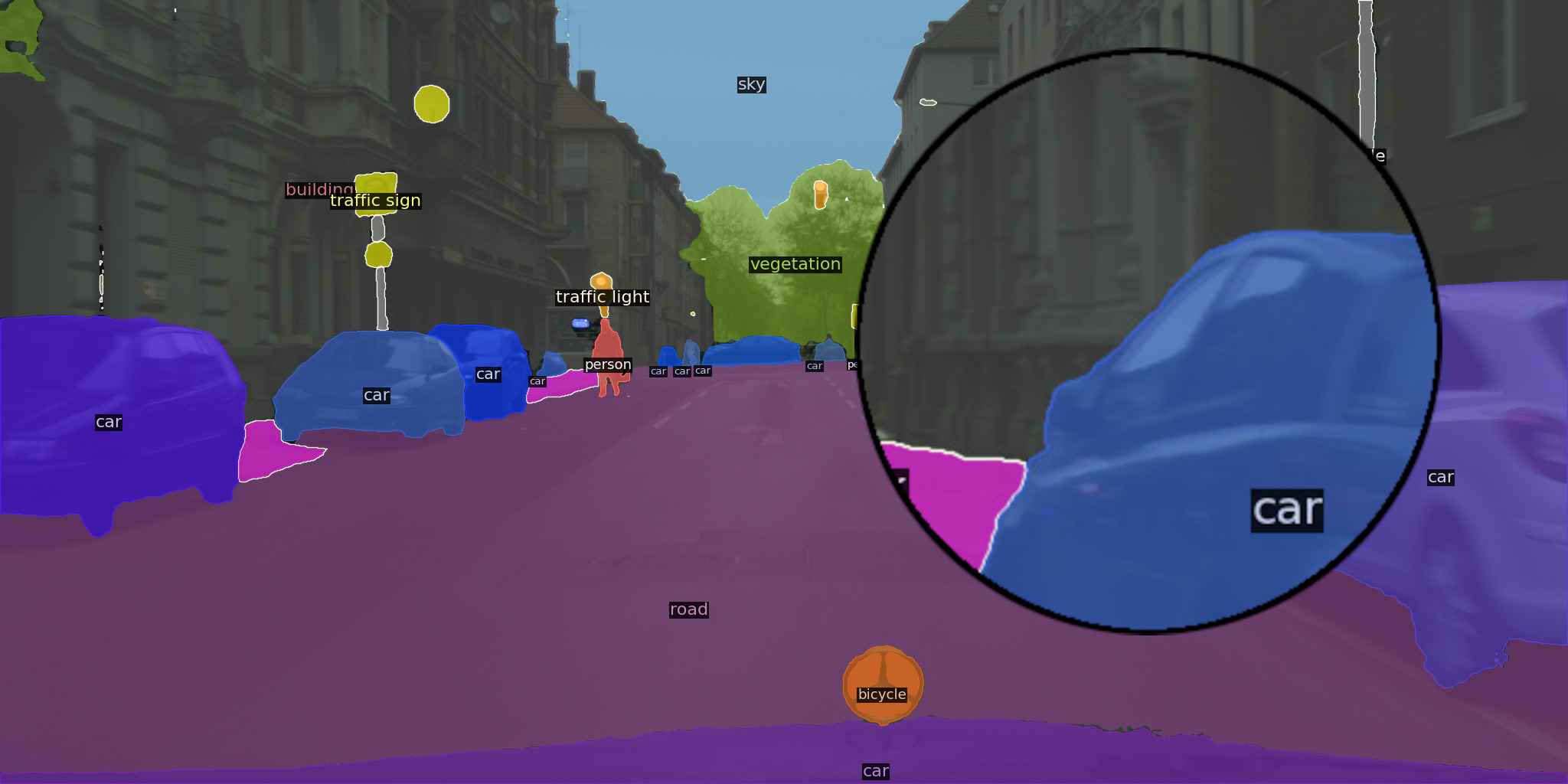} \\
    \end{tabular}
    \caption{Panoptic predictions in different training iterations. 
    Top row presents our consistency baseline, while the bottom row presents our approach. Best viewed zoomed in.}
    \label{fig:st_vs_ours}
\end{figure}

Figure~\ref{fig:st_vs_ours} illustrates
the difference in pseudo-label quality
of our consistency baseline (top)
and MC-PanDA (bottom)
at three training checkpoints.
We observe that 
the pseudo-label quality
improves as the training progresses.
Moreover, we notice that
the baseline
starts hallucinating 
false positive segments,
which significantly deteriorates
panoptic quality.
We believe this is due to the
learning on noisy pseudo-labels 
in a positive feedback loop.
On the other hand,
our method
reduces those hallucinations
by downscaling the gradients
of low-confidence masks and avoiding
pixels with uncertain mask assignment.
Note that both methods
suppress very small masks
during teacher inference.

We proceed by investigating
alternative formulations
of our proposed contributions.
Table~\ref{tab:confidence_calc_ablation} 
compares our mask-wide loss scaling (MLS)
with image-wide loss scaling (ILS)
that assigns each mask 
with the same 
image-wide scaling factor~\cite{Saha_2023_ICCV}.
We formalize the image-wide loss weight as follows: 
\begin{align}
    \lambda_i=\lambda^\text{ILS} = \frac{\sum_{r,c} \llbracket \Phi^\text{teach}_{r,c} > \tau_{\text{ILS}} \rrbracket}{HW}.
\end{align}
We experiment
with three different 
$\tau_\text{ILS}$ thresholds $\{0.95, 0.968, 0.99\}$
and report the 
performance of the best choice 
($\tau_\text{ILS}=0.968$)
as an optimistic baseline.
Top section shows results
of ILS and MLS 
without confidence-based 
point filtering (CBPF),
while the bottom section
includes CBPF.
We observe a clear advantage
of mask-wide loss weighting in both cases.
We believe this happens since
ILS down-scales the loss gradients for many
valid masks.
Table~\ref{tab:table_ubpf_variant_ablation}
validates the proposed loss 
subsampling according to CBPF.
The baseline point-sampling~\cite{kirillov2020pointrend}
favors points
with high student
uncertainty,
which increases
chances of sampling
incorrect pseudo labels.
CBPF
reduces this effect
by preventing training
in points with 
low teacher confidence
as presented in Figure~\ref{fig:mcpanda_comp_graph}.
We also experiment with
${\Phi}^\text{teach}_{i,r,c}=\sigma(|s^\text{teach}_i|)_{r,c}$
as an alternative formulation 
of the teacher confidence,
which considers only the mask 
that corresponds 
to the actual loss component.
This experiment also considers
completely random 
point samples 
(random sampling, $\beta=0.0$).
We observe that 
our formulation
(all-masks)
outperforms
the alternative (per-mask)
by 2.7 PQ points.
We believe that 
per-mask
underperformance
occurs due to 
over-confident
negative mask assignments where
many pixels 
get incorrectly predicted
as not belonging
to the corresponding mask.
\begin{table*}[h]
\parbox[t]{.48\linewidth}{
\centering
\caption{Comparison of image-wide (ILS) and mask-wide (MLS) loss scaling on Synthia$\rightarrow$Cityscapes. 
We report an optimistic ILS performance as a maximum across three different $\tau_\text{ILS}$ thresholds.}
\label{tab:confidence_calc_ablation}
\scriptsize
\begin{tabular}{cccccc}
\toprule
ILS & MLS & CBPF & SQ$_{16}$ & RQ$_{16}$ & PQ$_{16}$ \\
\midrule
\yes & \no & \no & \mstd{73.7}{0.5} & \mstd{50.5}{1.1} & \mstd{39.7}{0.8} \\
\no & \yes & \no &\mstd{75.3}{0.2} & \mstd{56.5}{1.0} & \mstd{44.7}{0.7} \\
\midrule
\yes & \no & \yes &\mstd{74.8}{0.4} & \mstd{53.4}{0.2} & \mstd{42.1}{0.5} \\
\no & \yes & \yes &\mstd{76.7}{0.4} & \mstd{59.3}{1.0} & \mstd{47.4}{0.8} \\
\bottomrule
\end{tabular}
}
\hfill
\parbox[t]{.48\linewidth}{
\centering
\caption{
    Validation of loss subsampling with
    confidence-based point filtering (CBPF)
    on Synthia$\rightarrow$Cityscapes.
The random sampling policy applies the loss
across a random set of image points.
}
\label{tab:table_ubpf_variant_ablation}
\scriptsize
\begin{tabular}{lccc}
\toprule
CBPF ($\tau_2=0.8)$ \\
filtering method & $\text{SQ}_{16}$ & $\text{RQ}_{16}$ & $\text{PQ}_{16}$ \\
\midrule
$\text{random sampling}$ & \mstd{74.9}{0.1} & \mstd{56.1}{1.4} & \mstd{44.2}{1.1} \\
$\text{per-mask}$ & \mstd{75.2}{0.4} & \mstd{56.5}{0.1} & \mstd{44.7}{0.1} \\
$\text{all-masks}~(\text{eq.}\ \ref{eq:conf_def})$ & \mstd{76.7}{0.4} & \mstd{59.3}{1.0} & \mstd{47.4}{0.8} \\
\bottomrule
\end{tabular}
}
\end{table*}

Table~\ref{tab:val_backbone}
compares our method 
with the current 
state-of-the-art \cite{Saha_2023_ICCV}
in experiments with matching backbones.
We retrain EDAPS with 
our Swin-B backbone \cite{liu2021swin}
and default hyperparameters,
as well as MC-PanDA
with the MiT-B5 backbone \cite{NEURIPS2021_segformer}.
We observe 3.5 PQ points advantage of our method
with MiT-B5 backbone,
and 8.1 PQ points with Swin-B.
Note that the number of parameters
of both models 
are roughly the same.
\begin{table}[t]
    \caption{
        Backbone validation on Synthia$\rightarrow$Cityscapes and comparison
        with EDAPS~\cite{Saha_2023_ICCV} using the same backbone. We report mean$_{\pm\text{std}}$ over three random seeds. $\dagger$ denotes our training with the publicly available source code.}
    \label{tab:val_backbone}
    \centering
    \begin{tabular}{l@{\quad}lcccc}
    \toprule
    Method & Backbone & \#Params & $\text{SQ}_{16}$ & $\text{RQ}_{16}$ & $\text{PQ}_{16}$ \\
    \midrule
    EDAPS \cite{Saha_2023_ICCV} & MiT-B5 & 104.9M & \mstd{72.7}{0.2} & \mstd{53.6}{0.5} & \mstd{41.2}{0.4}\\                        
    EDAPS$^{\dagger}$ & Swin-B & 110.8M & \mstd{73.1}{0.4} & \mstd{51.4}{1.6} & \mstd{39.3}{1.5}\\
    \midrule
    MC-PanDA & MiT-B5 & 103.2M & \mstd{75.1}{0.2} & \mstd{56.5}{0.3} & \mstd{44.7}{0.2} \\ 
    MC-PanDA & Swin-B & 107.0M & \mstd{76.7}{0.4} & \mstd{59.3}{1.0} & \mstd{47.4}{0.8} \\ 
    \bottomrule
    \end{tabular}
\end{table}

\subsection{Qualitative Analysis}
Figure~\ref{fig:mask_quality} illustrates advantages of our method on two qualitative examples.
The figure presents two masks at
three different training checkpoints
and the corresponding 
mask-wide loss-scaling factors $\lambda_i$.
We observe
a strong positive correlation
between the mask quality
and the scaling factor, 
which suggests
that our technique indeed
suppresses gradients
in low-quality masks.
This behavior is especially important
during early training when
many road and traffic-sign 
pixels get incorrectly excluded
from the corresponding masks.
\newcommand{\mywt}{0.33\textwidth}
\begin{figure}[h]
    \centering
    %\scriptsize
    \begin{tabular}{c@{\,}c@{\,}c}
    % \begin{tabular}{c@{\,}c@{\,}c}
         \tiny 15k & 
         \tiny20k & 
         \tiny40k \\
         \includegraphics[width=\mywt]{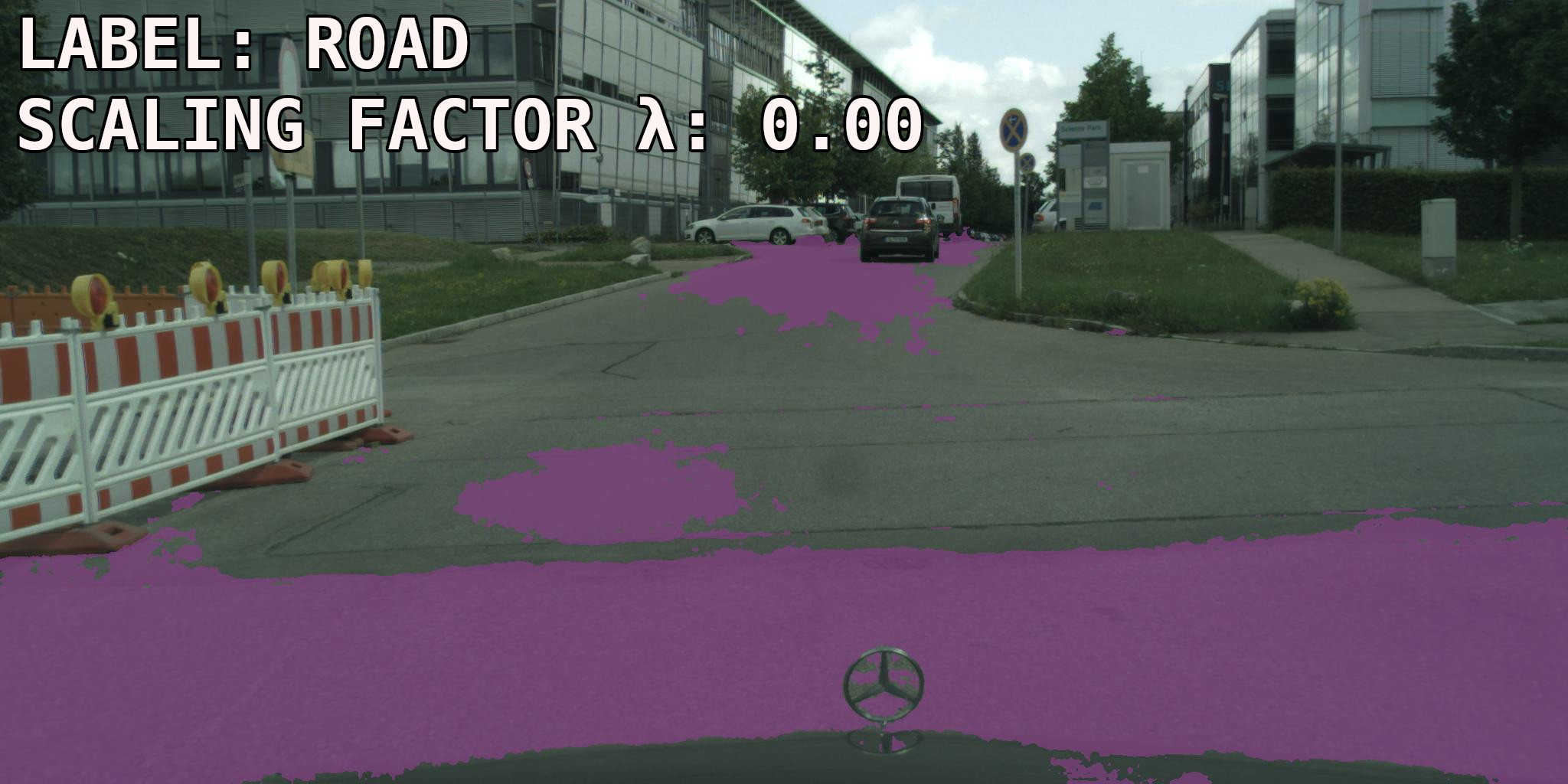} & 
         \includegraphics[width=\mywt]{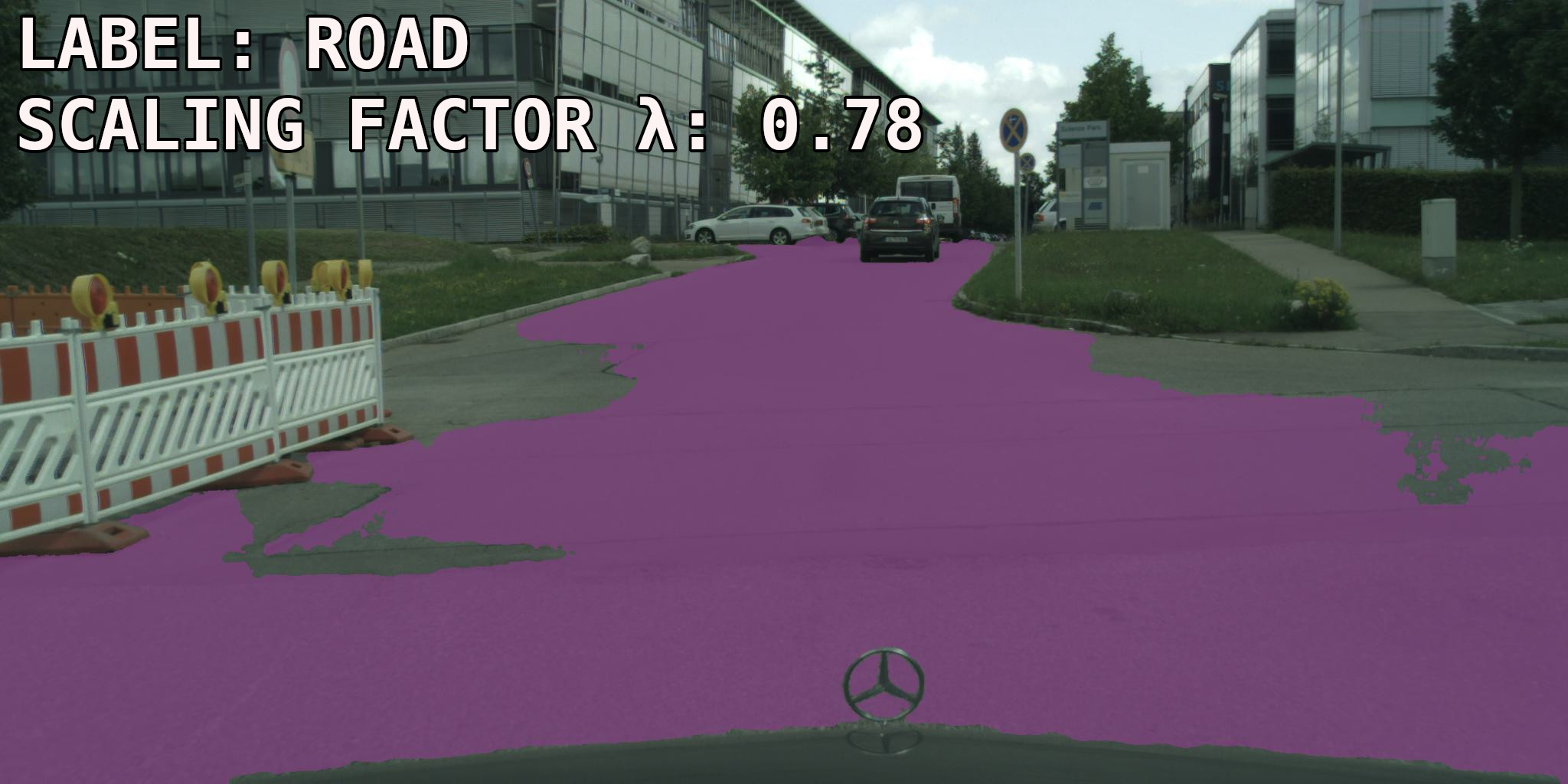} &
         \includegraphics[width=\mywt]{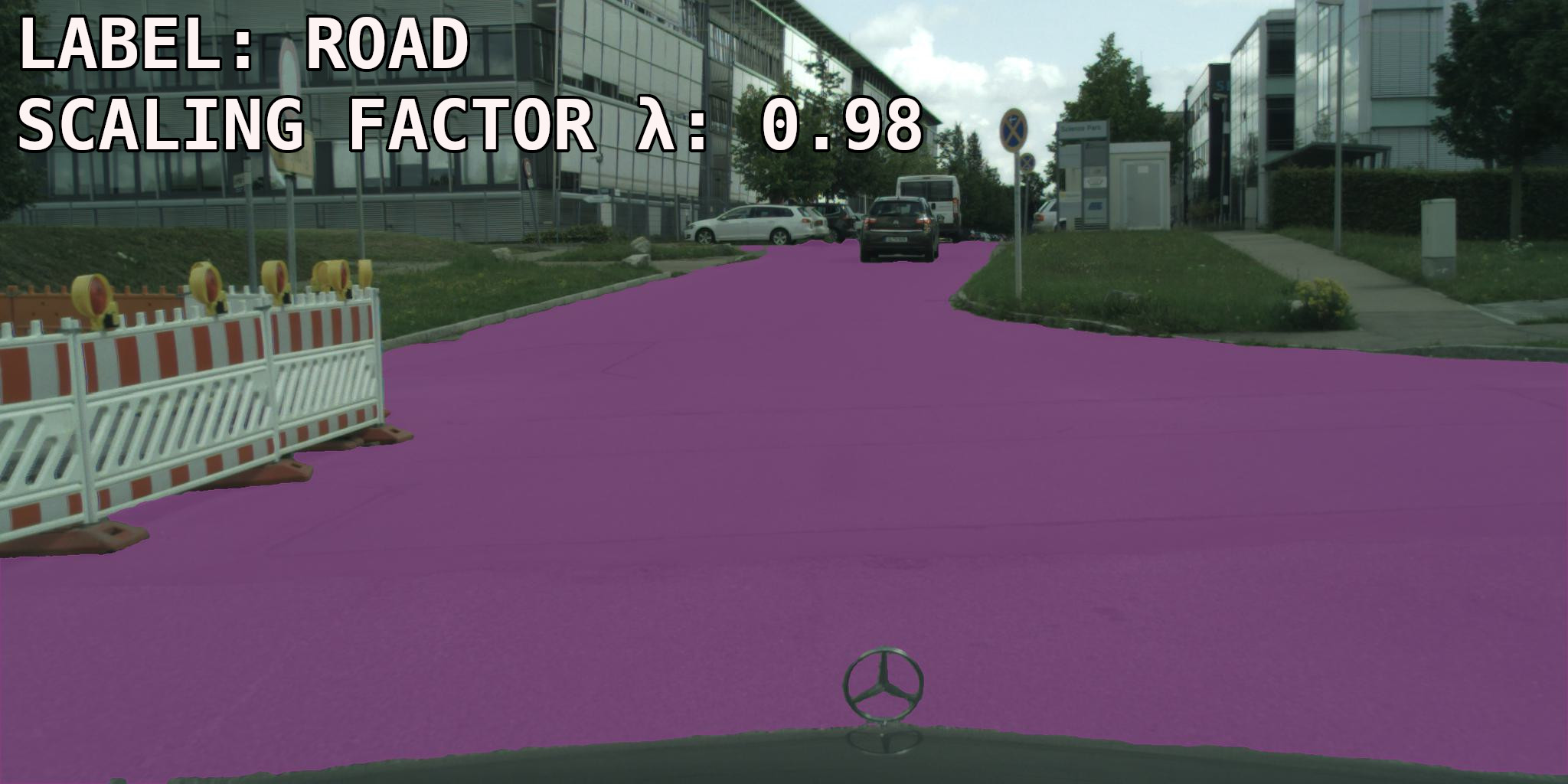} \\ [-0.2em]
         \includegraphics[width=\mywt]{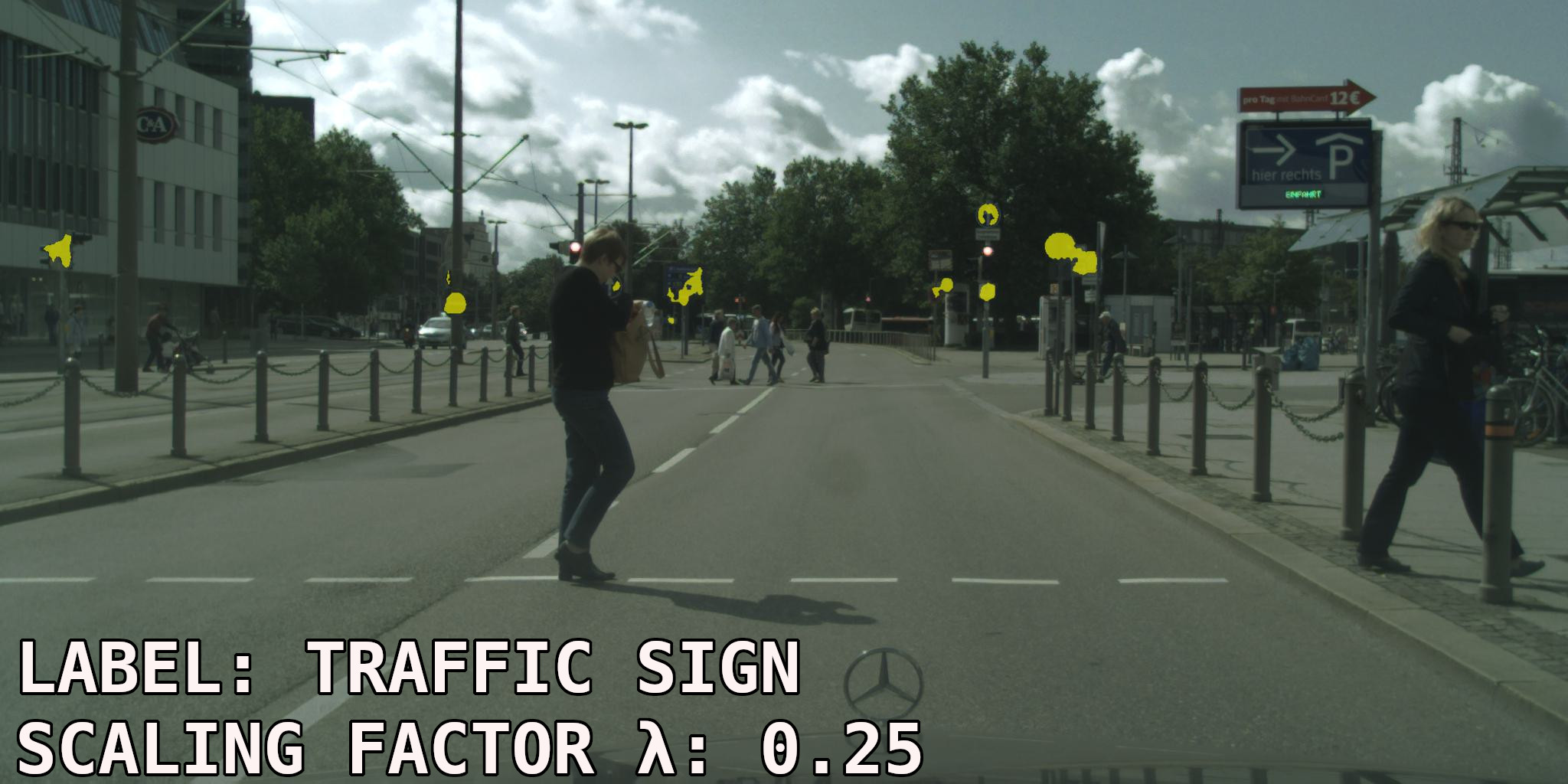} &
         \includegraphics[width=\mywt]{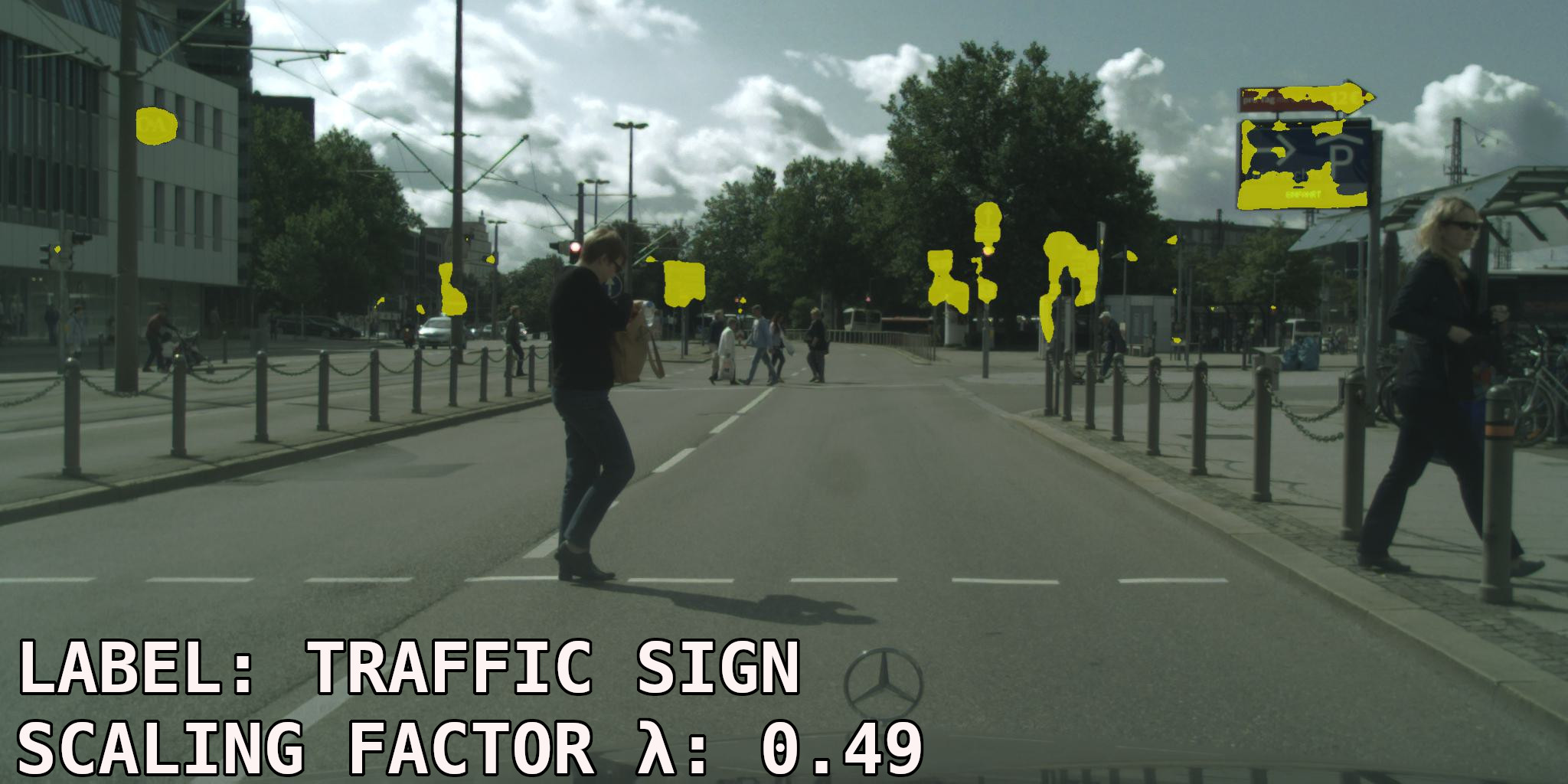} &
         \includegraphics[width=\mywt]{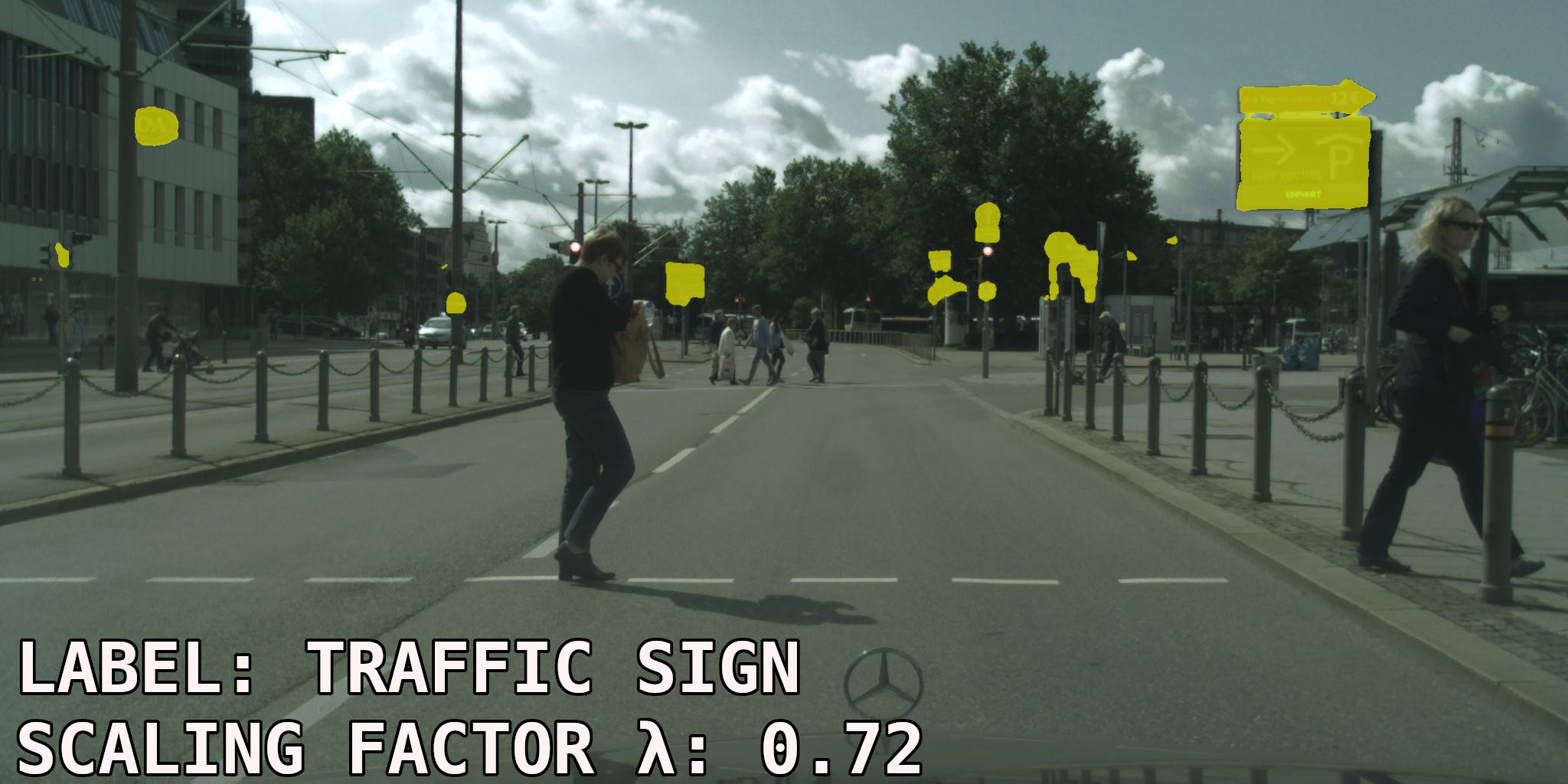} \\
    \end{tabular}
    \caption{
We visualize the masks corresponding to road and traffic signs,  and
their MLS factors
at three checkpoints.
There is
a strong correlation
between the MLS factor $\lambda$
and visual quality.}
    \label{fig:mask_quality}
\end{figure}

Figure~\ref{fig:low_conf_points}
shows 
that mask-wide loss-scaling 
and confidence-based point filtering 
exhibit complementary contributions 
to the selection of confident points
for domain-adaptive learning.
We observe that both the person mask (red, left) 
and the motorcycle mask (blue, right)
fail to attract all corresponding pixels.
We also observe that these false negative regions 
coincide with the teacher's low confidence points (red, middle). 
The figure shows that our two techniques 
prevent the student model 
to learn from incorrect pseudo-labels. 
\newcommand{\mywf}{0.325\textwidth}
\begin{figure}[h!]
    \centering
     \begin{tabular}{c@{\,}c@{\,}c}
         \includegraphics[width=\mywf]{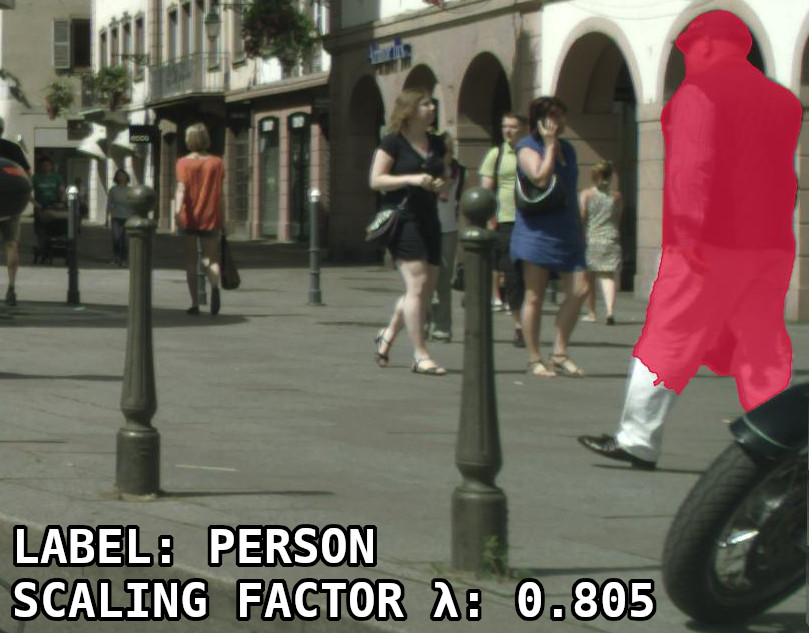} & 
         \includegraphics[width=\mywf]{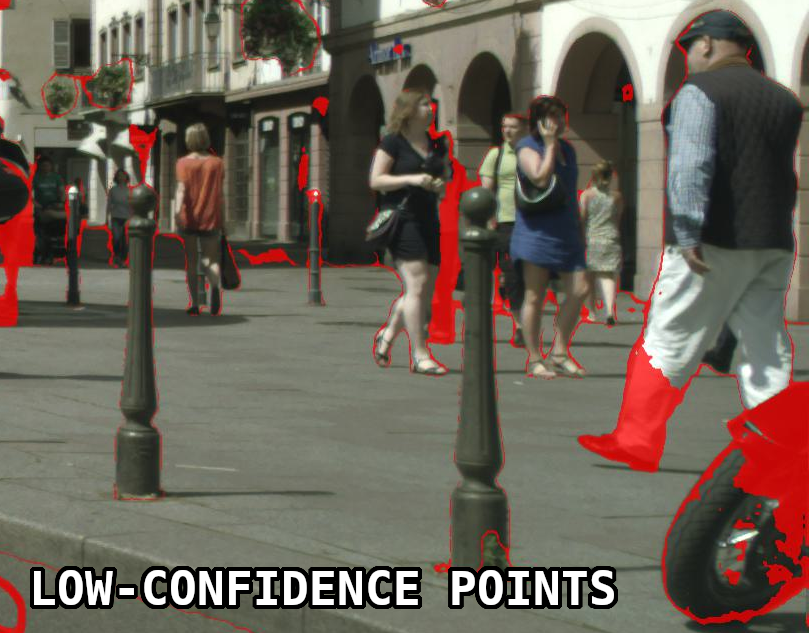} &
         \includegraphics[width=\mywf]{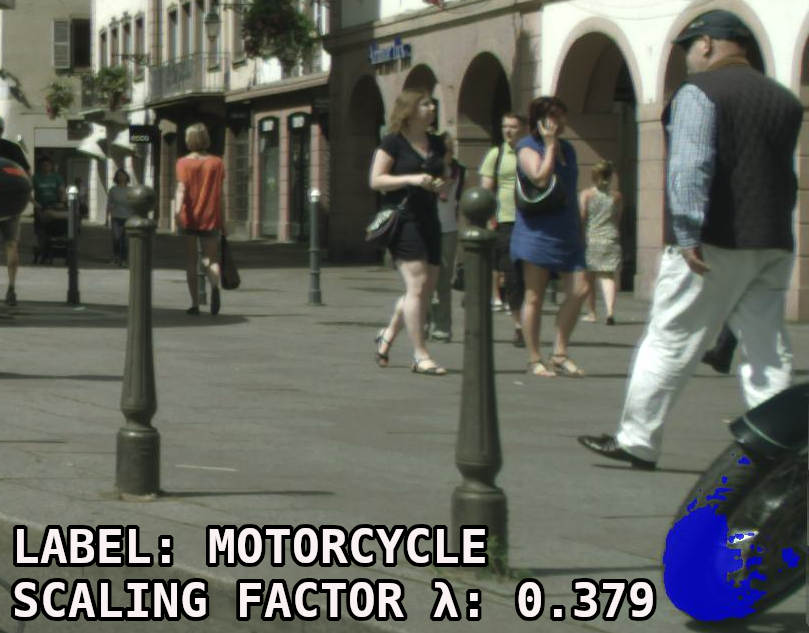} \\ 
    \end{tabular}
    \caption{
    Mask-wide loss-scaling and confidence-based point filtering exhibit
    complementary contributions to the selection of pixels 
    for domain-adaptive learning.
    The central image highlights the two regions
    where back-propagation is withheld
    due to low teacher confidence.
    The other two images show
    that these two regions correspond
    to false negative pixels in the person mask (left)
    and the motorcycle mask 
    (right).
}
    \label{fig:low_conf_points}
\end{figure}

\section{Limitations}
Experiments on Synthia$\rightarrow$Vistas show 
that our method may require
dataset-specific hyper parameters
in order to deliver the full extent 
of its generalization potential.
Our experiments have been performed
with batches of 2+2 images 
in order to fit on a single GPU with 40GB RAM.
We consider that as a limitation since
we imagine that many researchers
do not have access to such equipment.
We suspect that better performance
would be obtained with larger batches,
especially when training on datasets 
with high intra-class variation such as Vistas, 
since the recommended 
Mask2Former batch size is 16 \cite{cheng2022masked}.

\section{Conclusion}
All unsupervised domain adaptation methods
strive to avoid random solutions
due to noise amplification in the positive feedback.
Our method addresses that goal in the frame 
of a siamese consistency with hard predictions
by relying on fine-grained estimates of panoptic confidence.
In particular, we have proposed to weight
the dense self-supervised loss 
with mask-wide confidence, 
and subsample it with confidence-based point filtering.
This improves the training convergence
on target-domain images
by discouraging self-learning 
at uncertain image locations.
Extensive experimental evaluation reveals
consistent advantage 
with respect to the published state of the art.
In spite of encouraging experimental performance,
important challenges still remain.
We consider automatic
and 
per-class
selection of hyper-parameters
as prominent avenues 
for future work.
\section*{Acknowledgments}
This work has been
supported by
Croatian Recovery and Resilience Fund - NextGenerationEU 
(grant C1.4 R5-I2.01.0001),
Croatian Science Foundation 
(grant IP-2020-02-5851 ADEPT),
and the Advanced computing service
provided by the
University of Zagreb University Computing Centre - SRCE.
\bibliographystyle{splncs04}
\bibliography{references}
\clearpage
\appendix
\section*{Supplementary Material}
\section{Comparison with our Consistency Baseline}
Tables~\ref{tab:synthetic_to_real_baseline_vs_mcpanda} and~\ref{tab:city_to_other_baseline_vs_mcpanda}  present quantitative comparisons 
between the proposed domain adaptation method (MC-PanDA)
and our consistency baseline (BC).
We observe that our method 
consistently outperforms the baseline,  
which suggests a clear advantage
of the proposed contributions.
The advantage is the largest
in Table~\ref{tab:synthetic_to_real_baseline_vs_mcpanda}
where the supervision is available 
only for the synthetic domain.
\begin{table}[htb]
\caption{Comparison of MC-PanDA with our consistency baseline~(BC) on Synthia $\rightarrow$City and Synthia$\rightarrow$Vistas. All experiments are averaged over three random seeds.}
    \label{tab:synthetic_to_real_baseline_vs_mcpanda}
    \centering
    \begin{tabular}{l@{\,\,\,\,\,}ccc@{\quad}ccc}
        \toprule 
        & \multicolumn{3}{c}{Synthia$\rightarrow$City} & \multicolumn{3}{c}{Synthia$\rightarrow$Vistas} \\
        Method & $\text{SQ}_{16}$ & $\text{RQ}_{16}$ & $\text{PQ}_{16}$ & $\text{SQ}_{16}$ & $\text{RQ}_{16}$ & $\text{PQ}_{16}$ \\
        \midrule
        BC & \mstd{73.4}{0.6} & \mstd{50.0}{1.7} & \mstd{39.6}{1.5} & \mstd{72.3}{0.6} & \mstd{42.0}{1.1} & \mstd{32.2}{0.9}\\
        MC-PanDA & \mstd{76.7}{0.4} & \mstd{59.3}{1.0} & \mstd{47.4}{0.8} & \mstd{71.0}{0.7} & \mstd{49.8}{0.8} & \mstd{38.7}{1.0} \\ 
 
        \bottomrule
    \end{tabular}
\end{table}
\begin{table}[htb]
\caption{Comparison of MC-PanDA with our consistency baseline~(BC) on City$\rightarrow$Foggy and City$\rightarrow$Vistas. All experiments are averaged over three random seeds.}
    \label{tab:city_to_other_baseline_vs_mcpanda}
    \centering
    \begin{tabular}{l@{\,\,\,\,\,}ccc@{\quad}ccc}
        \toprule 
        & \multicolumn{3}{c}{Cityscapes$\rightarrow$Foggy} & \multicolumn{3}{c}{Cityscapes$\rightarrow$Vistas} \\
        Method & $\text{SQ}_{19}$ & $\text{RQ}_{19}$ & $\text{PQ}_{19}$ & $\text{SQ}_{19}$ & $\text{RQ}_{19}$ & $\text{PQ}_{19}$ \\
        \midrule
        BC & \mstd{82.6}{0.1} & \mstd{73.9}{0.5} & \mstd{61.7}{0.4} & \mstd{79.1}{0.1} & \mstd{59.8}{0.8} & \mstd{48.2}{0.7}\\
        MC-PanDA & \mstd{83.0}{0.2} & \mstd{73.9}{0.7} & \mstd{62.0}{0.4} & \mstd{79.6}{0.1} & \mstd{63.9}{0.3} & \mstd{51.7}{0.3} \\ 
 
        \bottomrule
    \end{tabular}
\end{table}
\section{{Impact of hyperparameters $\tau_1$ and $\tau_2$}}
{We did not optimize hyperparameters
to prevent over-optimistic results.
Table~\ref{tab:hyperparams-sensitivity} explores
the impact of $\tau_1$ and $\tau_2$ on
Synthia$\rightarrow$Cityscapes performance.
Due to limited GPU resources,
these experiments were conducted
with one random seed.
The experiments were run with 
$\tau_1 \in \{0.95, 0.96, 0.968, 0.98, 0.99\}$ 
and $\tau_2 \in \{0.7, 0.75, 0.8, 0.85, 0.9\}$.
The results show a minimum performance of 45.6~\pqs,
a maximum performance of 48.6~\pqs,
and a mean performance of \mstd{47.0}{0.8}~\pqs,
which represents improvements of 4.4, 7.4, and 5.8
percentage points over the state-of-the-art method
EDAPS~\cite{Saha_2023_ICCV}, respectively.}
\begin{table}[h]
    \centering
    \caption{{Impact of $\tau_1$ and $\tau_2$ on Synthia$\rightarrow$City \pqs~performance.}}
    \label{tab:hyperparams-sensitivity}
    \setlength{\tabcolsep}{6.0pt}
    \begin{tabular}{c|ccccc}
        \toprule
        $\tau_2 \backslash \tau_1$ & 0.95 & 0.96 & 0.968 & 0.98 & 0.99 \\
        \midrule
        0.7  & 46.9 & 47.7 & 47.8 & 46.7 & 46.4 \\
        0.75 & 48.6 & 48.0 & 48.3 & 47.8 & 47.3 \\
        0.8  & 46.1 & 46.0 & 47.2 & 46.4 & 46.3 \\
        0.85 & 46.0 & 45.6 & 46.7 & 46.8 & 48.1 \\
        0.9  & 47.0 & 46.5 & 46.7 & 47.1 & 46.6 \\
        \bottomrule
    \end{tabular}
\end{table}
\section{{Additonal ablations}}
{Table~\ref{tab:ablation_synthia_vistas}
presents the ablation study
of the proposed
mask-wide loss scaling (MLS)
and confidence-based point filtering (CBPF)
on Synthia$\rightarrow$Vistas.
The baseline consistency model (BC)
surpasses the supervised model
trained on Synthia by 7 points.
Including MLS or CBPF further
improves the performance
over the consistency baseline by
4.9 or 2.9 points, respectively.
Finally, the combination of MLS and CBPF
outperforms the consistency baseline by 6.5 points,
and demonstrates the
complementary effect of the proposed contributions.}
\begin{table}[h]
    \centering
    \caption{{Ablation study on Synthia$\rightarrow$Vistas: baseline consistency (BC), mask-level loss scaling (MLS), and confidence-based point filtering (CBPF). Top row represents supervised training on Synthia. We report mean$_{\pm\text{std}}$ over three random seeds.}}
    \label{tab:ablation_synthia_vistas}
    \begin{tabular}{ccccc}
    \toprule
      $\text{BC}$ & $\text{MLS}_{0.99}$ & 
         $\text{CBPF}_{0.9}$ & \multicolumn{2}{c}{$\text{PQ}_{16}$} 
         
      \\
      \midrule
      \no & \no  & \no  & 
        25.2 & \blarrow  
      \\
      \yes & \no  & \no  & 
        \mstd{32.2}{0.9} & 
        \textcolor{Green}{\textbf{+7.0}} 
      \\
      \yes & \yes & \no  & 
        \mstd{37.1}{1.0}& 
        \textcolor{Green}{\textbf{+11.9}}
      \\
      \yes & \no  & \yes & 
        \mstd{35.1}{1.3}& 
        \textcolor{Green}{\textbf{+9.9}}
      \\
      \yes & \yes & \yes & 
        \mstd{38.7}{1.0}& 
        \textcolor{Green}{\textbf{+13.5}}
      \\
      \bottomrule
    \end{tabular}
\end{table}

\section{Per-mask vs. All-mask Teacher Confidence for Point Filtering}
Figure~\ref{fig:app:per_mask_vs_all} 
provides a visual 
comparison between
all-mask 
confidence-based
point filtering (CBPF) 
used in our method
and per-mask CBPF,
as detailed in
Table~6 and Figure~6
of the main manuscript.
The left and middle columns
highlight complementary
contributions of mask-wide loss scaling (MLS)
and all mask CBPF.
The transition from the middle
to the right column demonstrates
the difference between
all-mask CBPF
and per-mask CBPF.
A closer examination of the
top-right image reveals that
per-mask CBPF is unable to detect
false negative assignments
far from the mask border.
Incorporating these points into 
the filtering process would
imply learning on a substantial
number of false negative pixels 
in the lower portion of the left leg.
These observations are in 
concordance with Table~6 
from the main manuscript,
which reports 
considerable advantage of
all-mask CBPF
in comparison 
with per-mask CBPF.
\newcommand{\mywfa}{0.325\textwidth}
\begin{figure}[h]
    \centering
    %\begin{tabular}{c@{\,}c@{\,}c@{\,}c}
     \begin{tabular}{c@{\,}c@{\,}c}
         \includegraphics[width=\mywfa]{figures/complementary/person_conf_v3.jpg} & 
         \includegraphics[width=\mywfa]{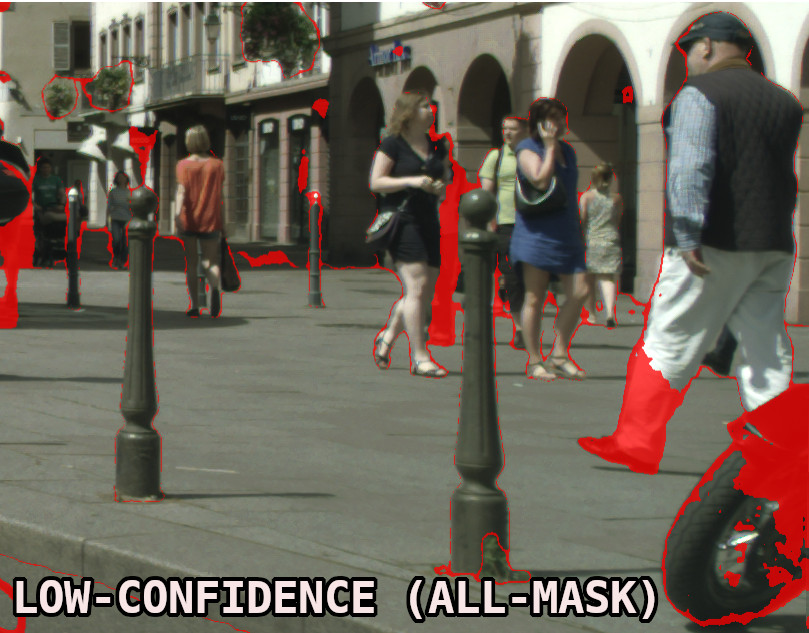} &
         \includegraphics[width=\mywfa]{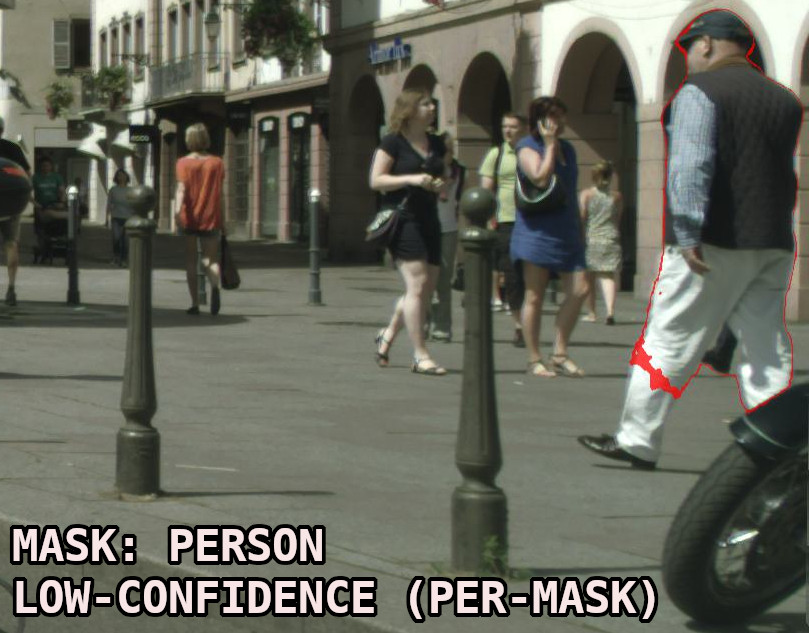} \\[0.5em]
         \includegraphics[width=\mywfa]{figures/complementary/motorcyc_conf_v3.png} & 
         \includegraphics[width=\mywfa]{figures/appendix/unc_per_mask_vs_all/uncertainty_better.jpg} &
         \includegraphics[width=\mywfa]{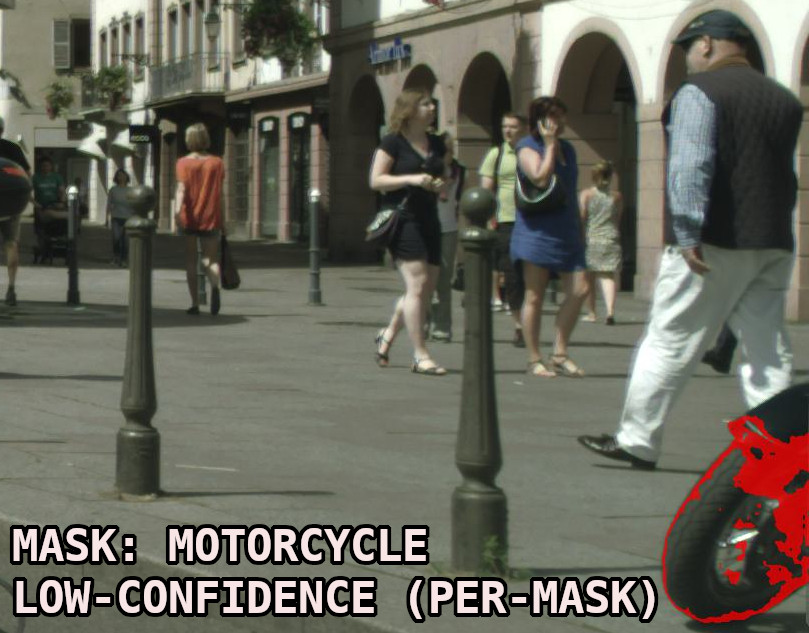} \\ 
    \end{tabular}
    \caption{Visual comparison of per-mask point filtering and all-mask point filtering (\cf~main manuscript - Table 6). The top row shows a mask classified as a person (left image), uncertain pixels with all-mask filtering (middle image), and uncertain pixels with per-mask filtering (right image). The bottom row shows the same intermediate results
    for a mask that corresponds to a motorcycle.
    We observe that all-mask filtering 
    outperforms per-mask filtering 
    as a detector of false negative pixel assignments 
    for the two considered masks.}
    \label{fig:app:per_mask_vs_all}
\end{figure}
\section{Failure Modes}
Figure~\ref{fig:app_failure_mods} illustrates
over-pessimistic mask-wide loss confidence estimates
during the initial phases of training as
produced by a teacher checkpoint 
after 25k iterations of domain adaptation learning.
We notice considerable discrepancies between
the observed localization performance and our mask-wide loss
confidence estimates $\lambda$.
This observation suggests that
our method might benefit from
either per-class or adaptively
defined thresholds $\tau_1$.
\newcommand{\mywff}{0.33\textwidth}
\begin{figure}[h!]
    \centering
    %\begin{tabular}{c@{\,}c@{\,}c@{\,}c}
     \begin{tabular}{c@{\,}c@{\,}c}
         \includegraphics[width=\mywff,height=0.2\textwidth]{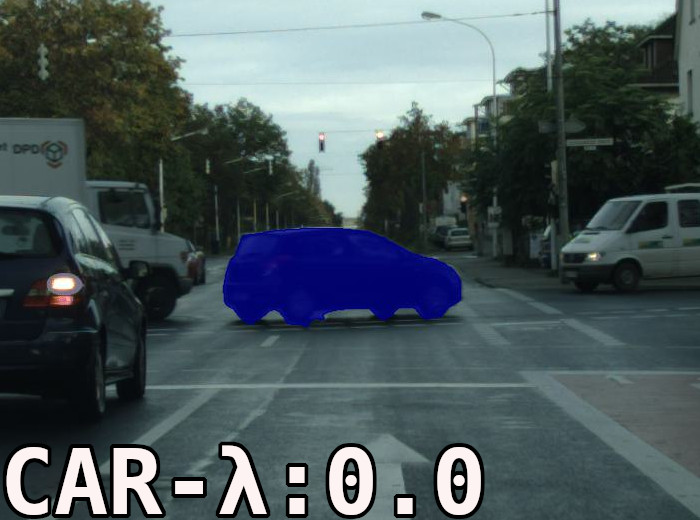} & 
         \includegraphics[width=\mywff,height=0.2\textwidth]{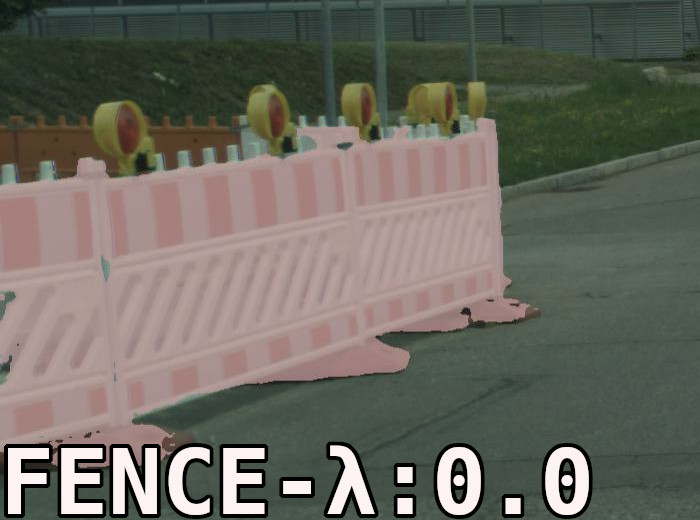} &
         \includegraphics[width=\mywff,height=0.2\textwidth]{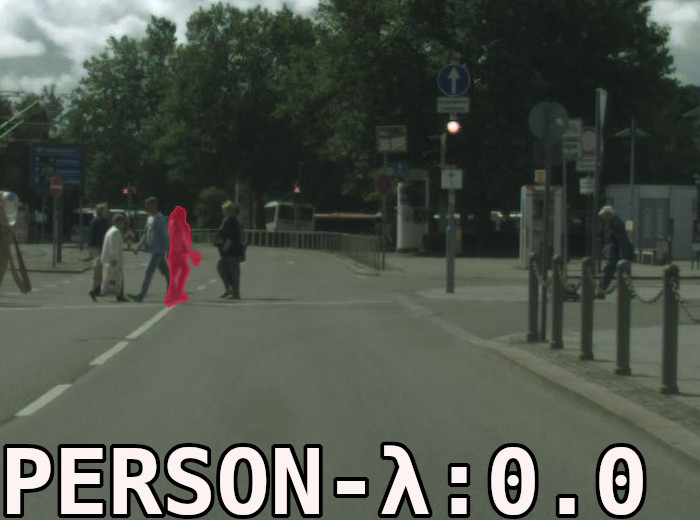} \\ 

         \includegraphics[width=\mywff,height=0.2\textwidth]{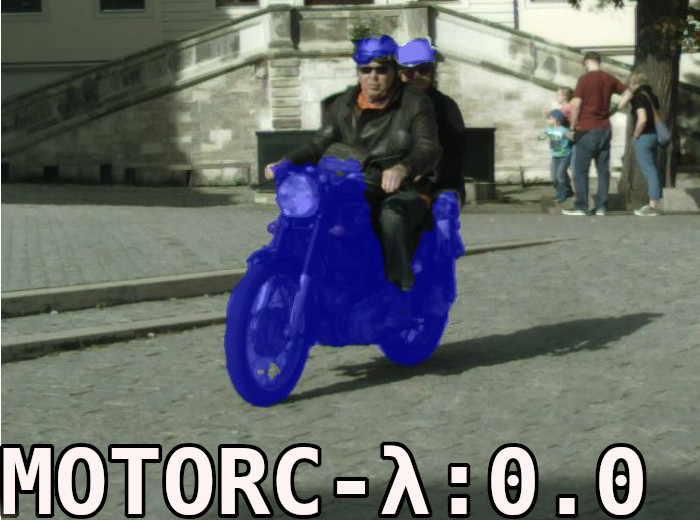} & 
         \includegraphics[width=\mywff,height=0.2\textwidth]{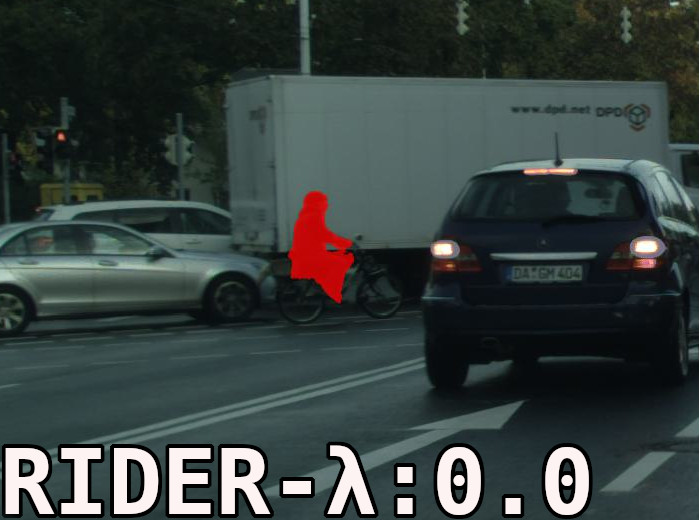} &
         \includegraphics[width=\mywff,height=0.2\textwidth]{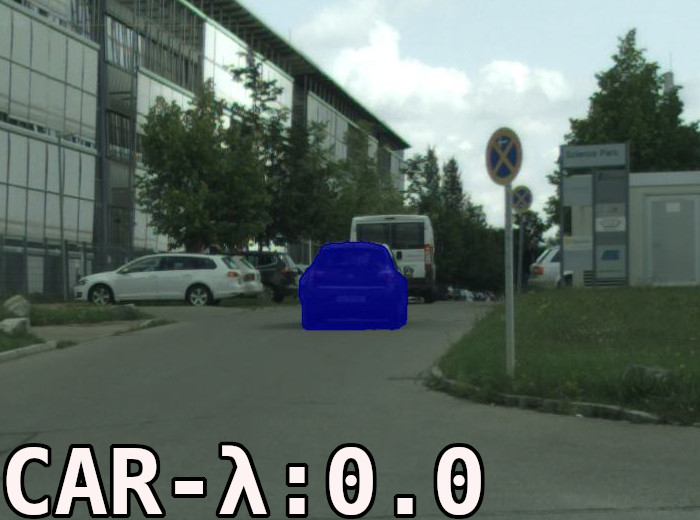} \\ 
    \end{tabular}
    \caption{Failures due to overly pessimistic estimates of mask-wide confidence in several examples generated by the teacher model during the early stages of training. Qualitatively, we observe masks that were assigned a mask-wide confidence of $\lambda=0$ in spite of being nearly perfectly localized.}
    \label{fig:app_failure_mods}
\end{figure}

\section{Further Implementation Details}
\subsection{Augmentations}
We follow the usual image augmentation 
pipeline from M2F~\cite{cheng2022masked} that consists of
random image resizing with constant aspect ratio, 
random cropping, 
horizontal flipping, 
and SSD color jittering \cite{liu2016ssd}. 
We use crop size of $512 \times 1024$ pixels in all our experiments.
We randomly sample 
the size of the shorter image side
from the following intervals:
\begin{itemize}
    \item[$\bullet$] \texttt{[512, 2048]} for Cityscapes and Foggy Cityscapes,
    \item[$\bullet$] \texttt{[640, 1408]} for Synthia, and
    \item[$\bullet$] \texttt{[1024, 4096]} for Vistas.    
\end{itemize}
We omit the color jitter
when preparing the unlabeled
image for the teacher.
We additionally perturb
the student image
using the following sequence
of transforms from \texttt{torchvision}~\cite{torchvision2016}:
\begin{enumerate}
    \item 
    \texttt{ColorJitter}(\\
    \texttt{\text{brightness}=(0.2, 1.8)},\\
    \texttt{\text{contrast}=(0.2, 1.8)}, \\
    \texttt{\text{saturation}=(0.2, 1.8)}, \\ 
    \texttt{\text{hue}=(-0.2, 0.2)})
    \item \texttt{RandomGrayscale(\text{\text{p}=0.2})}
    \item \texttt{RandomApply(GaussianBlur(\text{\text{sigma}=(0.1, 2)), \textbf{p}=0.5})}
\end{enumerate}
Finally, we apply SegMix 
as described in the main manuscript.
Figure~\ref{fig:train_ex_vis}
illustrates three random
training examples
from the Synthia$\rightarrow$Cityscapes experiment.
The first two columns show
augmented images 
from the source and the target domain.
The last two columns
show the student image
and the corresponding labels
after SegMix.
Note that the student labels
correspond to a mix of 
the teacher pseudolabels
and ground truth labels 
from the source domain.
We observe that in some cases
most of the student pixels
come from Synthia,
which may be suboptimal.
We believe that our
segment-oriented
mixing strategy
represents a fertile ground
for future upgrades.

\newcommand{\mywa}{0.245\textwidth}
\begin{figure}[h]
    \centering
    \begin{tabular}{c@{\,}c@{\,}c@{\,}c}
        \scriptsize Source Image & \scriptsize Target Image & \scriptsize Student Image & \scriptsize Augmented Pseudolabels \\
         \includegraphics[width=\mywa]{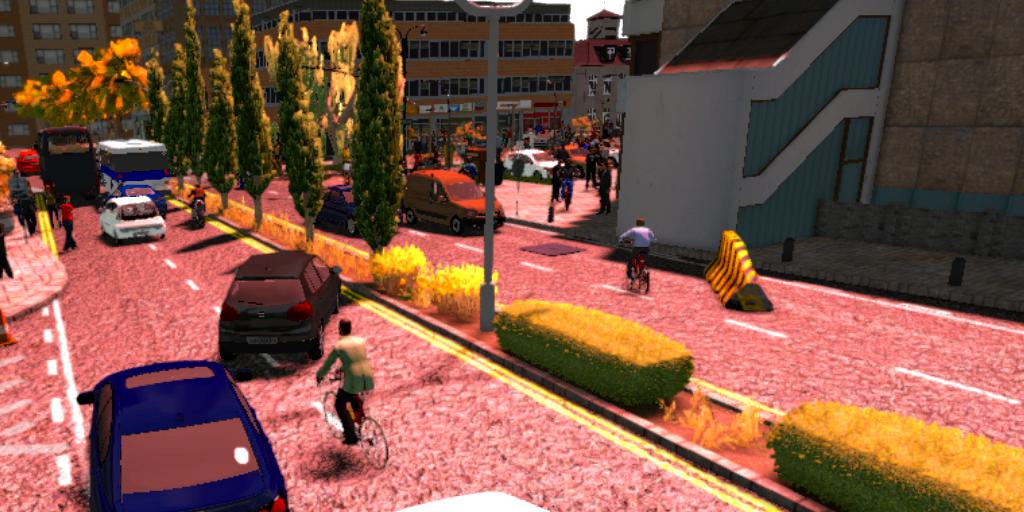} & 
         \includegraphics[width=\mywa]{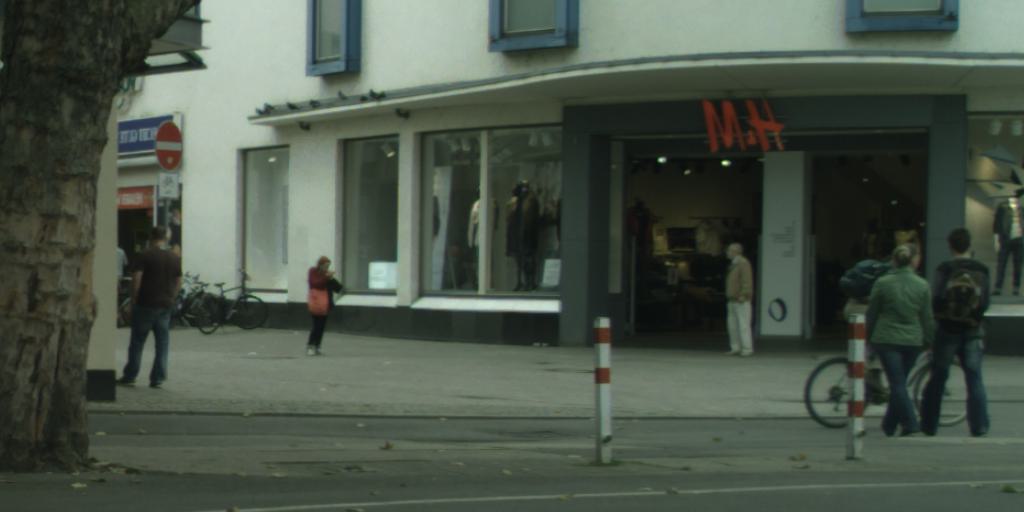} &
         \includegraphics[width=\mywa]{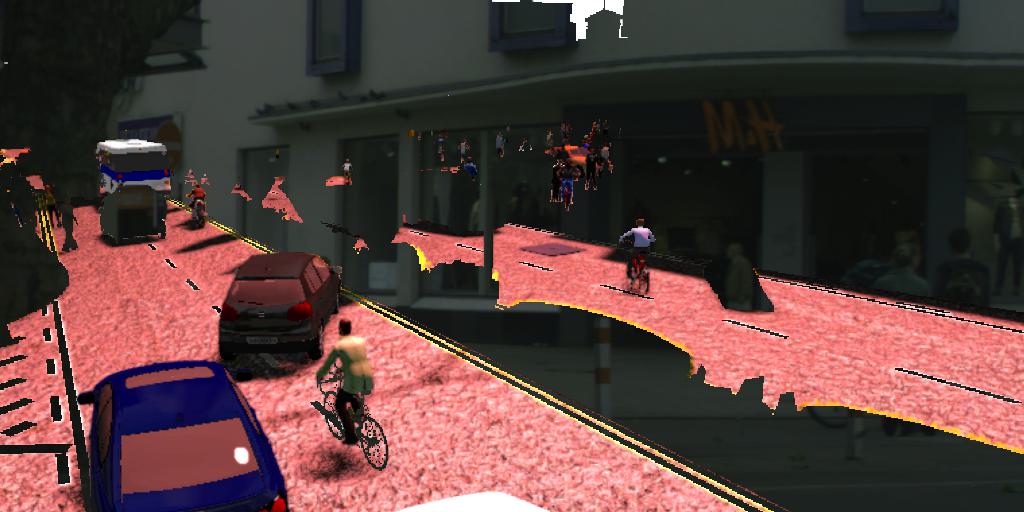} &
         \includegraphics[width=\mywa]{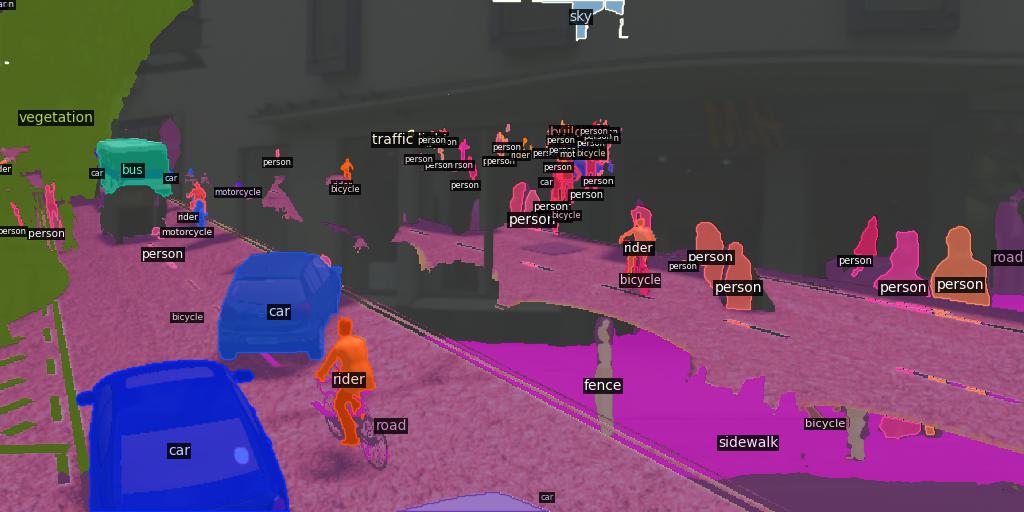} \\[-0.2em]

         \includegraphics[width=\mywa]{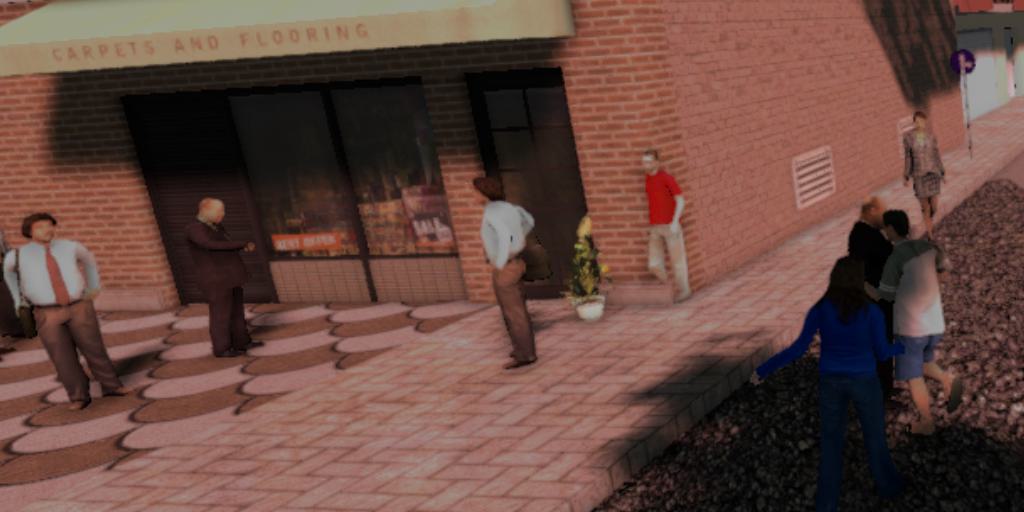} & 
         \includegraphics[width=\mywa]{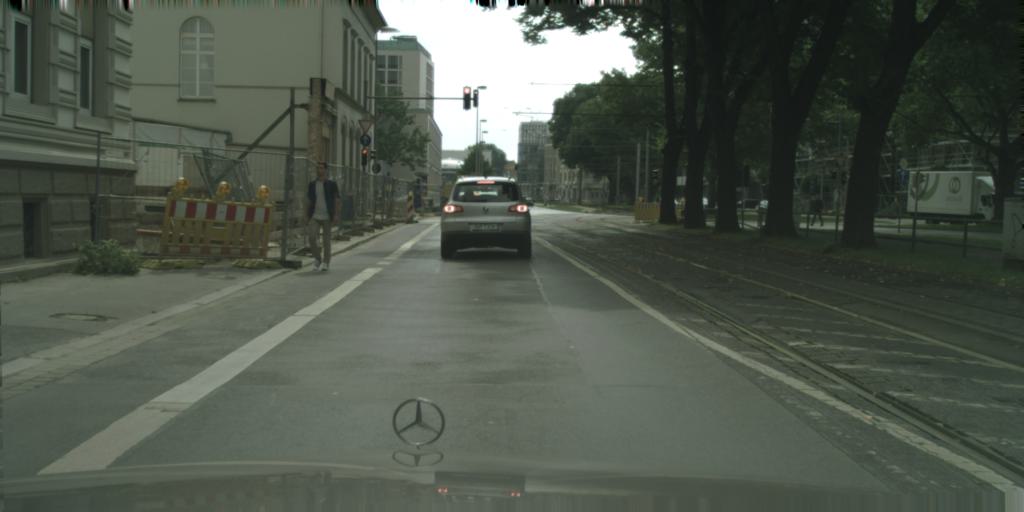} &
         \includegraphics[width=\mywa]{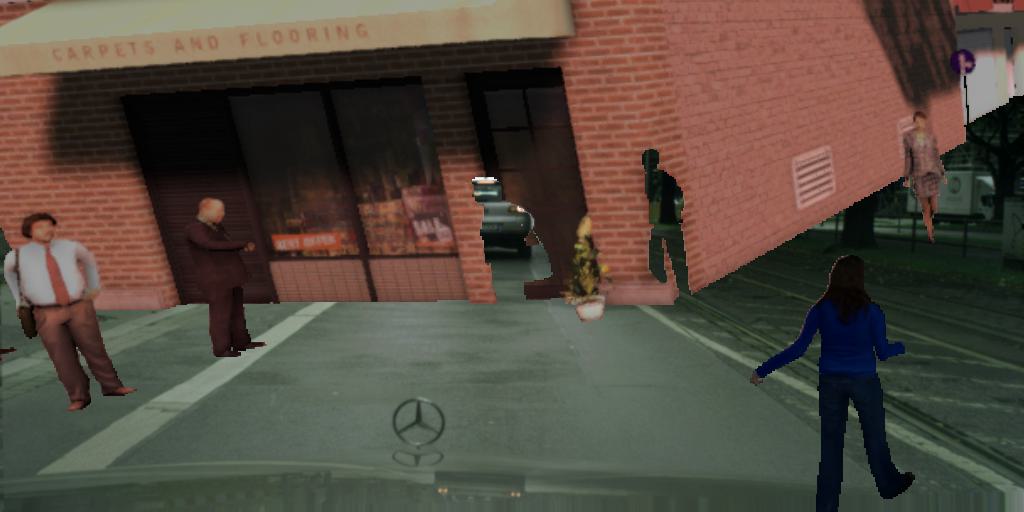} &
         \includegraphics[width=\mywa]{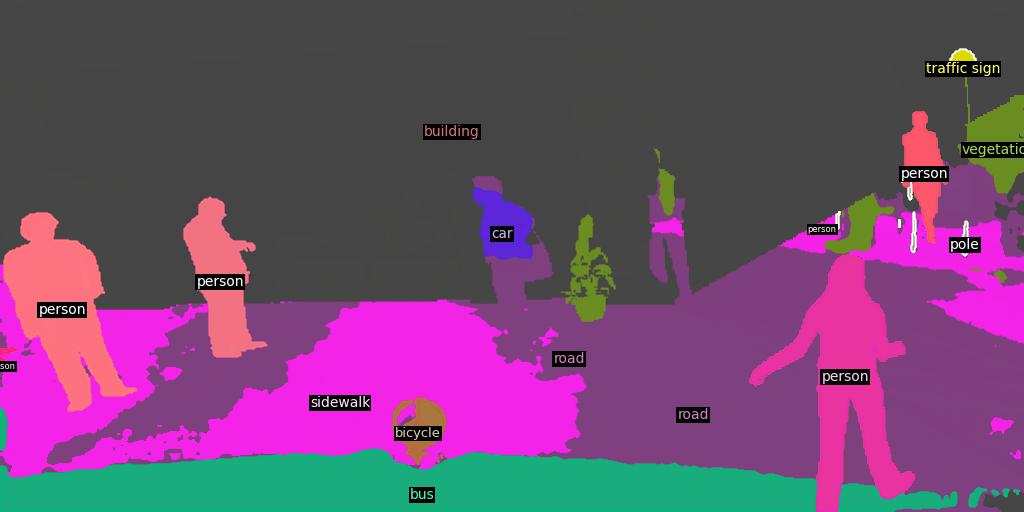} \\[-0.2em]

         \includegraphics[width=\mywa]{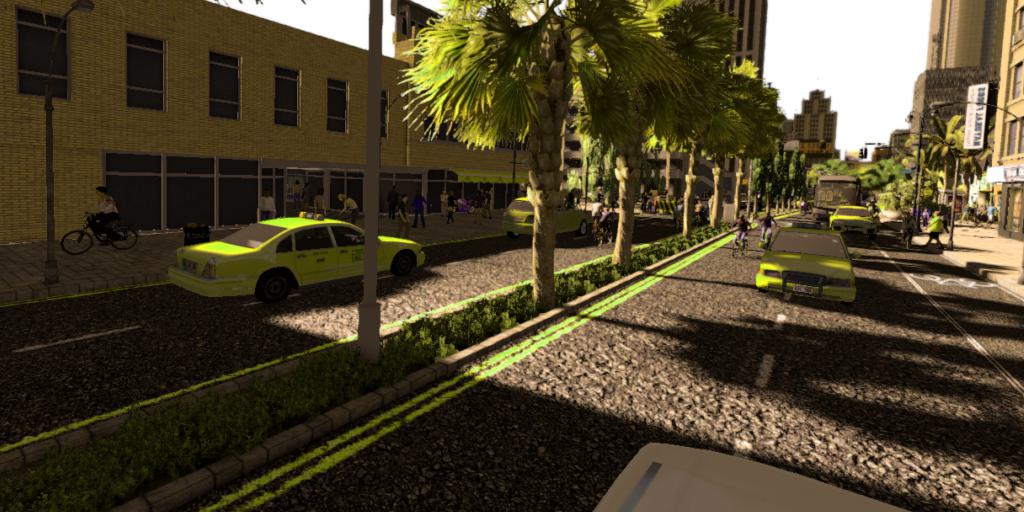} & 
         \includegraphics[width=\mywa]{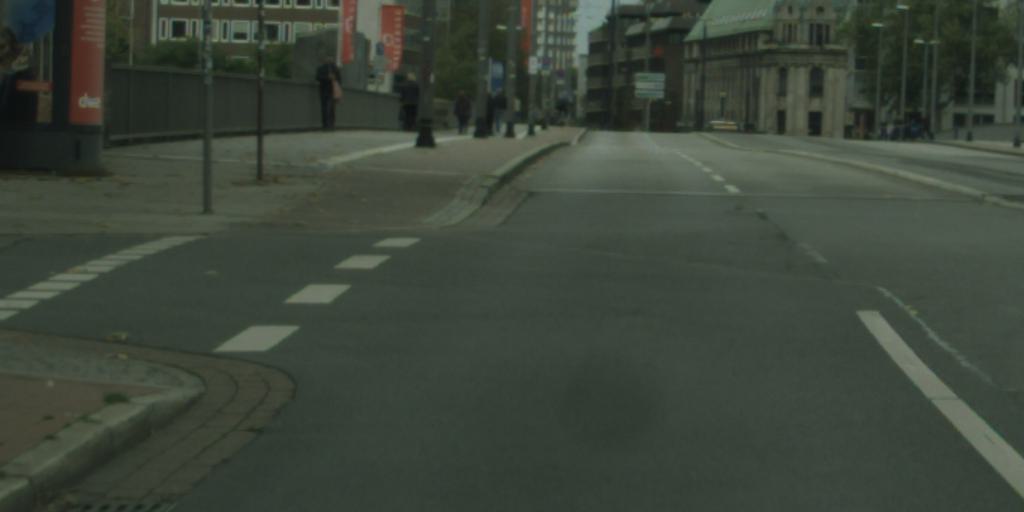} &
         \includegraphics[width=\mywa]{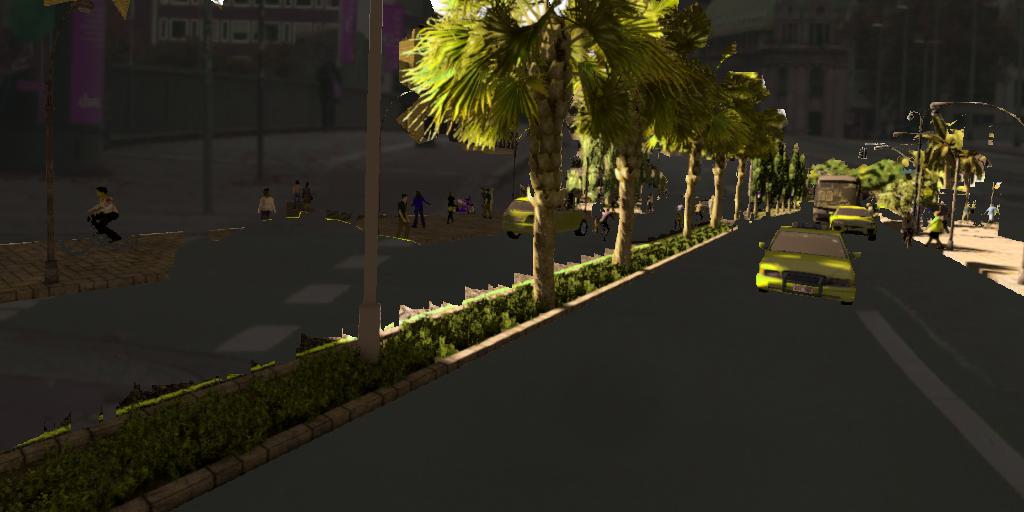} &
         \includegraphics[width=\mywa]{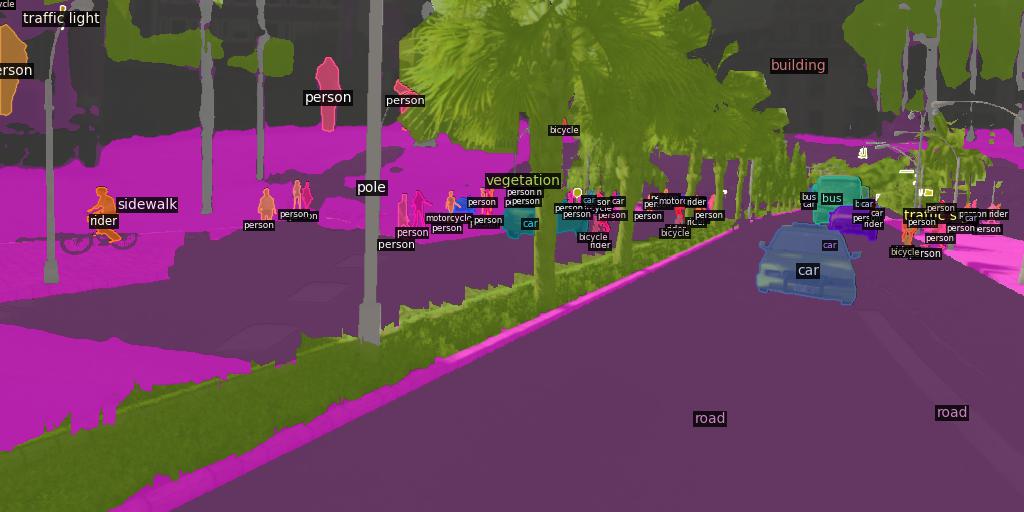} \\[-0.2em]

    \end{tabular}
    \caption{
        Target domain training examples
        are augmented by pasting masks
        from labeled images from the source domain.
        SegMix samples
        half of the segments 
        % (instead of classes) 
        from the source image
        and pastes them
        on top of the target image.
        Note that this is different from ClassMix,
        which samples half of the classes 
        present in the source image
        and pastes all corresponding segments. 
    }
    \label{fig:train_ex_vis}
\end{figure}
\subsection{Hyperparameters}
In experiments where Cityscapes and Synthia serve as the source domains, we set the number of Mask2Former queries to 100~\cite{cheng2022masked} and 200, respectively. We increase the number of queries for Synthia$\rightarrow$* due to larger average number of instances per image in Synthia (approximately 152). 
In contrast, average number of instances per image in Cityscapes is only 27. 
We follow~\cite{cheng2022masked} 
and sample $N_p=112\times112$
points for loss computation,
where $\beta=75\%$ of them
correspond to points with
largest sampling affinity.
\section{Qualitative Examples} 
\begin{figure}[h]
    \centering
    \begin{tabular}{c@{\,}c@{\,}c@{\,}c}
        \scriptsize Image & \scriptsize GT & \scriptsize EDAPS~\cite{Saha_2023_ICCV} & \scriptsize MC-PanDA \\
         \includegraphics[width=0.24\textwidth]{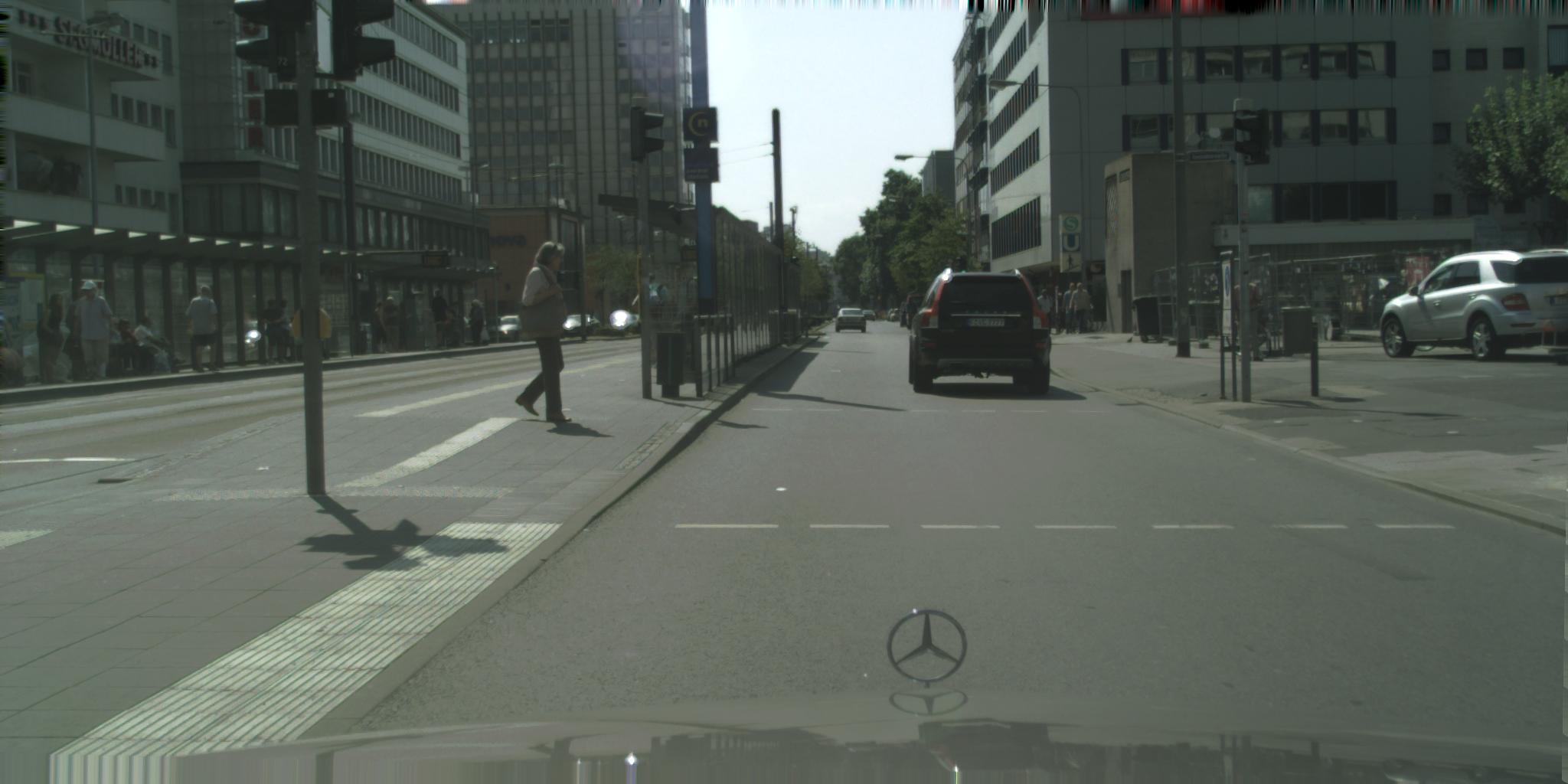} & 
         \includegraphics[width=0.24\textwidth]{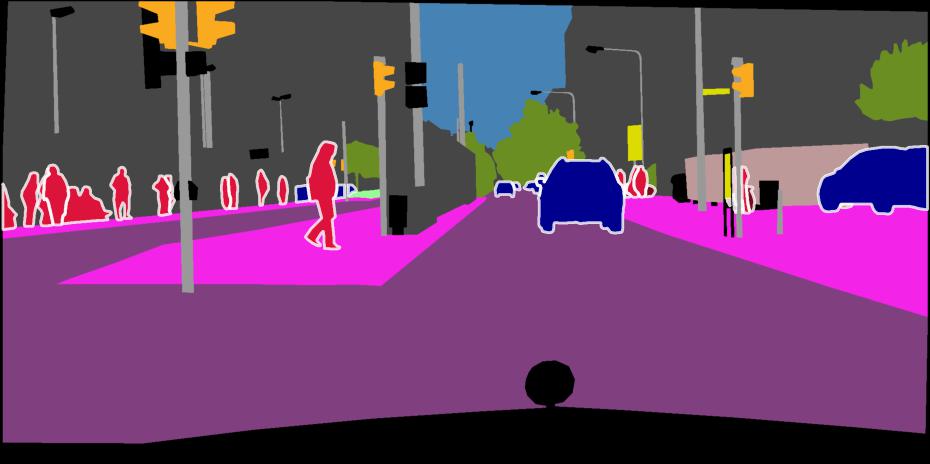} &
         \includegraphics[width=0.24\textwidth]{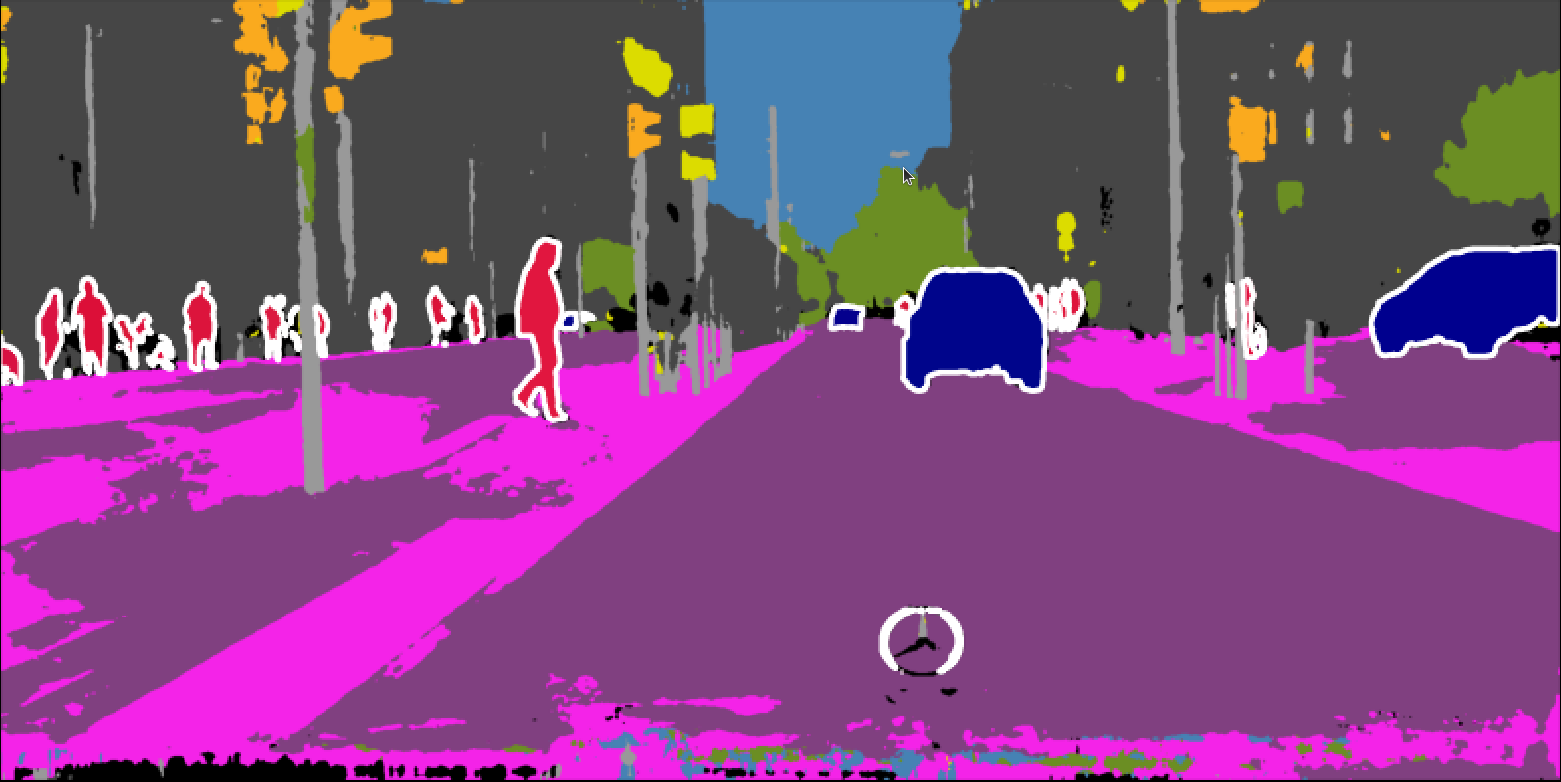} &
         \includegraphics[width=0.24\textwidth]{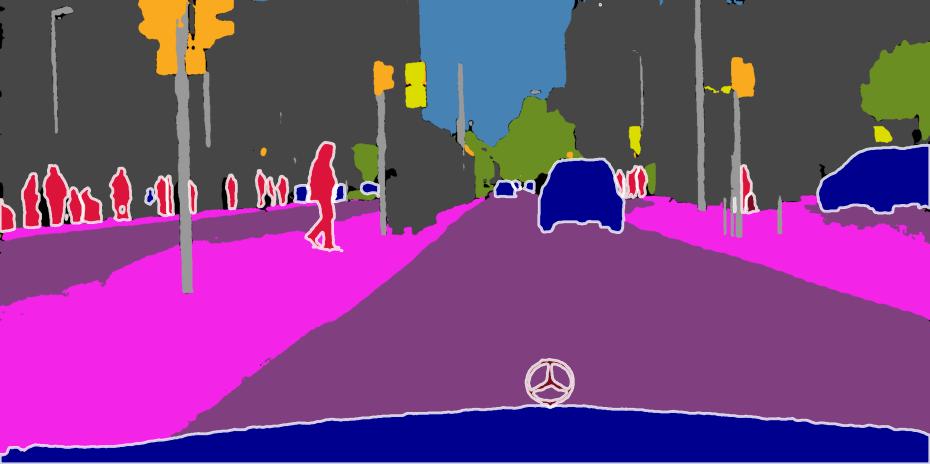} \\[-0.2em]

         \includegraphics[width=0.24\textwidth]{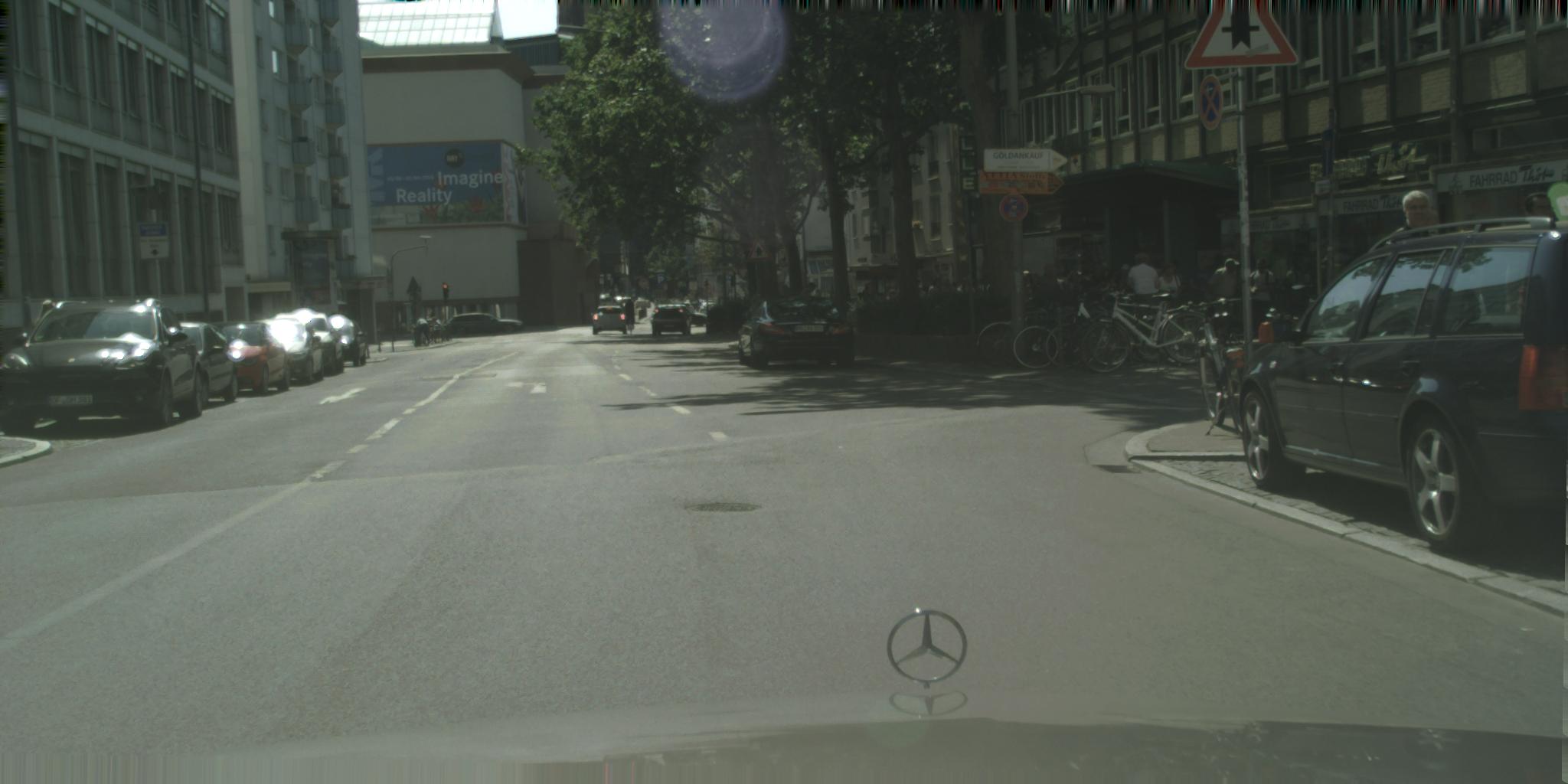} & 
         \includegraphics[width=0.24\textwidth]{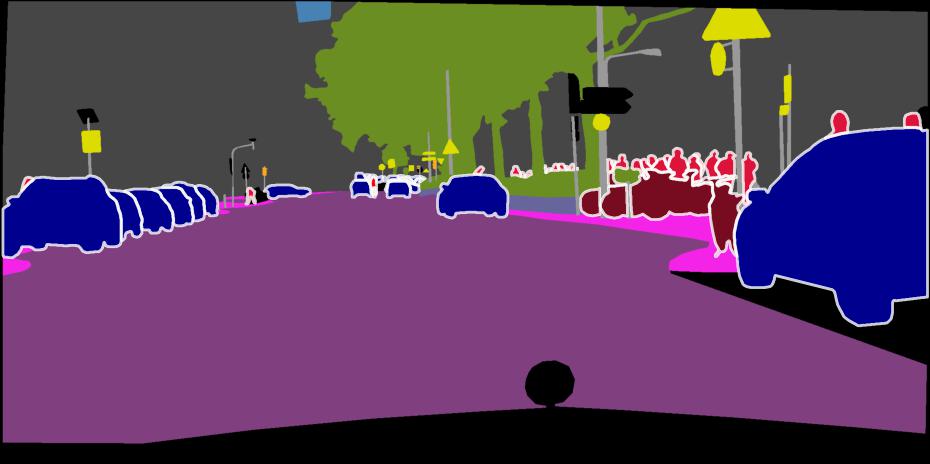} &
         \includegraphics[width=0.24\textwidth]{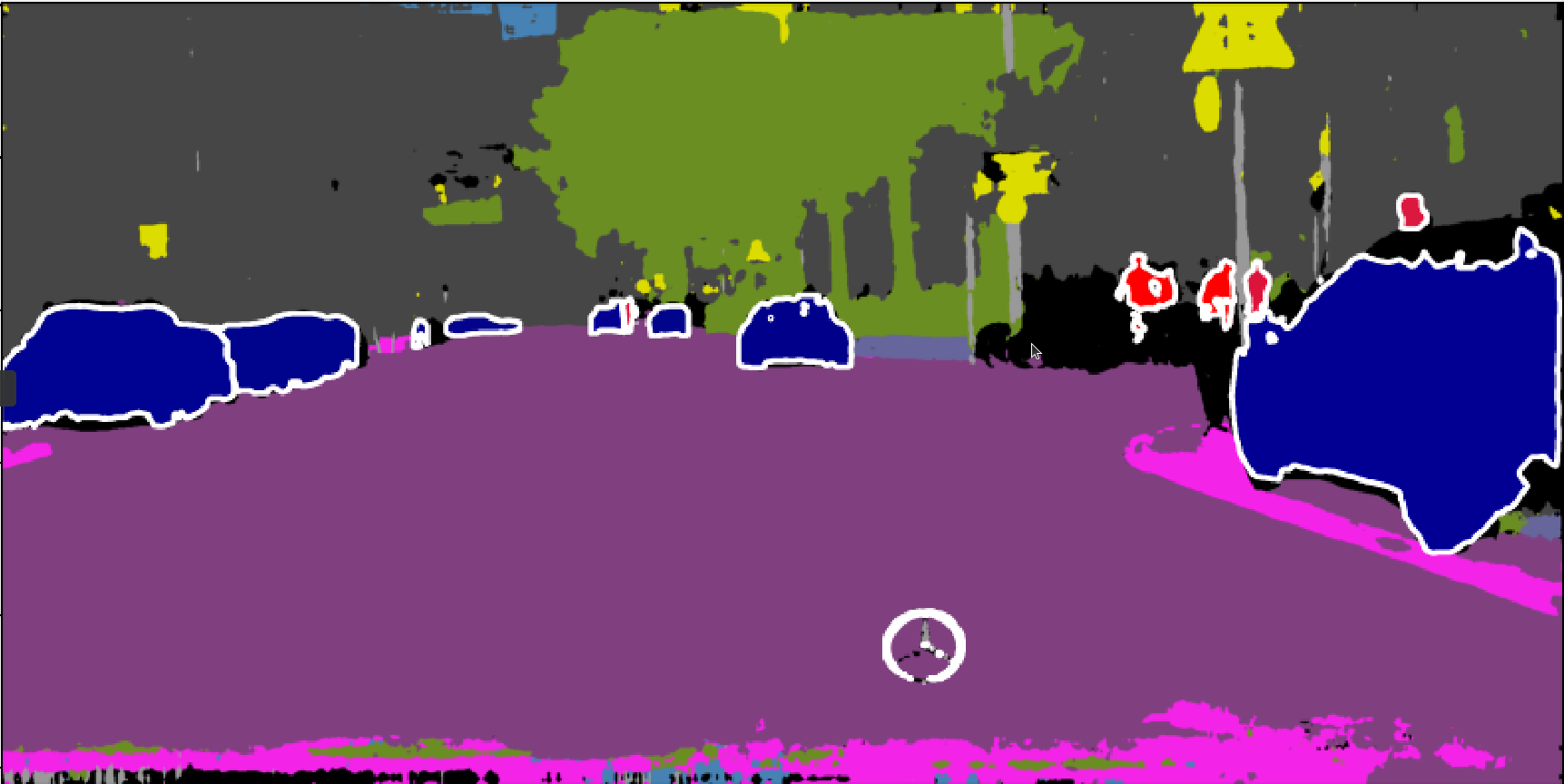} &
         \includegraphics[width=0.24\textwidth]{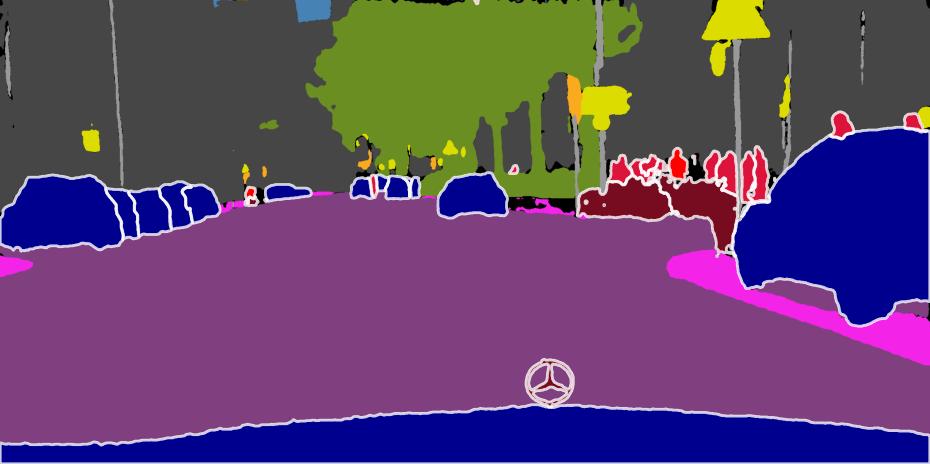} \\[-0.2em]
        
        \includegraphics[width=0.24\textwidth]{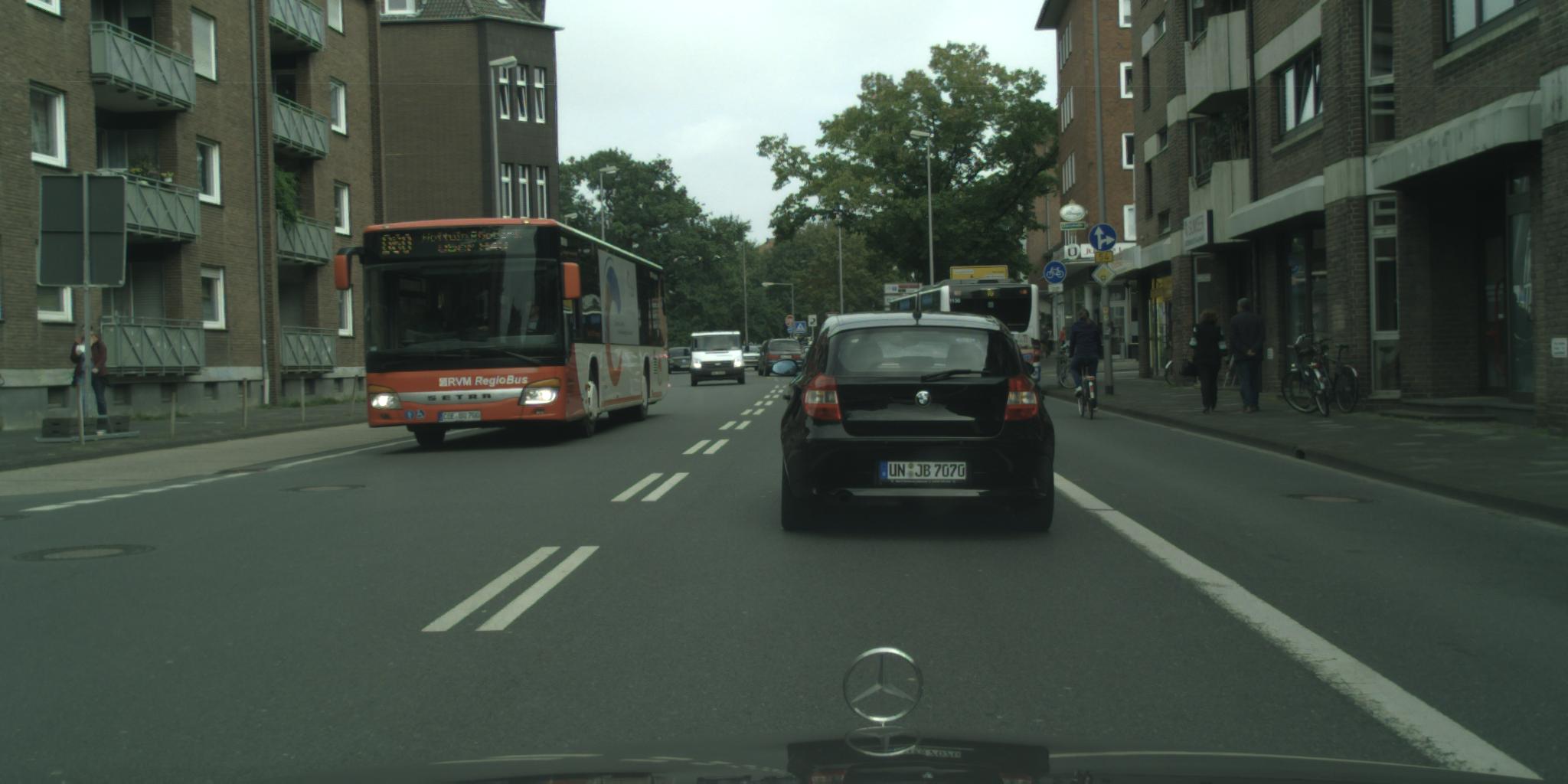} & 
         \includegraphics[width=0.24\textwidth]{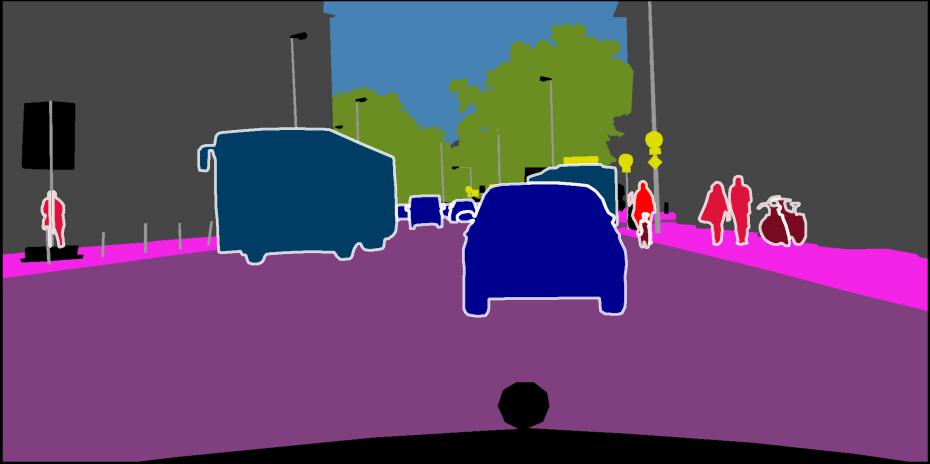} &
         \includegraphics[width=0.24\textwidth]{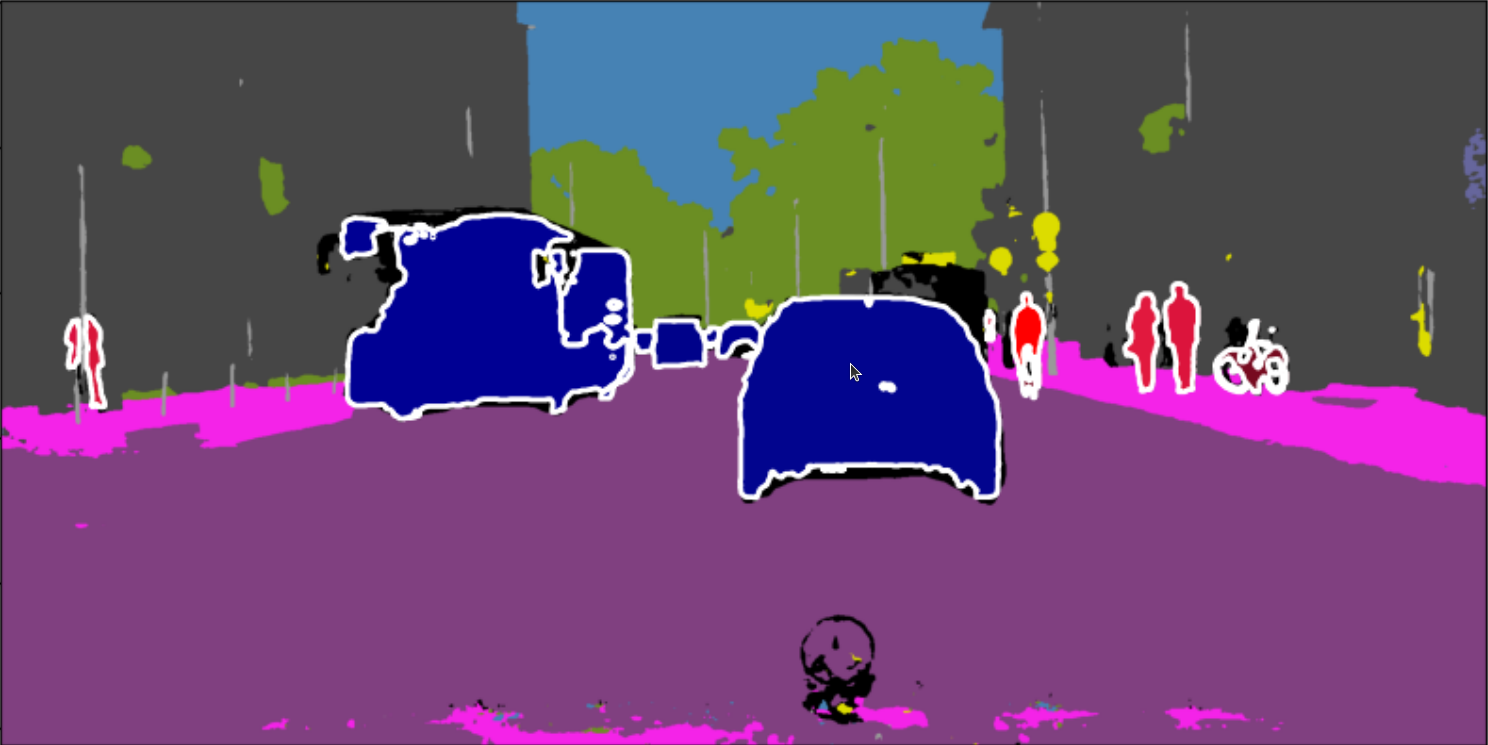} &
         \includegraphics[width=0.24\textwidth]{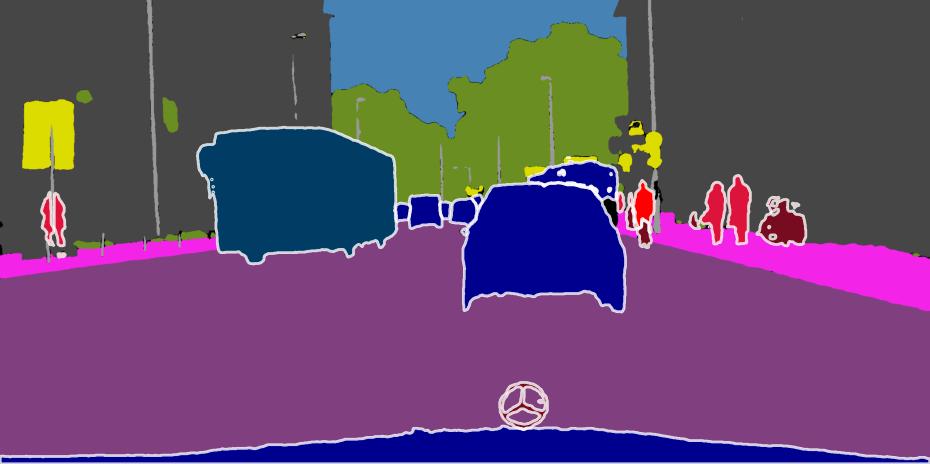} \\[-0.2em]

        \includegraphics[width=0.24\textwidth]{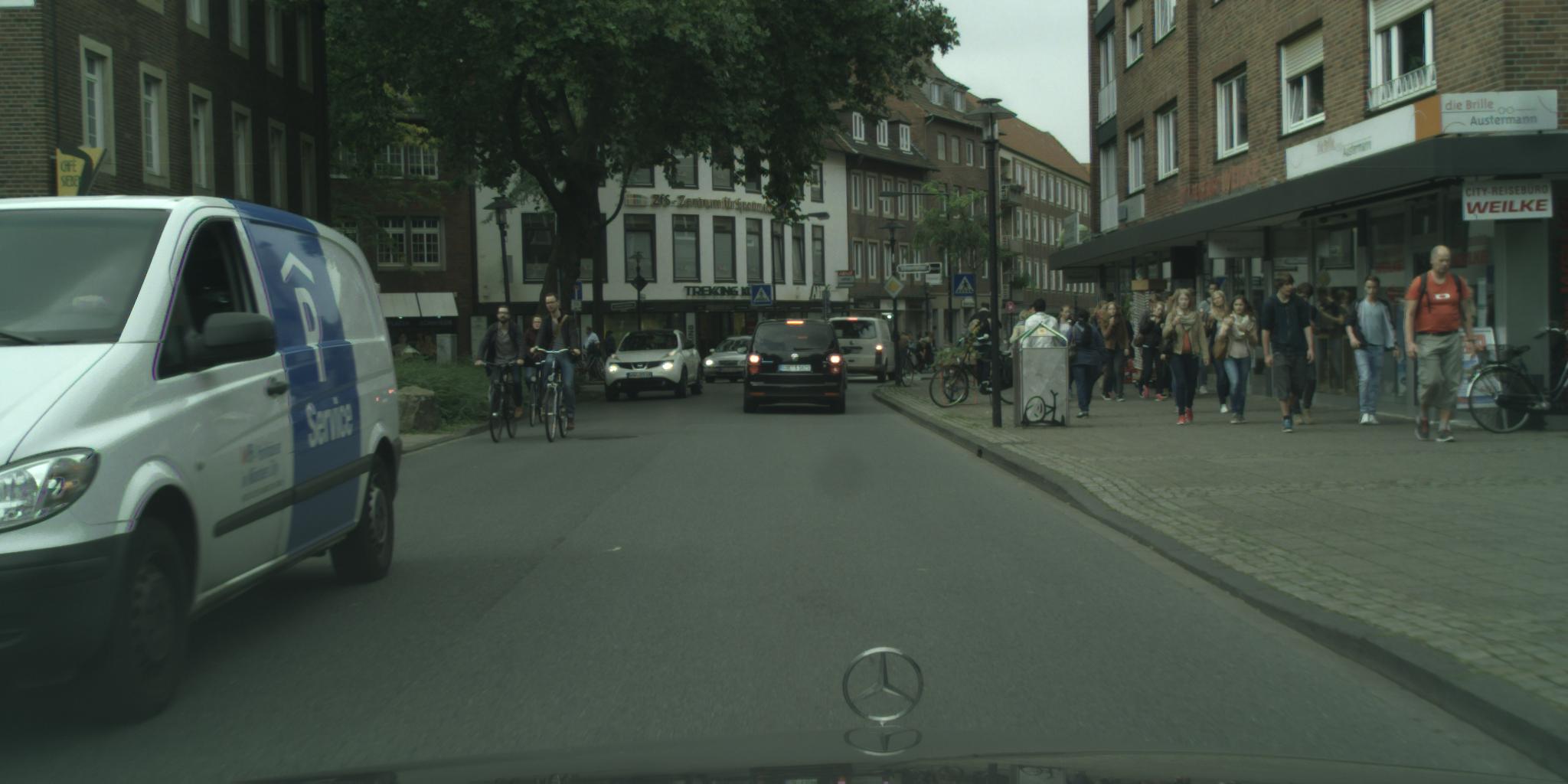} & 
         \includegraphics[width=0.24\textwidth]{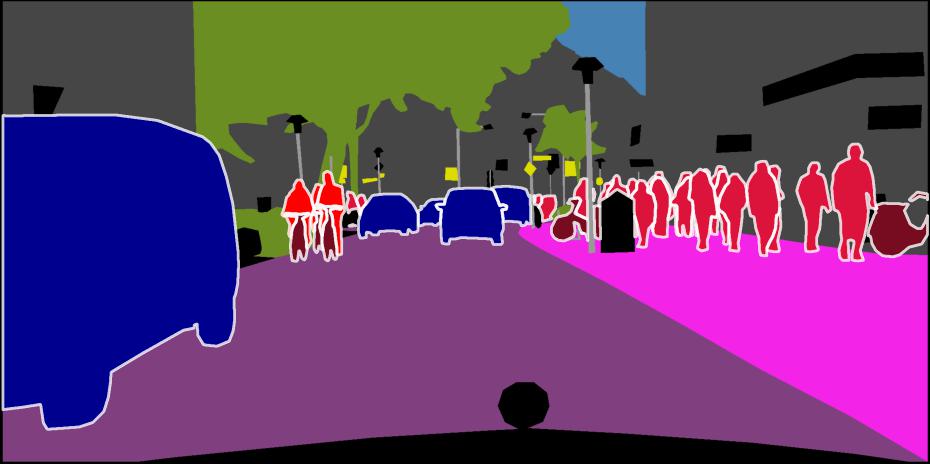} &
         \includegraphics[width=0.24\textwidth]{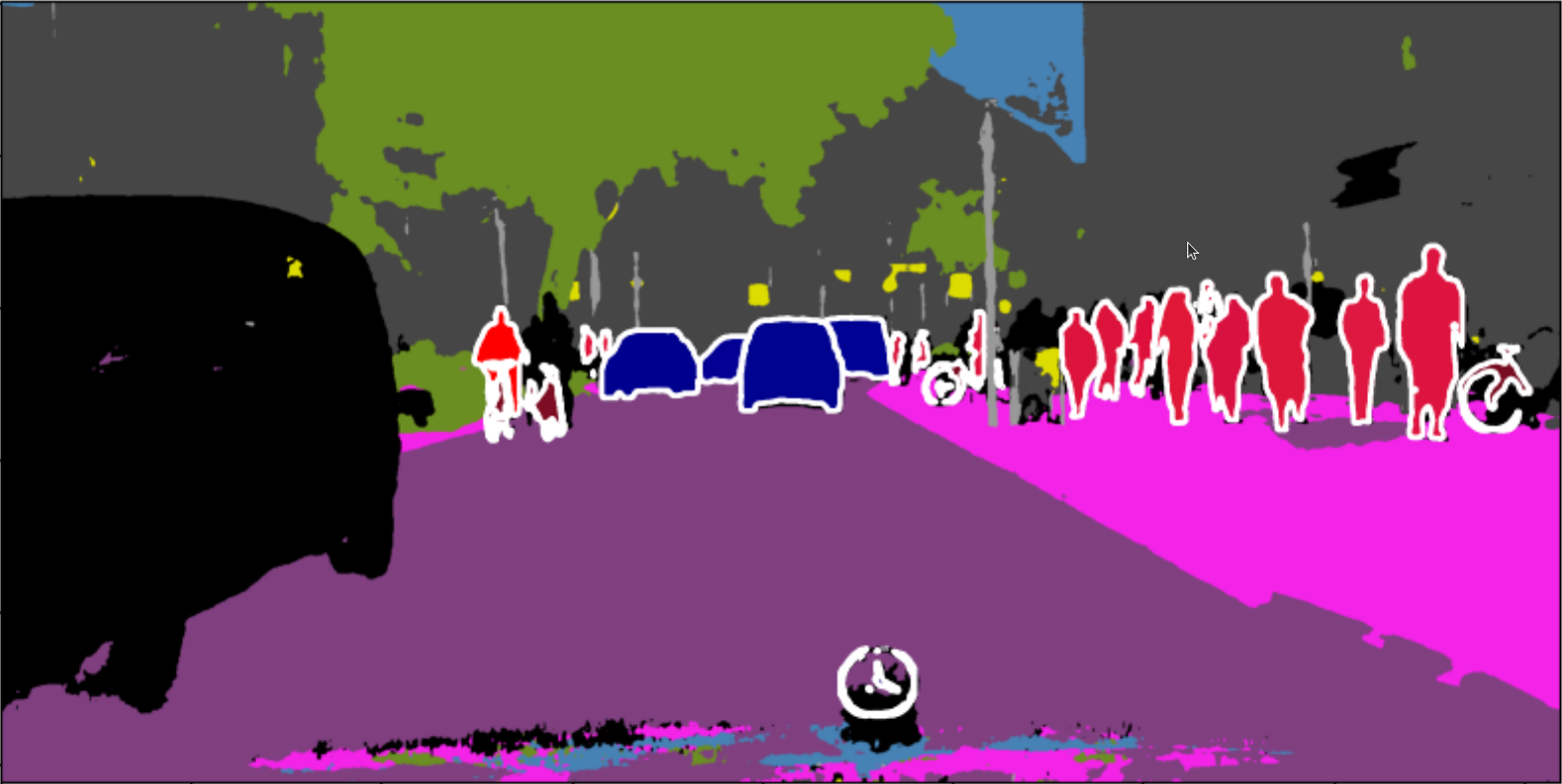} &
         \includegraphics[width=0.24\textwidth]{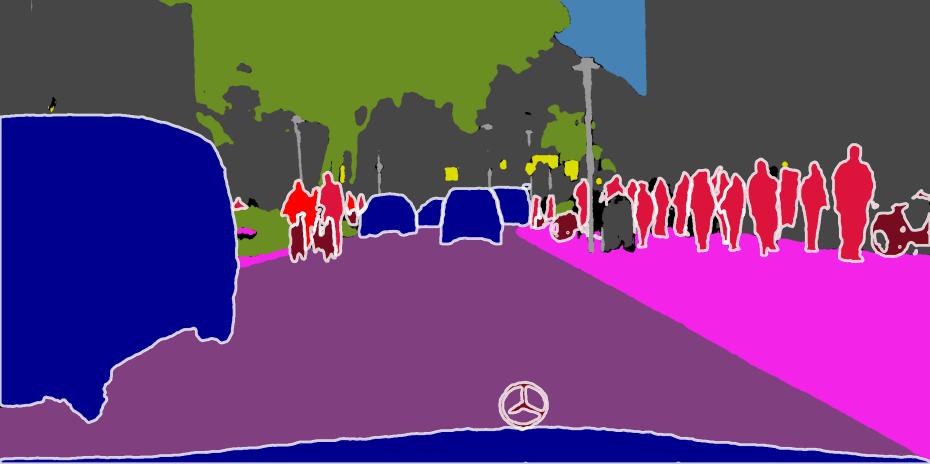} \\[-0.2em]

    \end{tabular}
    \caption{Qualitative comparison between our MC-PanDA and
     the state-of-the-art method EDAPS~\cite{Saha_2023_ICCV} on Synthia$\rightarrow$Cityscapes.}
    \label{fig:predictions_syn_cityscapes}
\end{figure}

\begin{figure}[t]
    \centering
    \begin{tabular}{c@{\,}c@{\,}c@{\,}c}
        \scriptsize Image & \scriptsize GT & \scriptsize EDAPS~\cite{Saha_2023_ICCV} & \scriptsize MC-PanDA \\
         \includegraphics[width=0.24\textwidth]{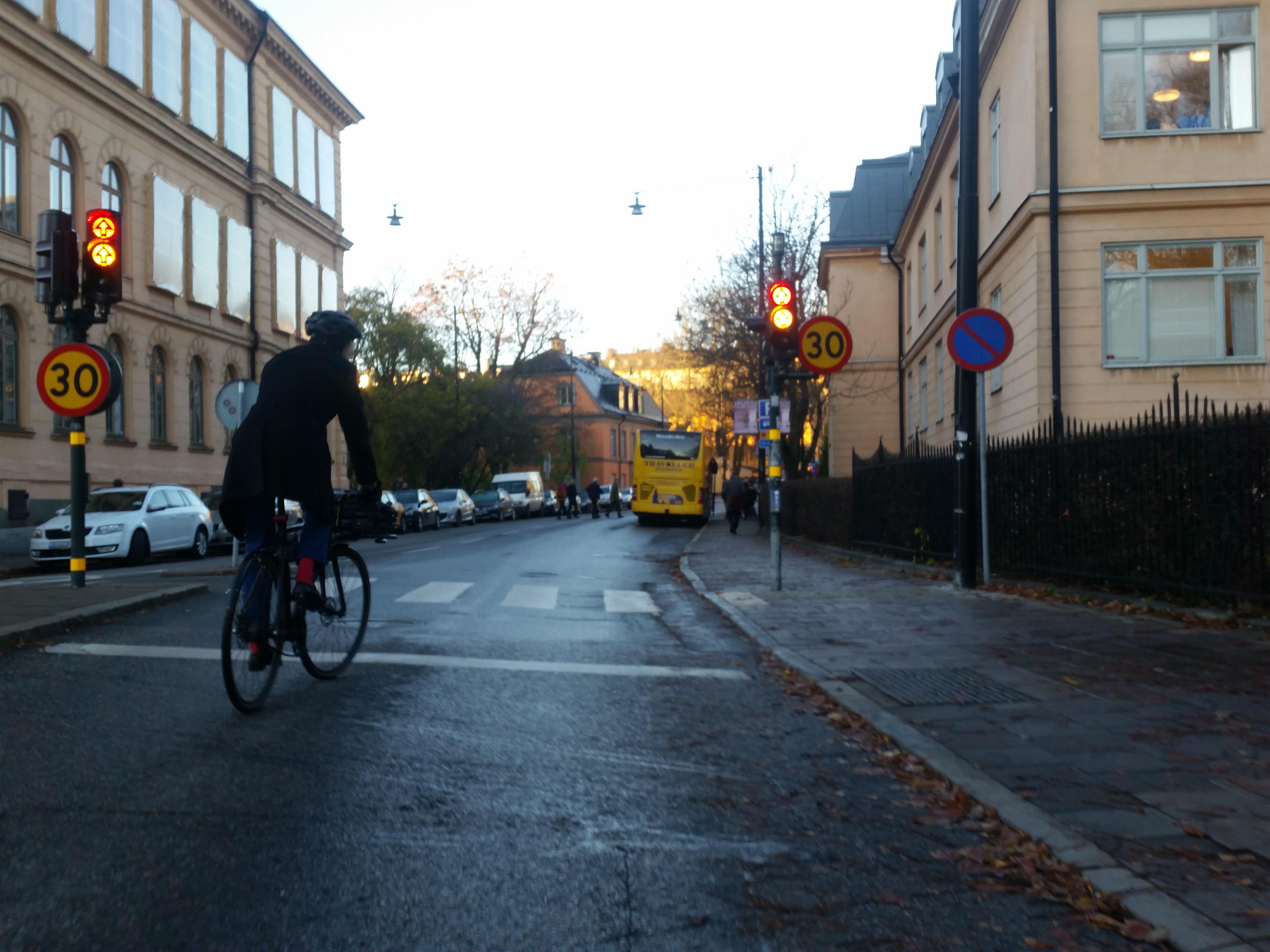} & 
         \includegraphics[width=0.24\textwidth]{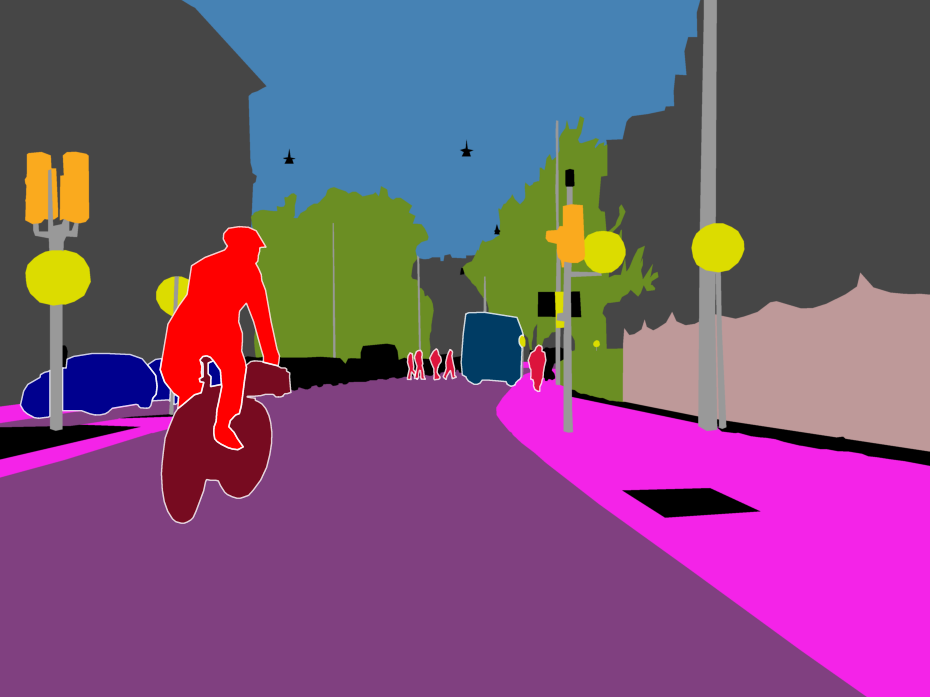} &
         \includegraphics[width=0.24\textwidth]{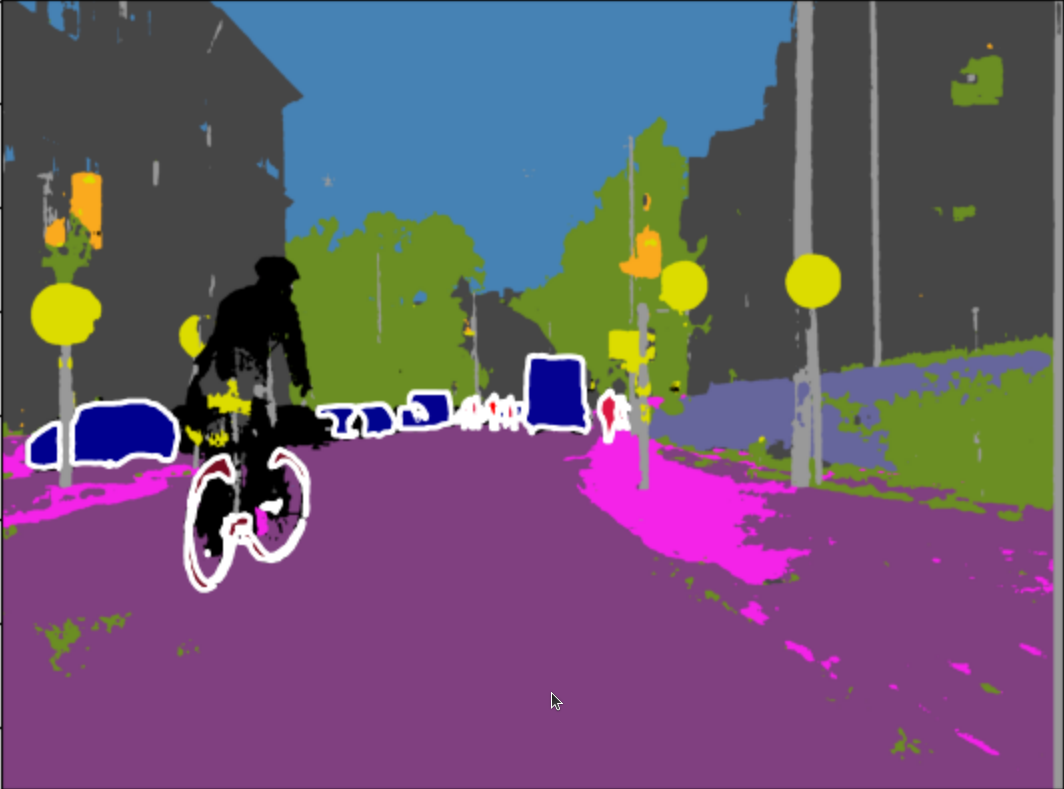} &
         \includegraphics[width=0.24\textwidth]{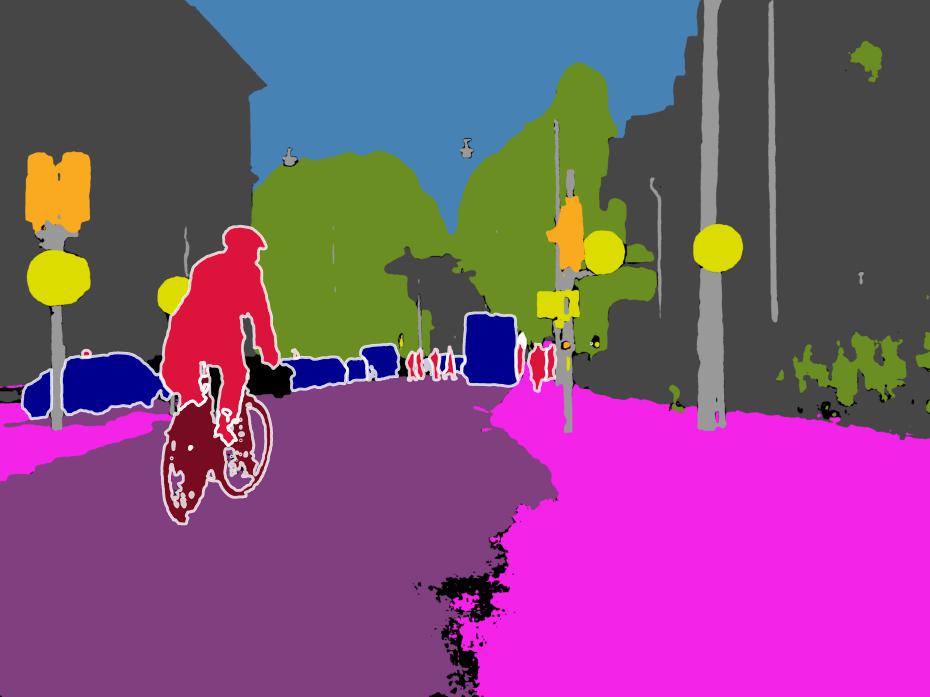} \\[-0.2em]

         \includegraphics[width=0.24\textwidth]{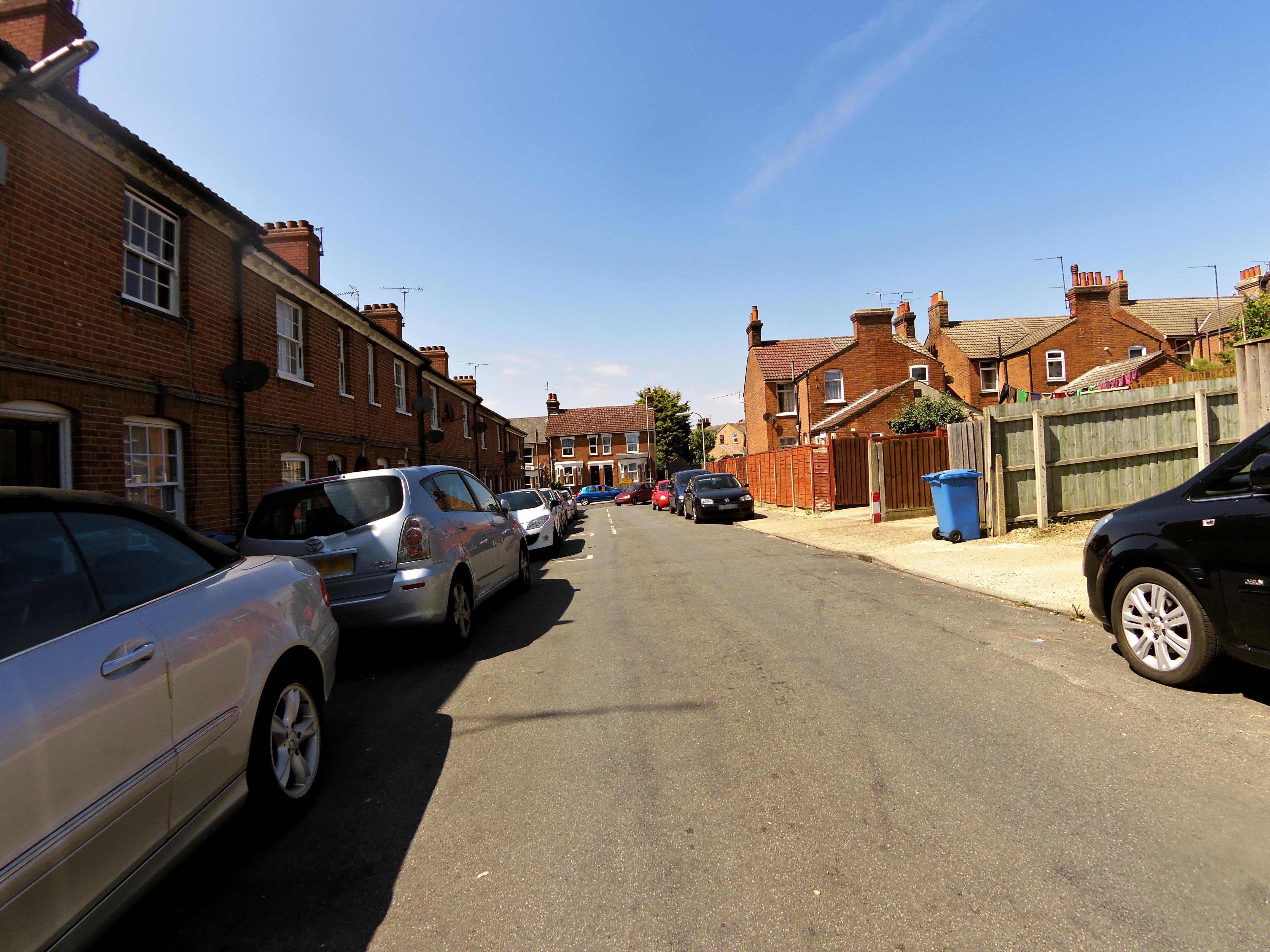} & 
         \includegraphics[width=0.24\textwidth]{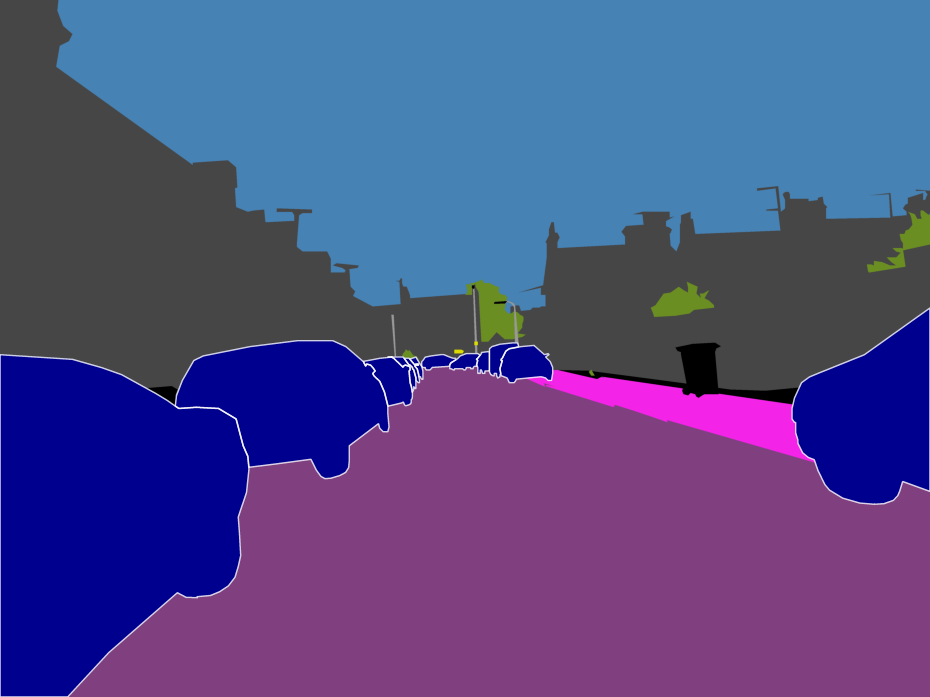} &
         \includegraphics[width=0.24\textwidth]{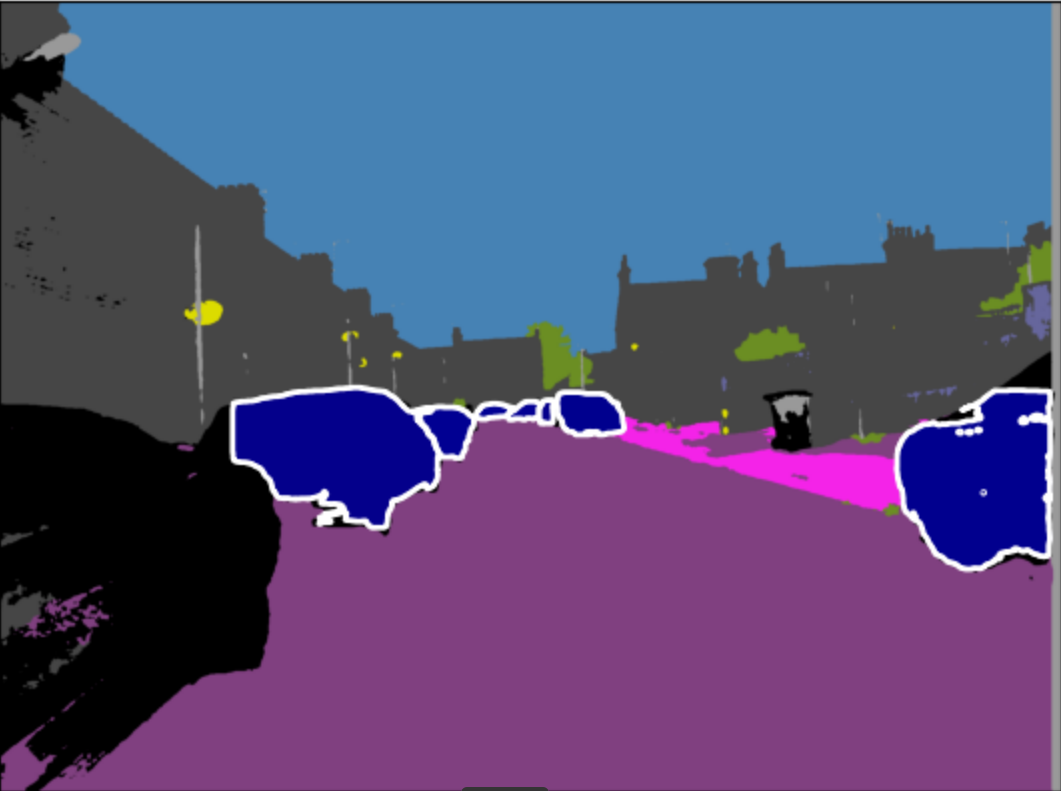} &
         \includegraphics[width=0.24\textwidth]{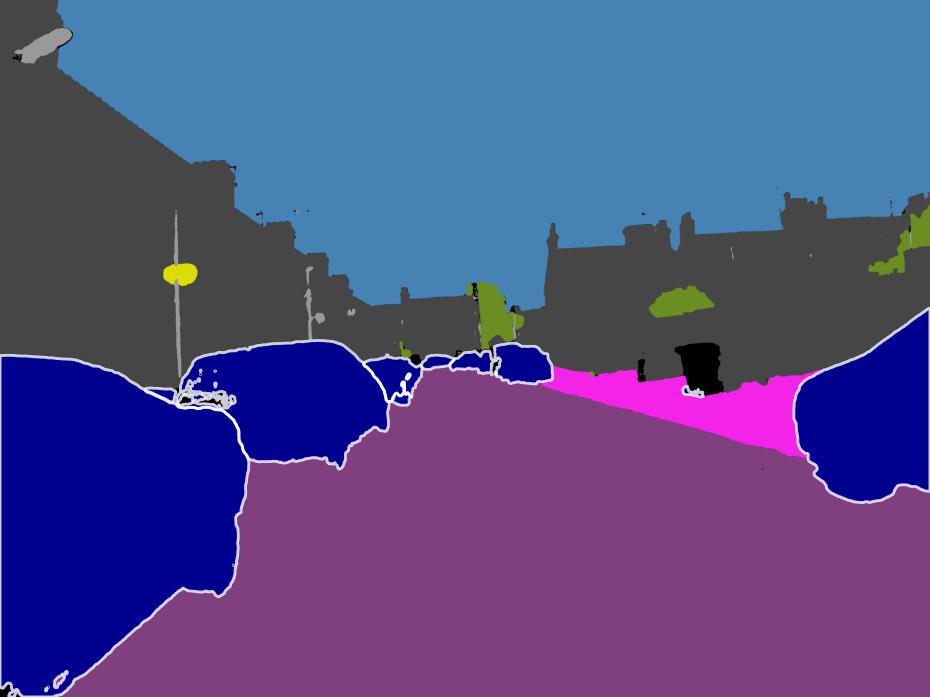} \\[-0.2em]

         \includegraphics[width=0.24\textwidth]{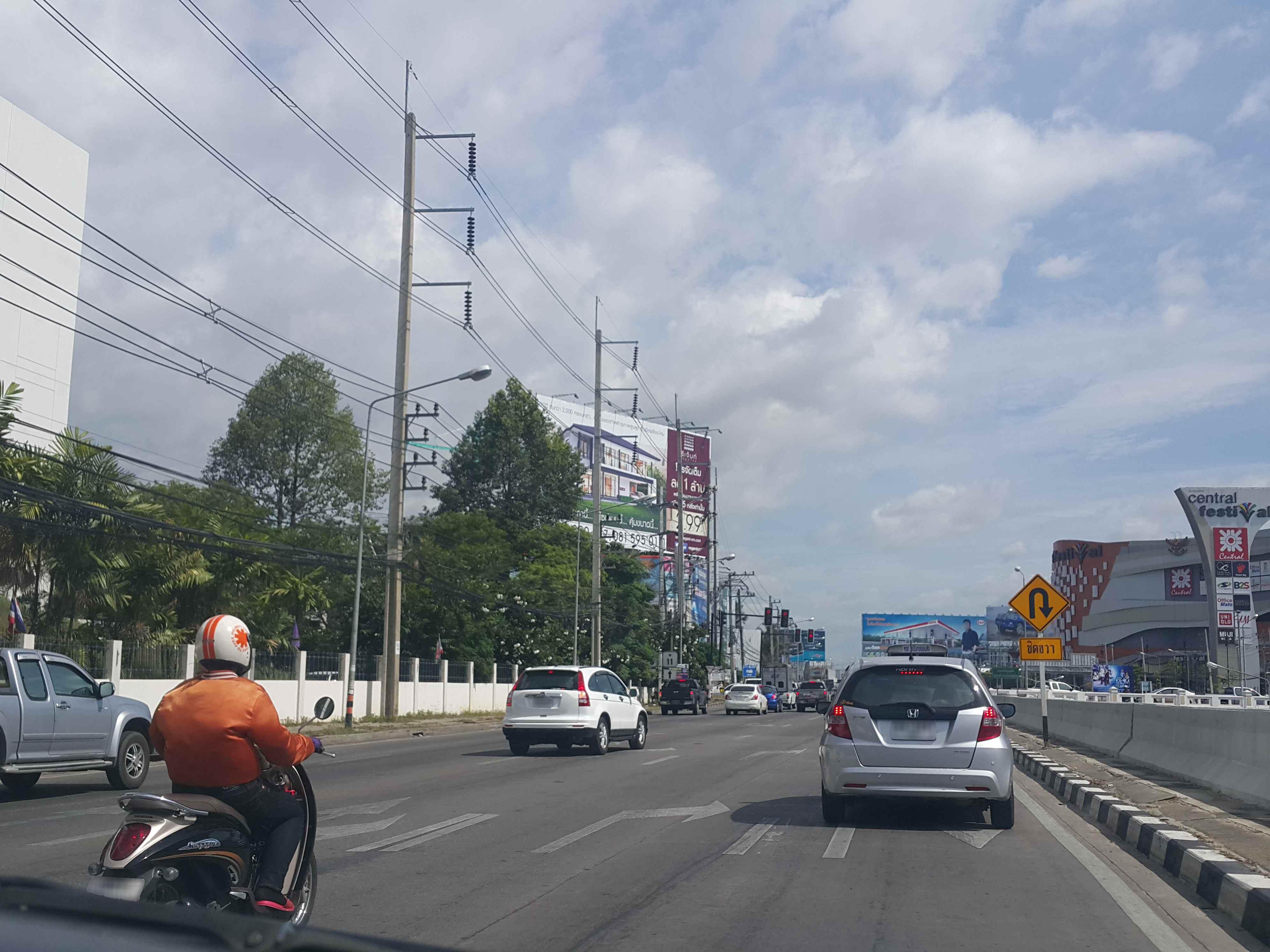} & 
         \includegraphics[width=0.24\textwidth]{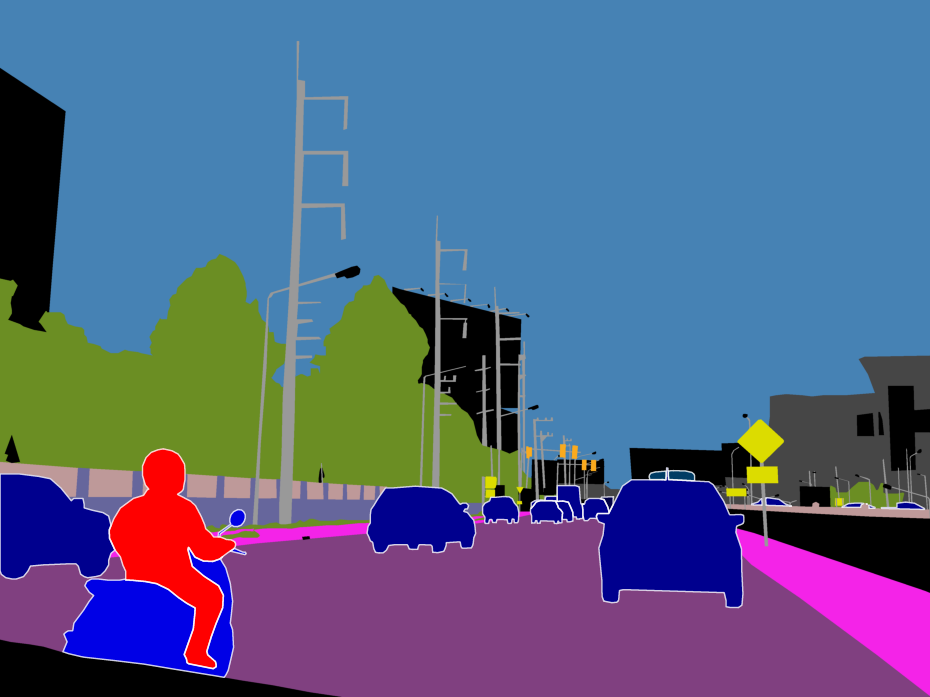} &
         \includegraphics[width=0.24\textwidth]{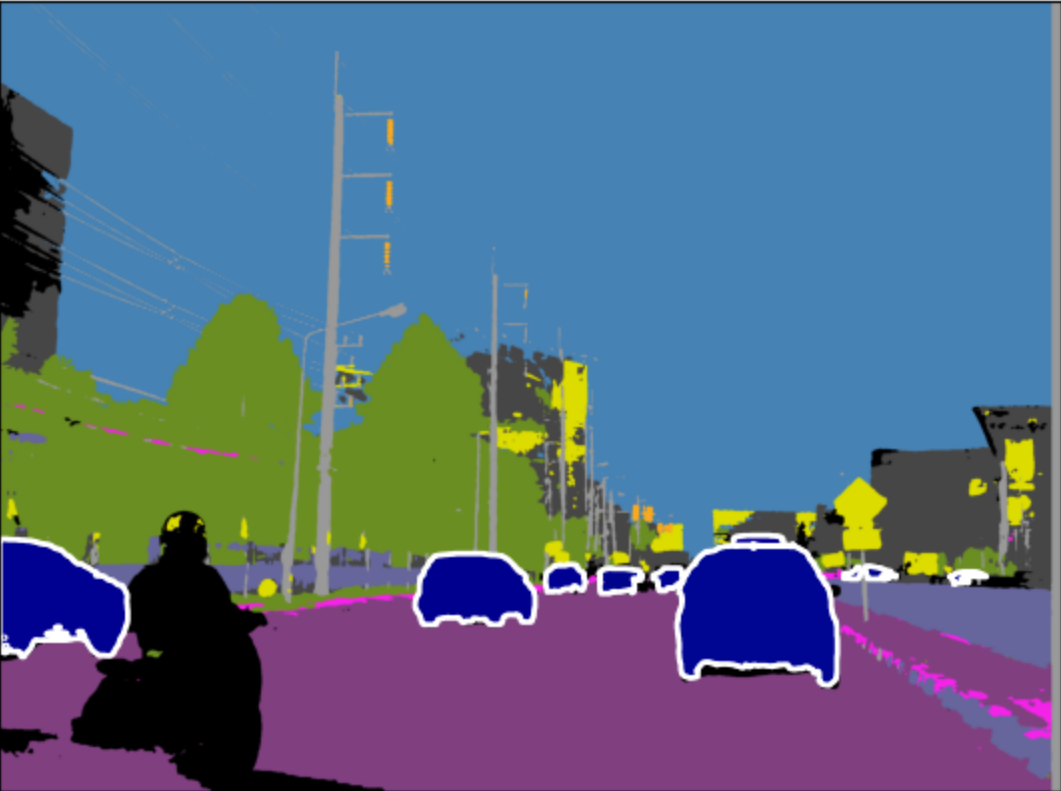} &
         \includegraphics[width=0.24\textwidth]{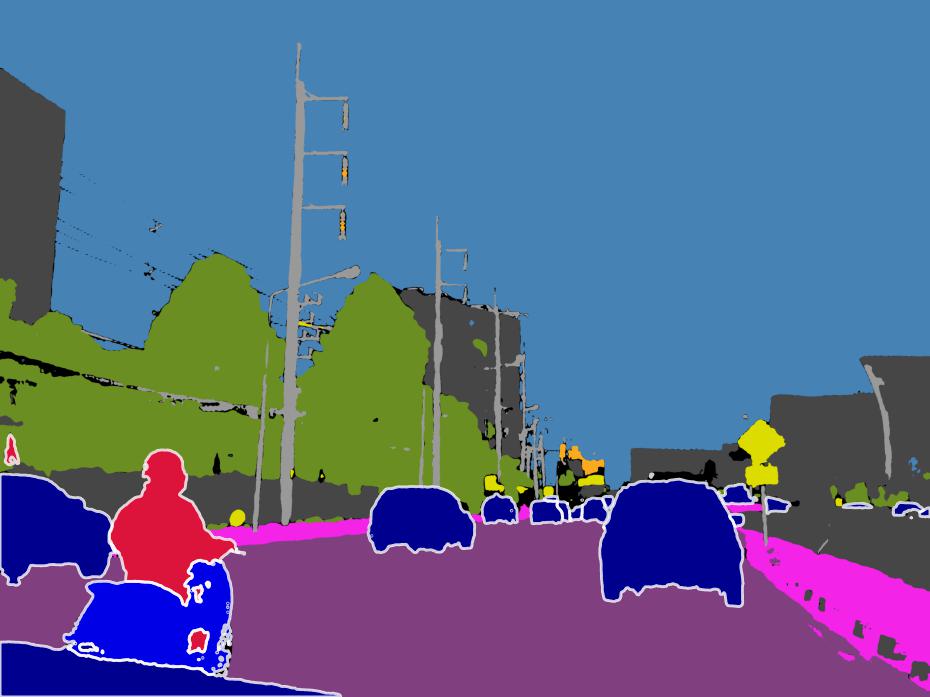} \\[-0.2em]

         \includegraphics[width=0.24\textwidth]{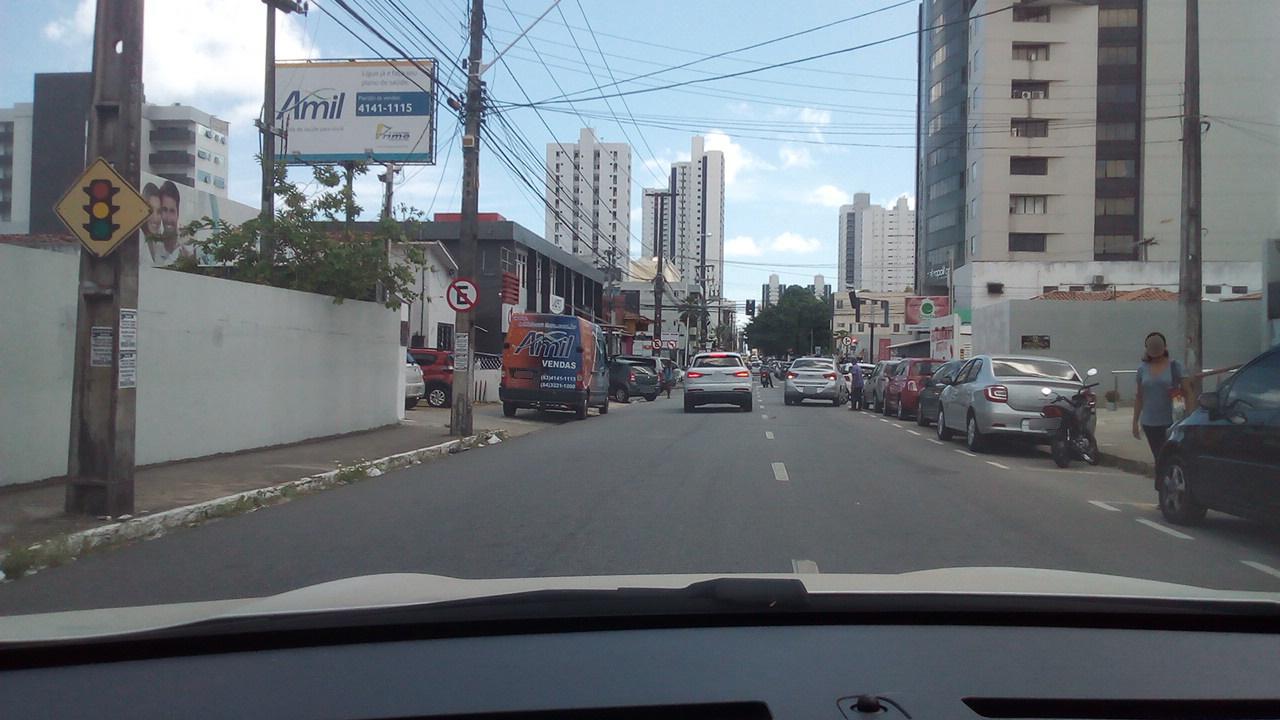} & 
         \includegraphics[width=0.24\textwidth]{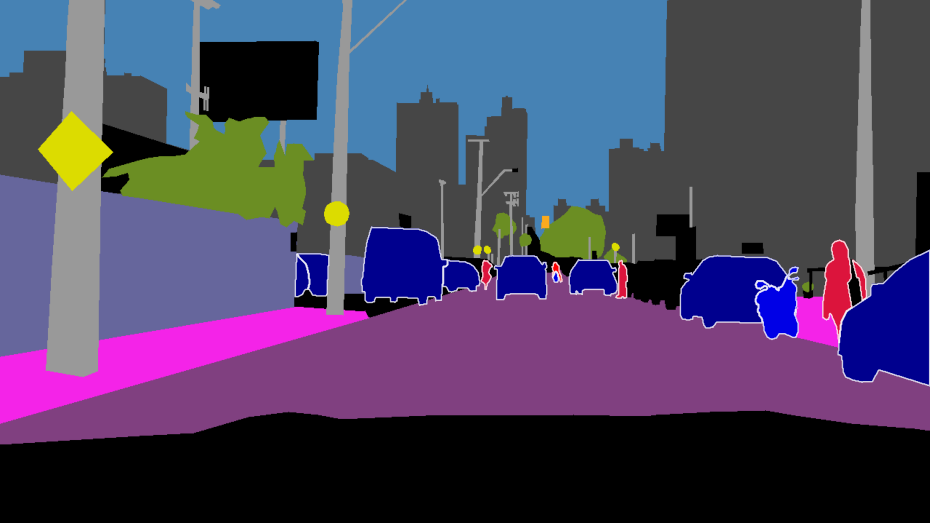} &
         \includegraphics[width=0.24\textwidth,height=0.135\textwidth]{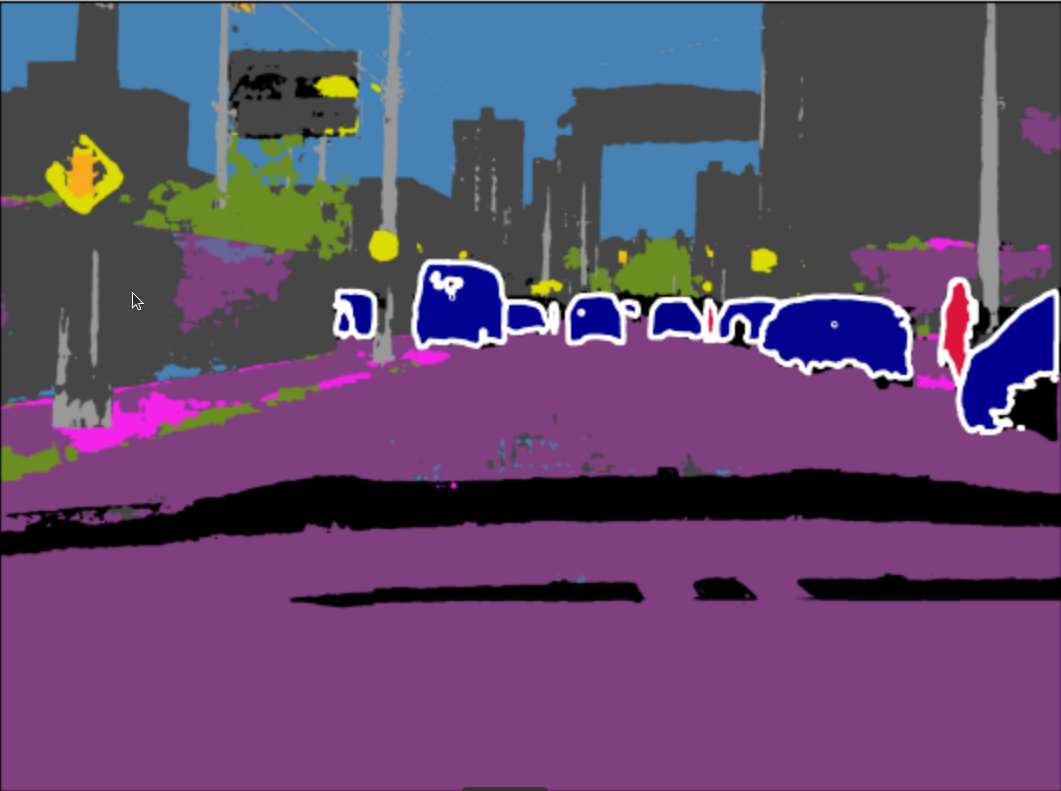} &
         \includegraphics[width=0.24\textwidth]{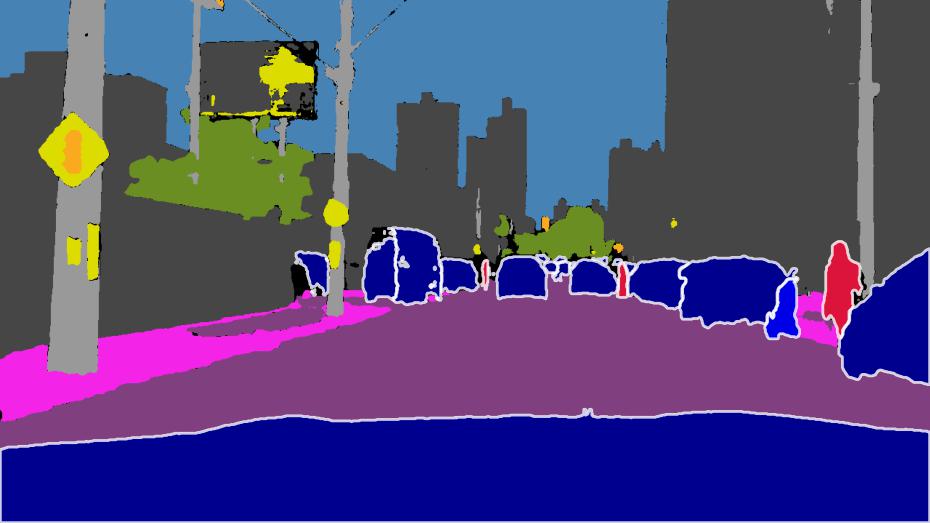} \\[-0.2em]

    \end{tabular}
    \caption{Qualitative comparison between our MC-PanDA and
     the state-of-the-art method EDAPS~\cite{Saha_2023_ICCV} on Synthia$\rightarrow$Vistas.}
    \label{fig:predictions_syn_vistas}
\end{figure}

\begin{figure}[t]
    \centering
    \begin{tabular}{c@{\,}c@{\,}c}
        \scriptsize Image & \scriptsize GT &  \scriptsize MC-PanDA \\
         \includegraphics[width=0.32\textwidth]{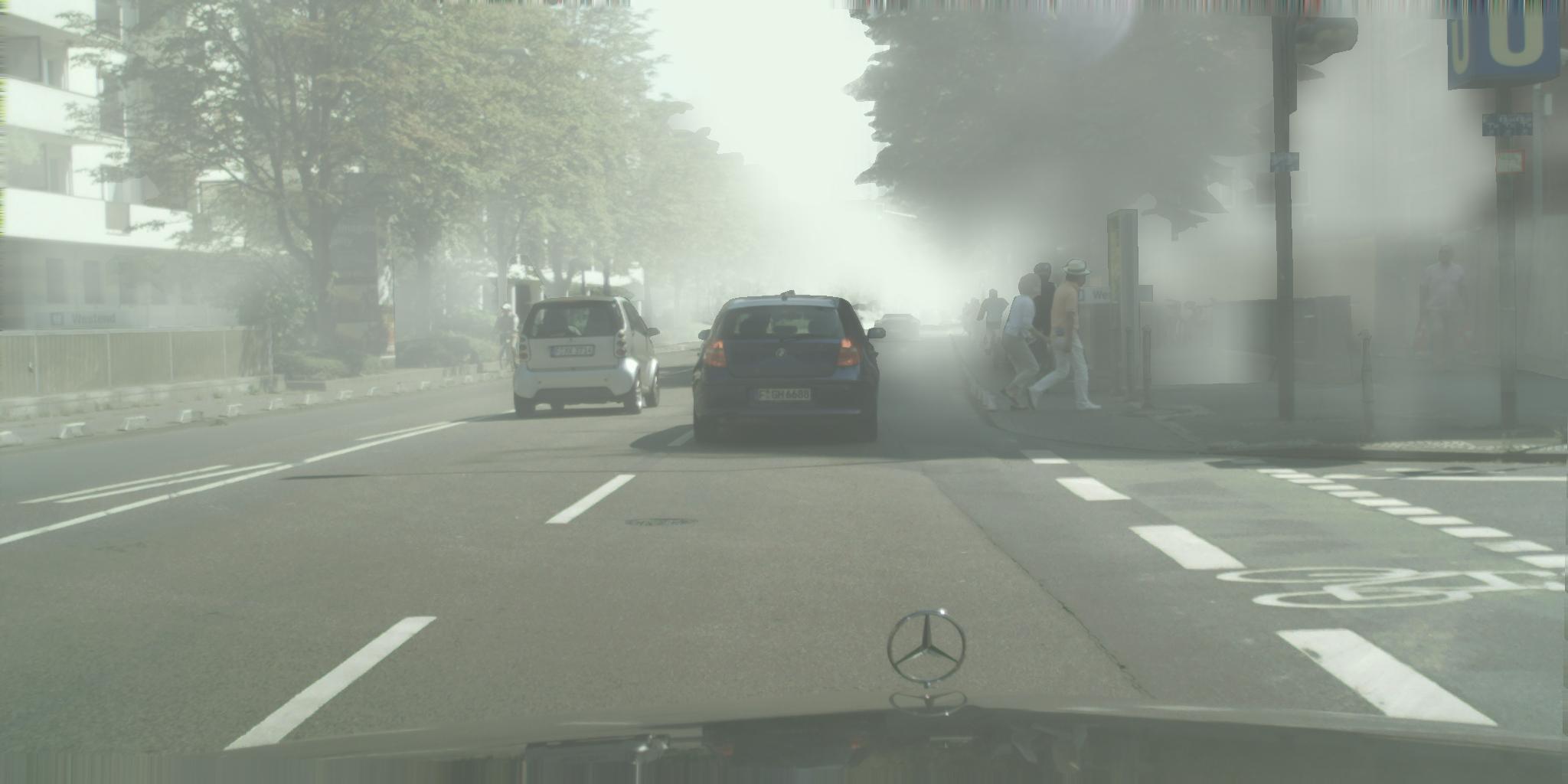} & 
         \includegraphics[width=0.32\textwidth]{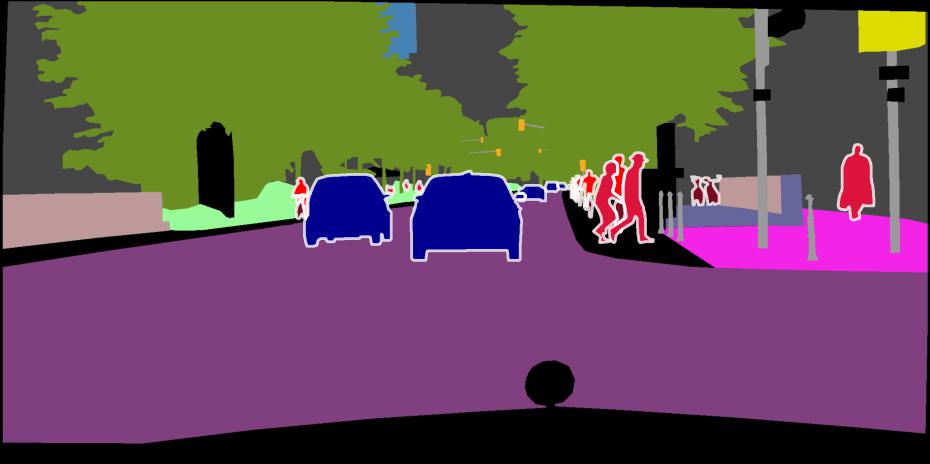} &
         \includegraphics[width=0.32\textwidth]{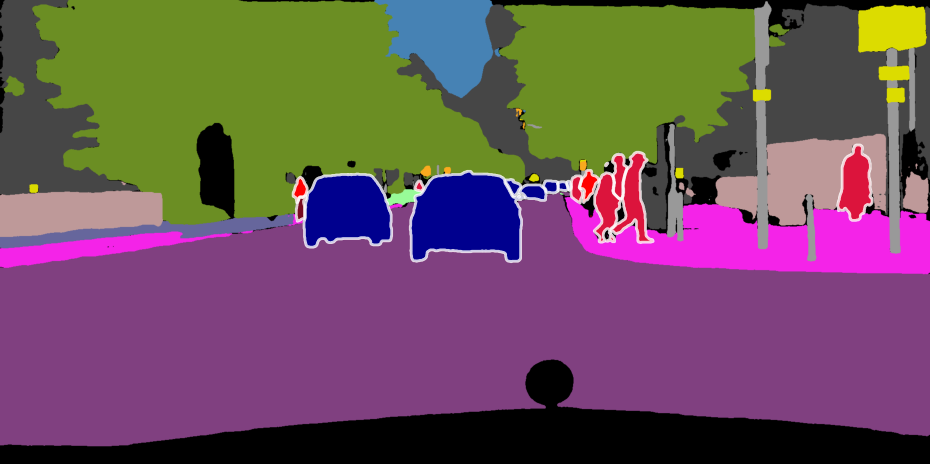} \\[-0.2em]

        \includegraphics[width=0.32\textwidth]{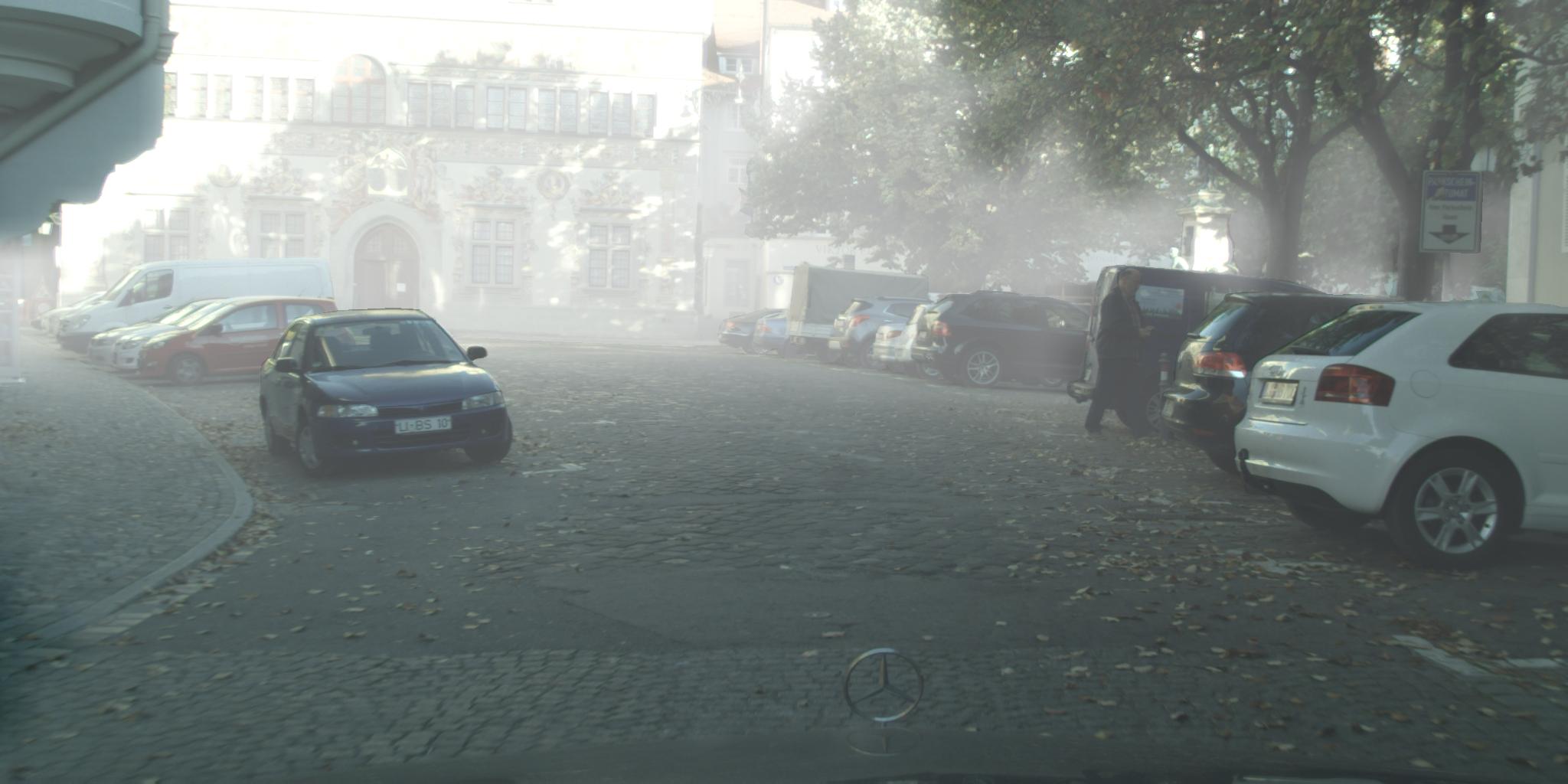} & 
         \includegraphics[width=0.32\textwidth]{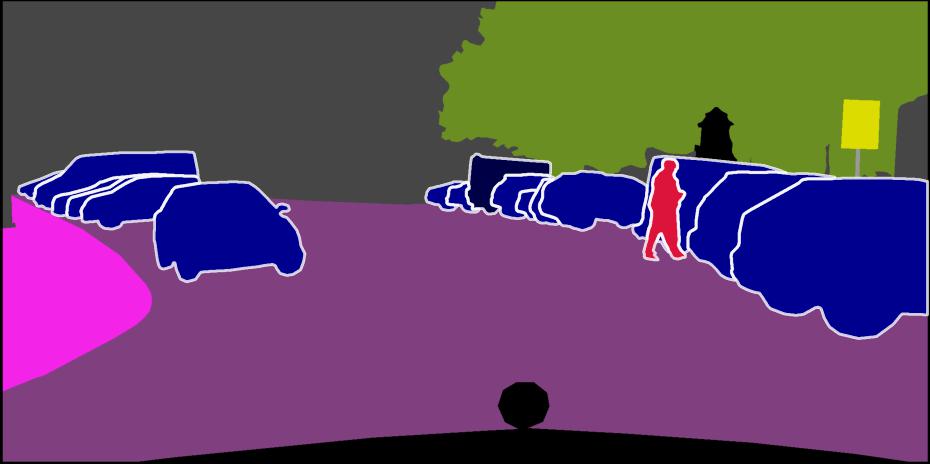} &
         \includegraphics[width=0.32\textwidth]{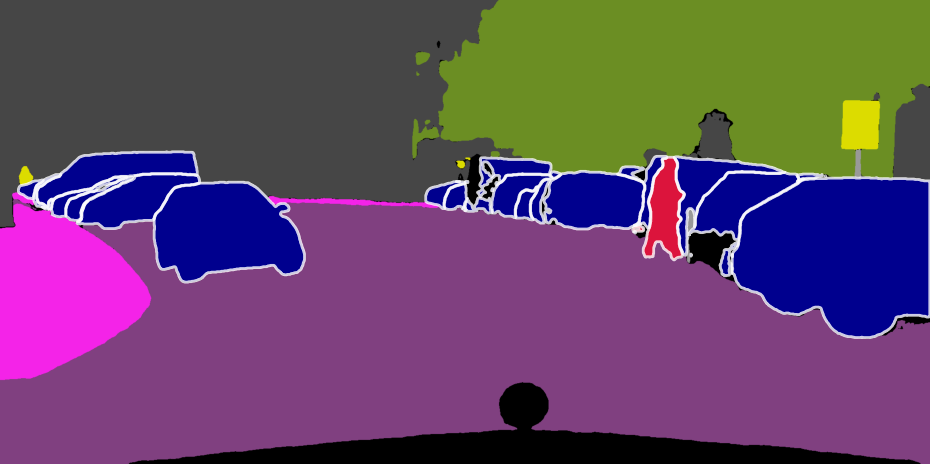} \\[-0.2em]

        \includegraphics[width=0.32\textwidth]{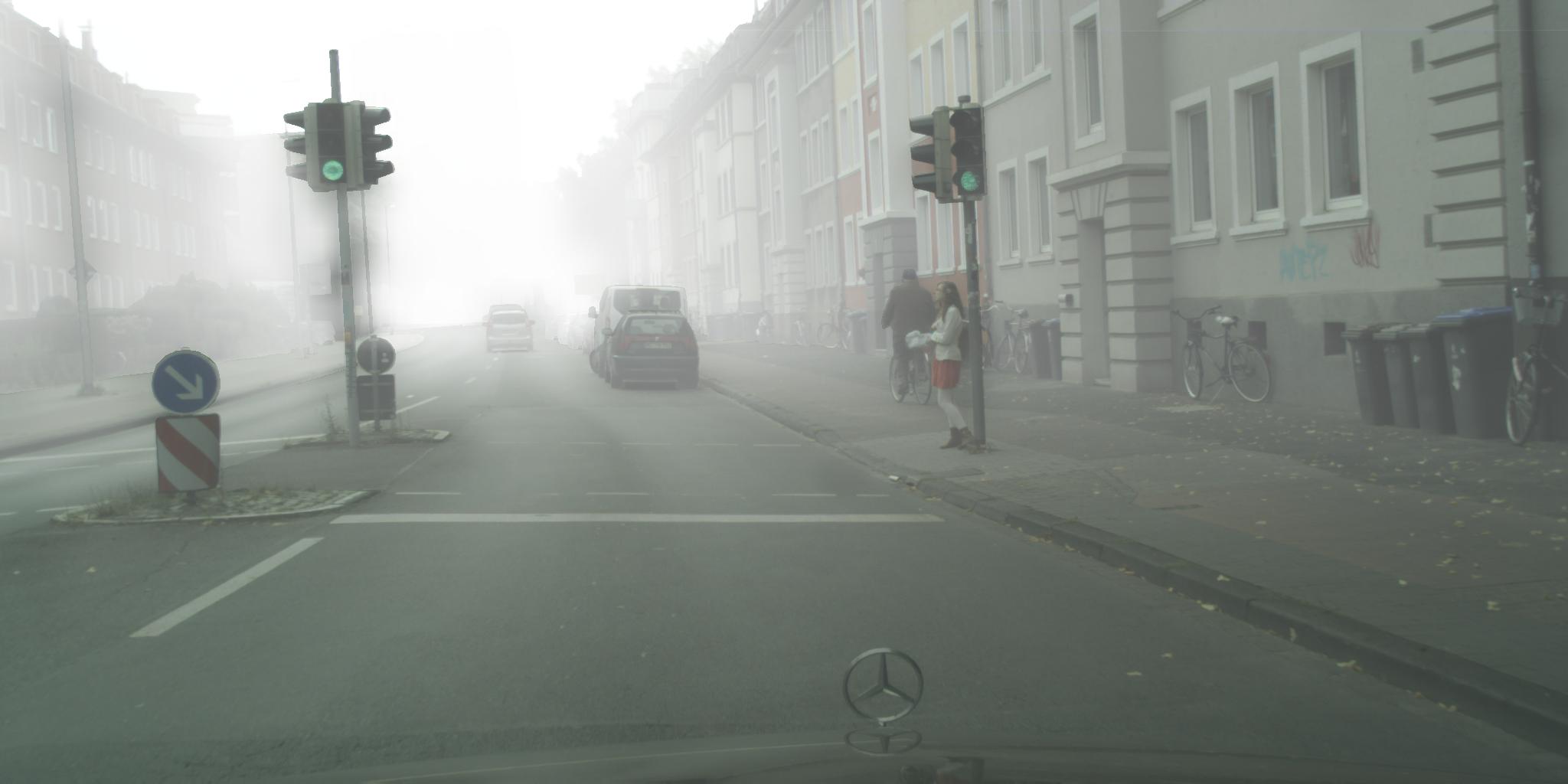} & 
         \includegraphics[width=0.32\textwidth]{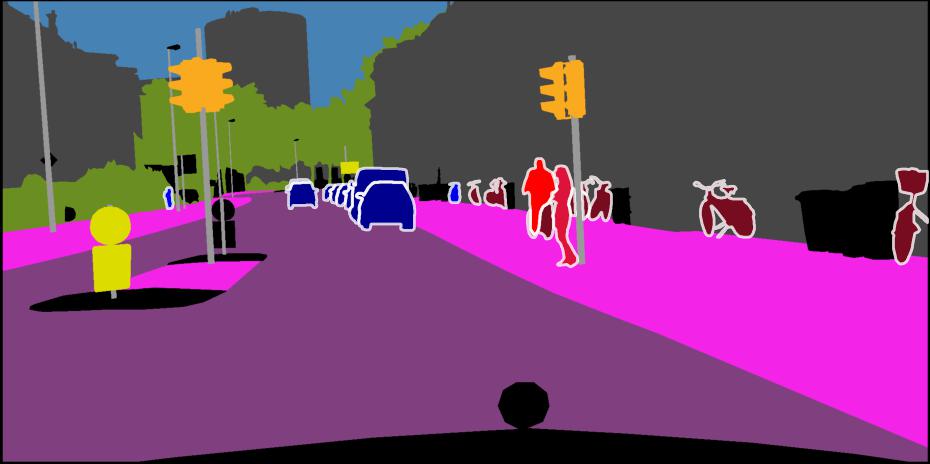} &
         \includegraphics[width=0.32\textwidth]{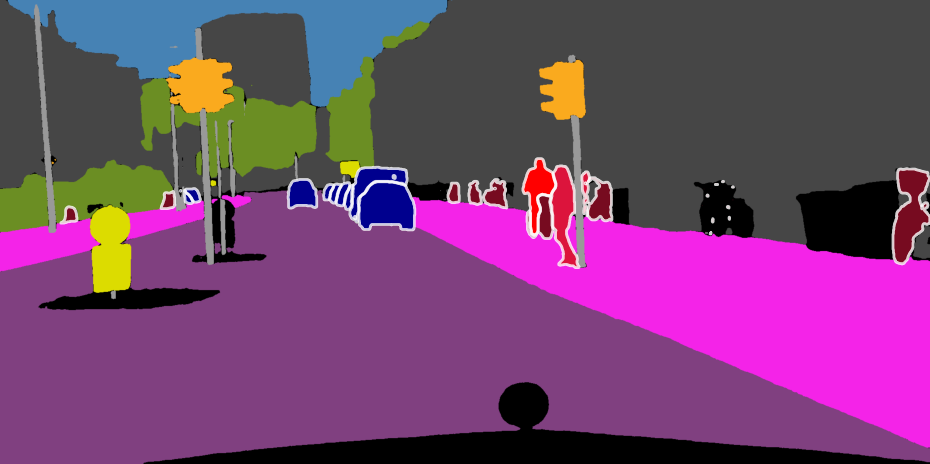} \\[-0.2em]

    \end{tabular}
    \caption{Prediction examples on Cityscapes $\rightarrow$ Foggy Cityscapes.}
    \label{fig:predictions_city_foggy}
\end{figure}

\begin{figure}[t]
    \centering
    \begin{tabular}{c@{\,}c@{\,}c}
        \scriptsize Image & \scriptsize GT &  \scriptsize MC-PanDA \\
        
        \includegraphics[width=0.32\textwidth]{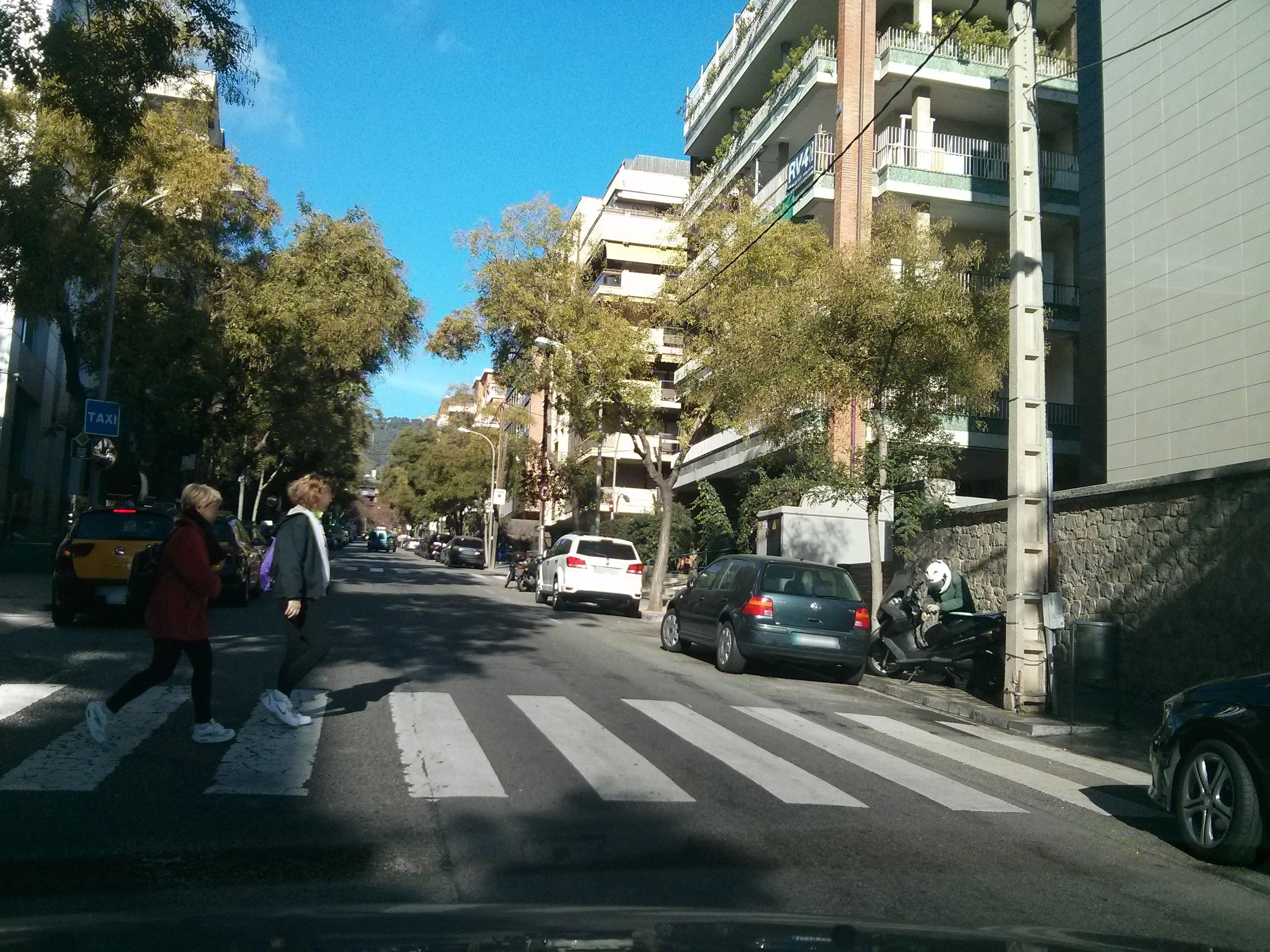} & 
         \includegraphics[width=0.32\textwidth]{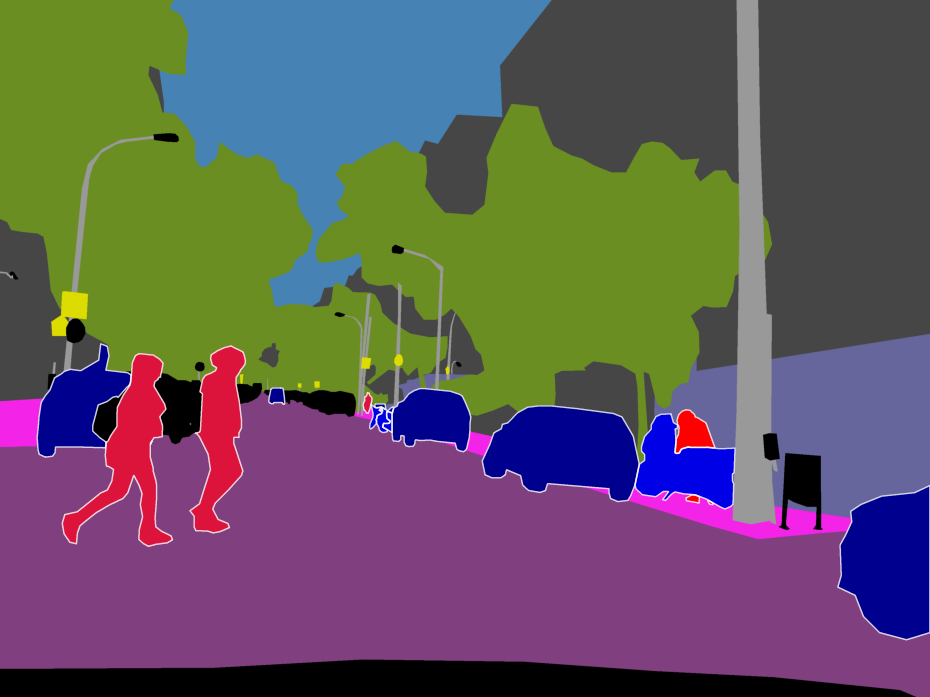} &
         \includegraphics[width=0.32\textwidth]{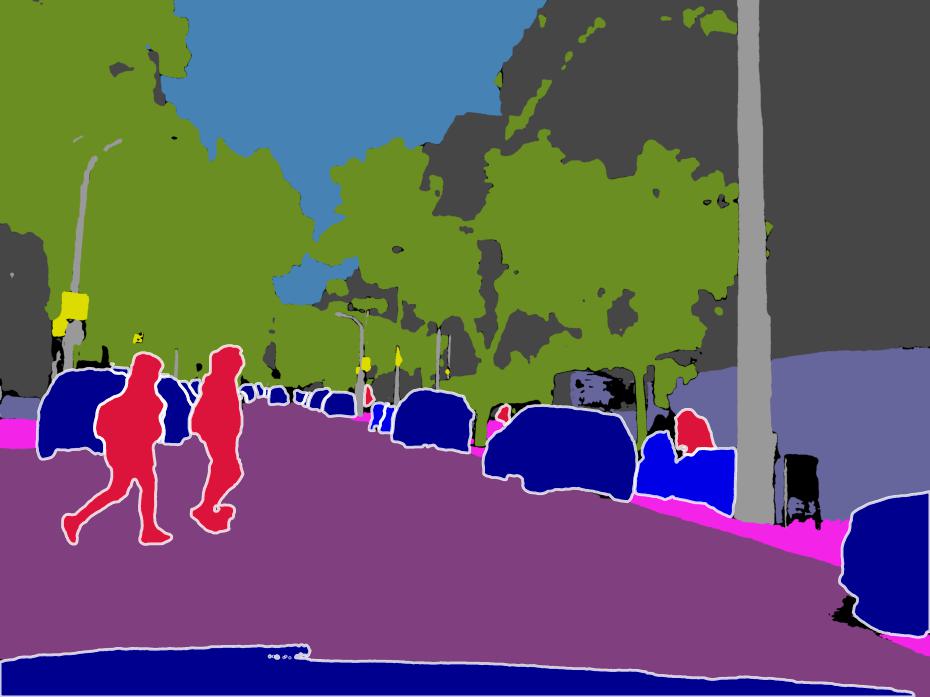} \\[-0.2em]

        \includegraphics[width=0.32\textwidth]{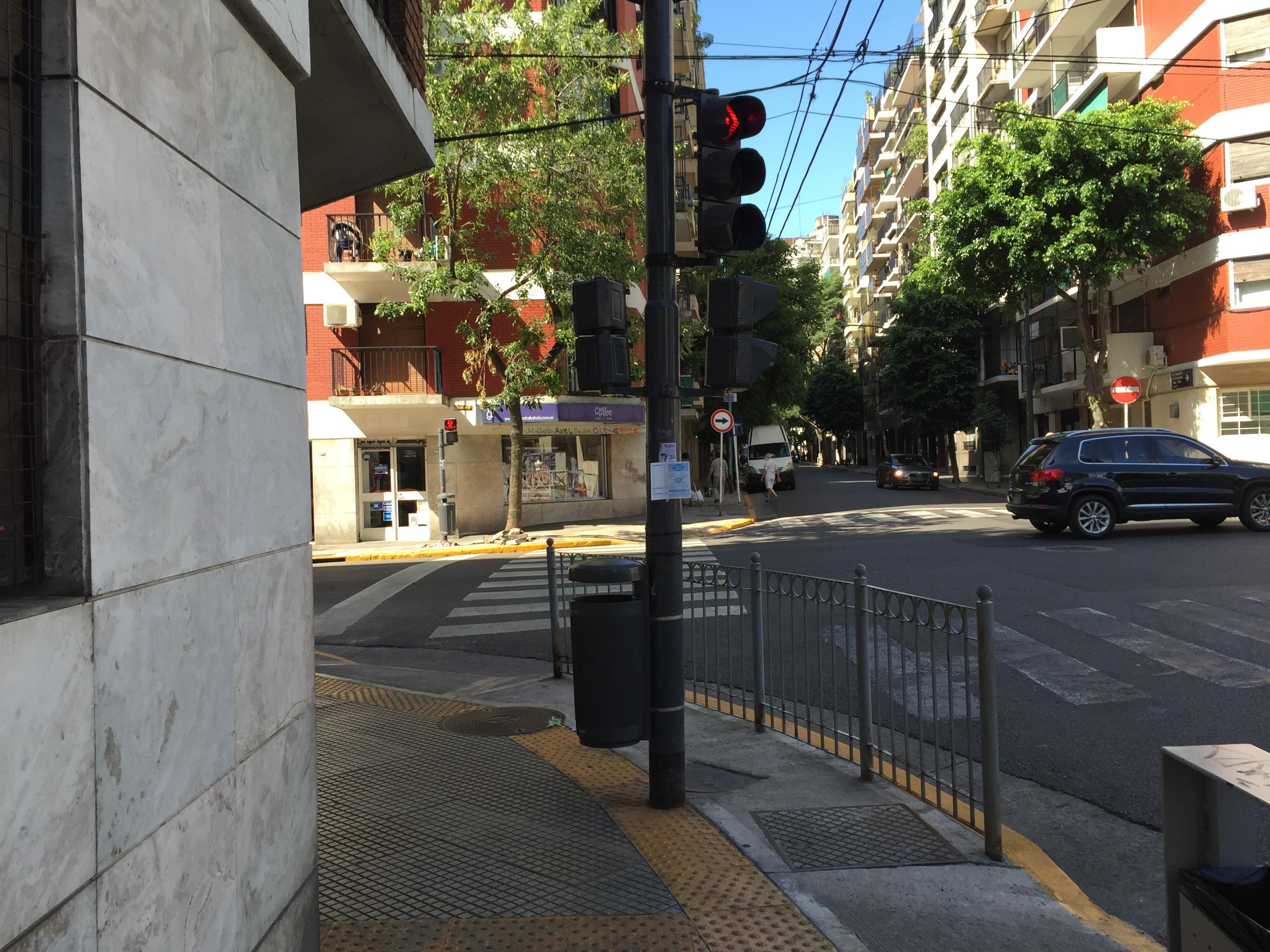} & 
         \includegraphics[width=0.32\textwidth]{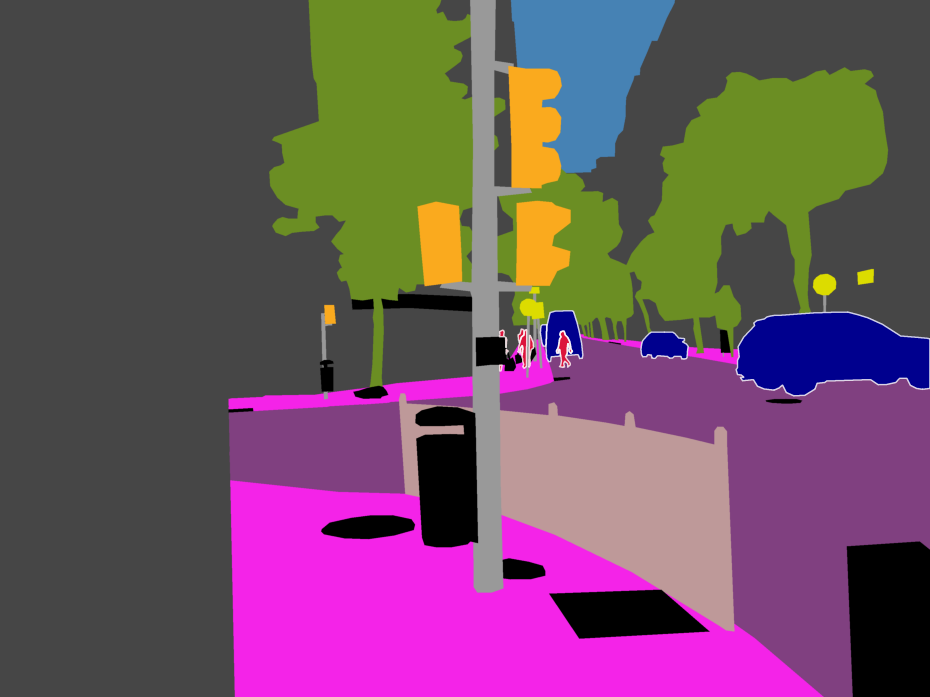} &
         \includegraphics[width=0.32\textwidth]{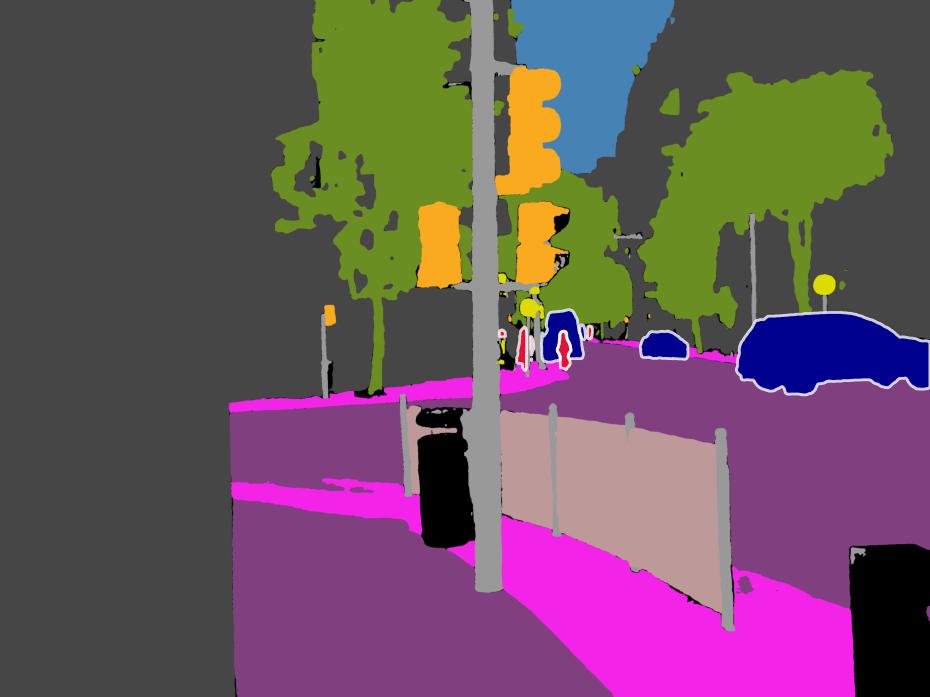} \\[-0.2em]

        \includegraphics[width=0.32\textwidth]{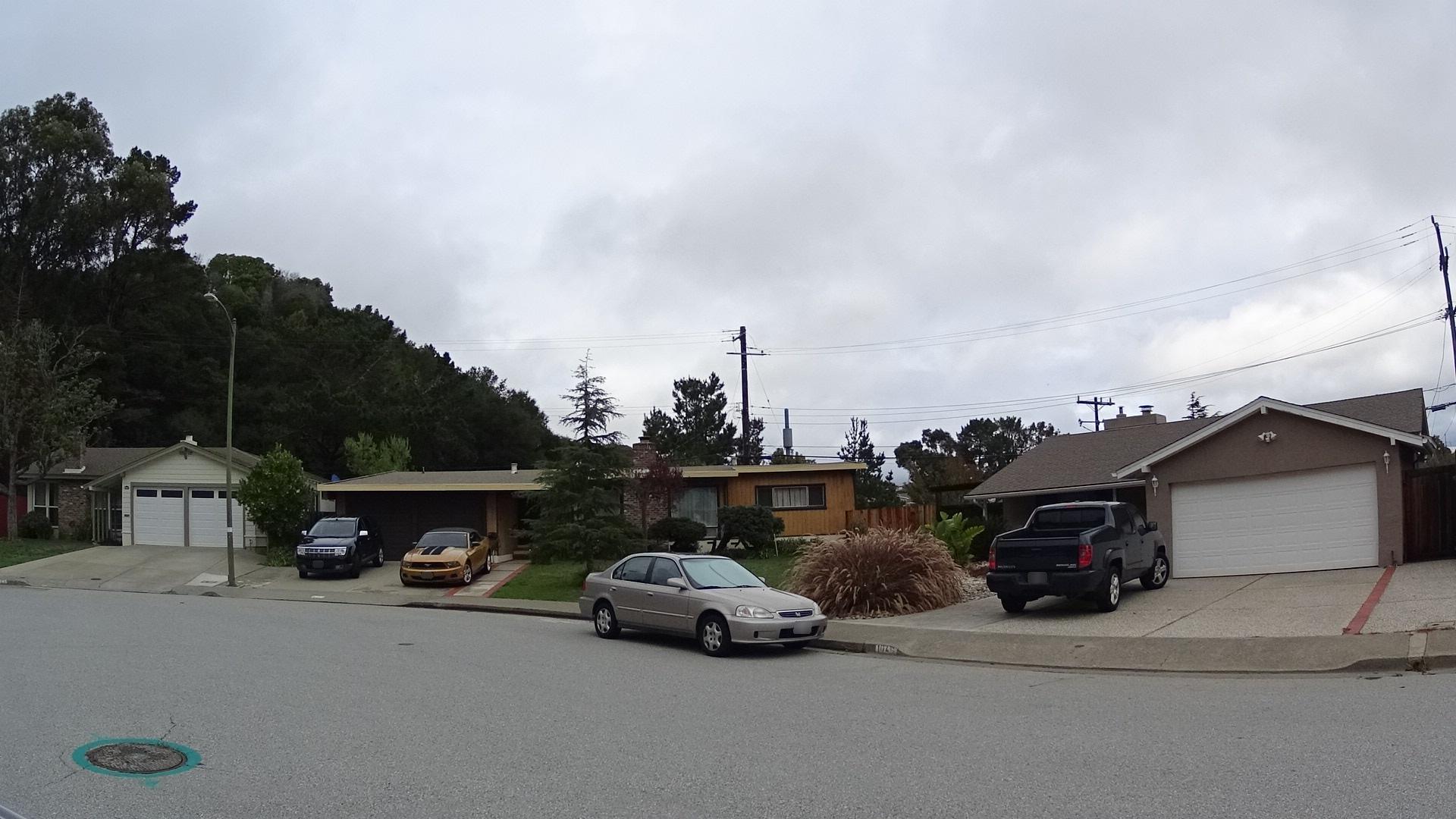} & 
         \includegraphics[width=0.32\textwidth]{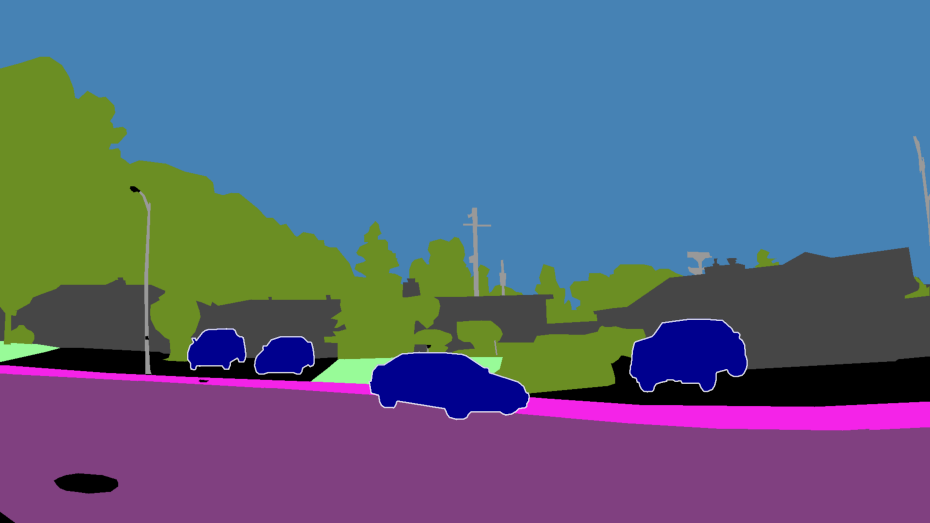} &
         \includegraphics[width=0.32\textwidth]{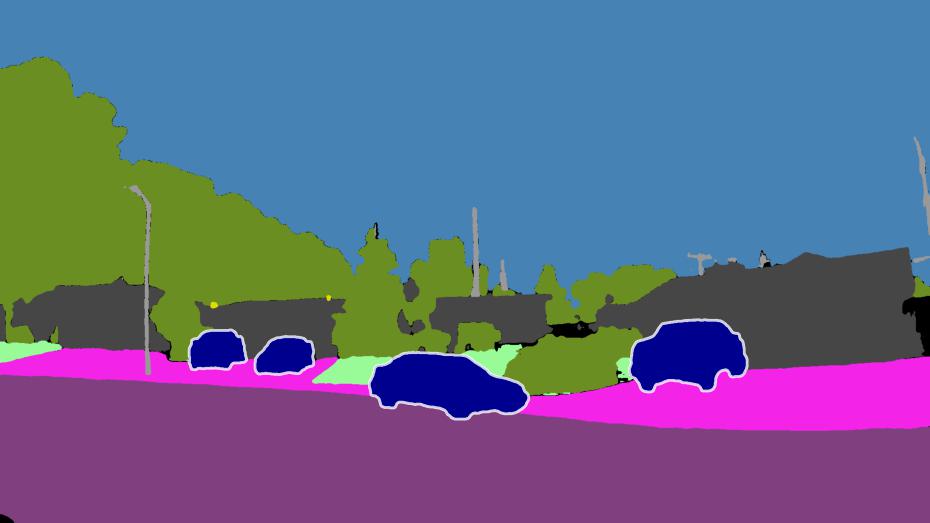} \\[-0.2em]

    \end{tabular}
    \caption{Prediction examples on Cityscapes$\rightarrow$Vistas.}
    \label{fig:predictions_city_vistas}
\end{figure}

We complete this appendix by 
presenting qualitative experiments 
on test target images 
for the four domain adaptation benchmarks
from the main manuscript. 
Figures~\ref{fig:predictions_syn_cityscapes} 
and~\ref{fig:predictions_syn_vistas} 
illustrate predictions of MC-PanDA (single-scale inference) 
for the Synthia$\rightarrow$Cityscapes 
and Synthia$\rightarrow$ Vistas, respectively, 
and offer a comparison 
with the state-of-the-art method EDAPS~\cite{Saha_2023_ICCV}. 
We generate EDAPS predictions 
by applying single-scale 
inference with the
publicly available weights\footnote{\url{https://github.com/susaha/edaps}}.
Figures~\ref{fig:predictions_city_foggy} 
and~\ref{fig:predictions_city_vistas} 
present predictions for
the Cityscapes$\rightarrow$Foggy 
and Cityscapes $\rightarrow$ Vistas scenarios, respectively. 
EDAPS predictions are not included 
due to lack 
of public weights 
for these configurations.

\end{document}